\PassOptionsToPackage{table,dvipsnames}{xcolor}
\documentclass[]{viocean}
\pdfoutput=1

\usepackage{latexsym}
\usepackage{url}
\usepackage{amssymb}
\usepackage[utf8]{inputenc}
\usepackage{pifont}
\usepackage{makecell}
\usepackage{paralist}
\usepackage{xspace}
\usepackage{colortbl}
\usepackage{adjustbox}
\usepackage{tikz}
\usepackage{amsfonts}

\usetikzlibrary{shapes,arrows.meta,positioning,fit,backgrounds,calc,decorations.pathreplacing}

\makeatletter
\DeclareRobustCommand\onedot{\futurelet\@let@token\@onedot}
\def\@onedot{\ifx\@let@token.\else.\null\fi\xspace}
\def\eg{\emph{e.g}\onedot}

\def\vs{\emph{vs}\onedot}
\def\wrt{w.r.t\onedot}

\makeatother

\newcommand{\ourBench}{Diagram-MMU\xspace}

\newcommand{\bcmark}{\color{Violet}{\ding{51}}}%
\newcommand{\cmark}{\color{OliveGreen}{\ding{51}}}%
\newcommand{\xmark}{\color{Maroon}{\ding{55}}}%

\newcommand{\notcheckmark}{\textcolor{black}{\bcmark\kern-1.1ex\raisebox{.7ex}{\rotatebox[origin=c]{125}{--}}}\color{black}}

\definecolor{dark-green}{rgb}{0,0.5,0}

\definecolor{taskblue}{RGB}{66,133,244}
\definecolor{softgray}{RGB}{245,245,248}
\definecolor{headerblue}{RGB}{66,103,178}
\definecolor{metricgreen}{RGB}{52,168,83}
\definecolor{contextorange}{RGB}{251,146,60}
\definecolor{domainpurple}{RGB}{139,92,246}
\definecolor{lightblue}{RGB}{219,234,254}
\definecolor{lightgreen}{RGB}{209,250,229}
\definecolor{lightorange}{RGB}{254,235,200}
\definecolor{lightpurple}{RGB}{237,233,254}
\definecolor{rowgray}{RGB}{248,250,252}
\definecolor{bestcell}{RGB}{220,252,231}

\definecolor{SciColor}{HTML}{4A6FD4}
\definecolor{ForestGreen}{HTML}{228B22}

\definecolor{dccolor}{HTML}{4A6FD4}   
\definecolor{decolor}{HTML}{B35A83}   
\definecolor{ducolor}{HTML}{9B4FCF}   

\definecolor{d2c-header}{HTML}{4472C4}   
\definecolor{d2c-bg}{HTML}{EEF2F6}       
\definecolor{d2e-header}{HTML}{3E7C6E}   
\definecolor{d2e-bg}{HTML}{EDF9F5}       
\definecolor{dqa-header}{HTML}{E45300}   
\definecolor{dqa-bg}{HTML}{FBF3EE}       

\colorlet{basebg}{gray!15}             
\colorlet{ctxbg}{lightblue!35}         
\colorlet{toolbg}{lightorange!40}      
\colorlet{statebg}{lightpurple!35}     
\colorlet{planbg}{lightgreen!40}       

\definecolor{basetxt}{gray}{0.40}                
\definecolor{ctxtxt}{RGB}{41,98,204}             
\definecolor{tooltxt}{RGB}{194,108,22}           
\definecolor{statetxt}{RGB}{119,62,196}          
\definecolor{plantxt}{RGB}{190,60,110}           

\tcbset{
  caseheader/.style={
    enhanced, sharp corners, boxrule=0pt,
    coltext=white, fontupper=\fontfamily{ptm}\selectfont\bfseries\normalsize,
    left=6pt, right=6pt, top=3pt, bottom=3pt,
    before skip=6pt, after skip=0pt,
  },
  casebox/.style={
    enhanced, boxrule=0.4pt, arc=0pt, sharp corners,
    left=4pt, right=4pt, top=4pt, bottom=4pt,
    fontupper=\fontfamily{ptm}\selectfont\small,
    before=\par\noindent,
    after=\par,
  },
}

\newcommand{\caseleftwidth}{0.27\linewidth}
\newcommand{\casecenterwidth}{0.45\linewidth}
\newcommand{\caseshowcasequestion}[1]{%
  {%
    \RaggedRight
    \setlength{\parindent}{0pt}%
    \setlength{\parskip}{0pt}%
    \hangindent=1.55em
    \hangafter=1
    \noindent\makebox[1.55em][l]{\textbf{\footnotesize Q:}}%
    {\fontfamily{ptm}\selectfont\fontsize{8pt}{6.8pt}\selectfont #1}\par
  }%
}
\newcommand{\casegroupheader}[2]{%
  \par
  \begingroup
  \dimen0=\dimexpr\pagegoal-\pagetotal\relax
  \ifdim\dimen0<0.42\textheight
    \newpage
  \fi
  \endgroup
  \begin{tcolorbox}[caseheader, colback=#1]
  #2
  \end{tcolorbox}%
}

\lstdefinestyle{tinycode}{
  language=[LaTeX]TeX,
  basicstyle=\fontfamily{ptm}\selectfont\fontsize{6pt}{7pt}\selectfont,
  keywordstyle=\color{blue},
  commentstyle=\color{gray},
  texcsstyle=*\color{blue!80!black},
  breaklines=true, breakautoindent=true,
  frame=none,
  xleftmargin=0pt, xrightmargin=0pt,
  aboveskip=1pt, belowskip=1pt,
  columns=fullflexible,
  keepspaces=true,
  escapeinside={(*@}{@*)},
}

\newlength\savewidth

\newcolumntype{x}[1]{>{\centering\arraybackslash}p{#1pt}}
\newcolumntype{y}[1]{>{\raggedright\arraybackslash}p{#1pt}}
\newcolumntype{z}[1]{>{\raggedleft\arraybackslash}p{#1pt}}

\definecolor{baselinecolor}{gray}{.9}

\newcolumntype{*}{>{\global\let\currentrowstyle\relax}}
\newcolumntype{^}{>{\currentrowstyle}}

\definecolor{dt}{gray}{0.7}

\usepackage{amsmath}
\usepackage{enumitem}
\usepackage{array}
\usepackage{longtable}
\usepackage{listings}
\usepackage{fvextra}
\fvinlineset{breaklines=true, breakanywhere=true}
\usepackage{ragged2e}
\usepackage{titletoc}

\title{Diagram-MMU: A Multi-Modal Benchmark for \\ Scientific Diagrams}

\author{%
    Weihao Bo\textsuperscript{1,2}\thanks{Equal contribution \quad $\dagger$Corresponding Author\quad$\ddagger$Project Lead}\quad
    Shan Zhang\textsuperscript{3}\footnotemark[2]\quad
    Yanpeng Sun\textsuperscript{4}\footnotemark[1]\quad
    Jie Liu\textsuperscript{2,5}\quad
    Yongke Yao\textsuperscript{2,6}\\
    Jinhao Du\textsuperscript{7}\quad
    Wei He\textsuperscript{2}\quad
    Kai Zou\textsuperscript{8}\quad
    Zechao Li\textsuperscript{1}\footnotemark[2]\quad
    Jingdong Wang\textsuperscript{2$\ddagger$}\\[1em]
    \affiliation[1]{Nanjing University of Science and Technology\quad $^2$Baidu Inc\quad $^3$AIML, Adelaide University \quad $^4$SUTD\quad $^5$Southeast University\quad $^6$East China Normal University\quad $^7$University of Oxford\quad $^8$NetMind.ai\newline}
}

\abstract{  Multimodal Large Language Models (MLLMs) have been growing the capability for scientific writing and collaboration. For example, OpenAI Prism is a free workspace for scientific writing and collaboration. One important feature in Prism is turning scientific diagrams directly into \LaTeX{} TikZ code. In this paper, we build a benchmark, \ourBench{},
a multi-modal  benchmark designed to assess MLLMs' ability for scientific diagram parsing and understanding. \ourBench{} features 3.7k curated diagrams and 18.3k human-validated questions across six domains. 
It evaluates MLLMs on three tasks common in vibe writing workspaces: diagram-to-code parsing, diagram-to-code editing, and diagram question answering, alongside agentic settings per task. The evaluation of 12 MLLMs reveals that diagram-to-code tasks are more challenging than diagram question answering: models can reason well over diagrams but struggle to parse and edit them, underscoring the need for methods to enhance MLLMs' capability in diagram-to-code generation.  Under agentic settings, most models improve parsing and editing performance but degrade on question answering,  while Claude-4.6 Opus consistently improves across all three tasks.  
}

\date{\today}
\connection{\email{shan.zhang@adelaide.edu.au, yanpeng\_sun@sutd.edu.sg}}
\checkdata[Project Page]{\url{https://vi-ocean.github.io/projects/diagram-mmu}}
\abslogo{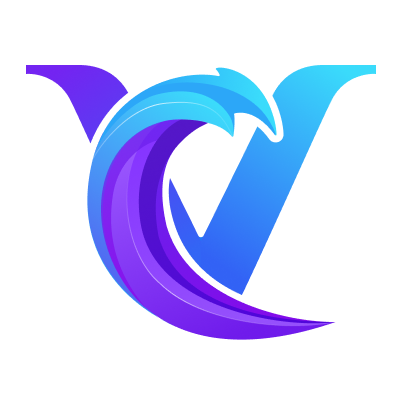}

\begin{document}
\maketitle

\section{Introduction}
\label{sec:intro}

In the wake of significant advancements in large language models (LLMs)~\cite{minimaxm25, kimik2openagentickimiteam2026,glm5team2026glm5vibecodingagentic}, multimodal large language models (MLLMs)~\cite{gemini31pro,qwen3.5,gpt52,seed20pro} have rapidly evolved to extend their versatility across vision--language tasks~\cite{yue2024mmmu,yue2025mmmu,sun2024descriptive,bo2025agentic}. To bring these capabilities into everyday research workflows, 
OpenAI Prism~\cite{prism} provides a free vibe writing workspace where any researcher who writes in \LaTeX{} can draft papers, generate figures, and convert scientific diagrams into TikZ code.

Scientific diagrams are the primary visual medium through which researchers convey experimental results, mathematical relationships, molecular structures, and circuit topologies~\cite{wang2024charxiv,yangchartmimic,sun2025mathblind}.
A researcher working with diagrams needs to:
{(1)}~{parse} a diagram into editable TikZ code so it can be included in a manuscript;
{(2)}~{edit} a diagram--changing colors, labels, data values, or layout--to fit specific requirements; and {(3)}~{describe or reason} about diagram to interpret domain-specific semantics of symbols, or even perform hypothetical reasoning (what-if) about how answers change conditioned on modified diagrams.

\begin{table*}[t]
  \centering
    \caption{Comparison with diagram-to-code and question-answering benchmarks. {\#Diag.}: unique diagrams across diagram types ({\#Diag. Type}). {Tasks} include three fundamental tasks (evaluating how to think): Diagram-to-Code Parsing (D2C-P), Diagram-to-Code Editing (D2C-E), and Diagram Question Answering (DQA), alongside agentic settings (evaluating how to act).  Evaluation spans multi-levels: object-level F1 for fine-grained element grounding, CrystalBLEU for TikZ syntax correctness, image-level for visual appearance similarity, and answer accuracy (Acc.) for DQA. Benchmarks without TikZ Code ({\xmark}) target Python charts, HTML WebUIs, SVG graphics, or no code (ChartE$^3$). \notcheckmark\ is partially supported (limited editing or text-only input). $\oslash$ denotes the synthetic diagrams.}  
  \label{tab:comparison}
  \renewcommand{\arraystretch}{1.}
  \setlength{\tabcolsep}{3.5pt}
  \resizebox{\textwidth}{!}{%
  \begin{tabular}{l ccc cccc cccc}
  \toprule
  & \multicolumn{3}{c}{\textbf{Data}}
  & \multicolumn{4}{c}{\textbf{Tasks}}
  & \multicolumn{4}{c}{\textbf{Evaluation Metrics}} \\
  \cmidrule(lr){2-4} \cmidrule(lr){5-8} \cmidrule(lr){9-12}
  \textbf{Benchmark}
  & \textbf{\makecell{TikZ\\ Code}} & \textbf{\makecell{\#Diag. \\ Type}} & \textbf{\#Diag.}
  & \textbf{D2C-P} & \textbf{D2C-E} & \textbf{DQA} & \textbf{Agentic}
  & \textbf{Object} & \textbf{CBLEU} & \textbf{Image} & \textbf{\makecell{Answer\\Acc.}} \\
  \midrule

  \multicolumn{12}{l}{\textit{Diagram-to-Code Parsing}} \\[2pt]
  ChartMimic $\oslash$~\cite{yangchartmimic}
  & \xmark & 1 & 2.4k
  & \cmark & \notcheckmark & \xmark & \xmark
  & \cmark & \cmark & \cmark & \xmark \\
  InfiBench-V~\cite{jiang2025viscodex}
  & \xmark & 2 & 0.3k
  & \cmark & \xmark & \xmark & \xmark
  & \xmark & \cmark & \cmark & \xmark \\

  Plot2Code~\cite{wu2025plot2code}
  & \xmark & 1 & 0.1k
  & \cmark & \xmark & \xmark & \xmark
  & \xmark & \cmark & \cmark & \xmark \\

  SVG-Diagrams~\cite{rodriguez2025starvector}
  & \xmark & 4 & 0.5k
  & \cmark & \xmark & \xmark & \xmark
  & \xmark & \xmark & \cmark & \xmark \\

  DiagramGenBenchmark~\cite{wei2025words}
  & \cmark & 2 & 0.4k
  & \cmark & \notcheckmark & \xmark & \xmark
  & \xmark & \cmark & \cmark & \xmark \\

  AutomaTikZ~\cite{belouadi2024automatikz}
  & \cmark & 4 & 1.0k
  & \notcheckmark & \xmark & \xmark & \xmark
  & \xmark & \cmark & \cmark & \xmark \\

  DeTikZify~\cite{belouadi2024detikzify}
  & \cmark & 4 & 1.5k
  & \cmark & \xmark & \xmark & \xmark
  & \xmark & \cmark & \cmark & \xmark \\

  Image2Struct~\cite{roberts2024image2struct}
  & \cmark & 4 & 0.3k
  & \cmark & \xmark & \xmark & \xmark
  & \xmark & \xmark & \cmark & \xmark \\

  \midrule
  \multicolumn{12}{l}{\textit{Diagram-to-Code Editing}} \\[2pt]
    \rowcolor{gray!10}
  ChartE$^3$~\cite{li2026charte}
    & \xmark & 1 & 0.8k
    & \xmark & \cmark & \xmark & \xmark
    & \cmark & \xmark & \cmark & \xmark \\
    ChartM$^3$$\oslash$~\cite{yang2025chartm3}
    & \xmark & 1 & 1.0k
    & \xmark & \cmark & \xmark & \xmark
    & \xmark & \cmark & \cmark & \xmark \\

  \midrule
  \multicolumn{12}{l}{\textit{Diagram Question Answering}} \\[2pt]

  CharXiv~\cite{wang2024charxiv}
    & \xmark & 1 & 2.3k
    & \xmark & \xmark & \cmark & \xmark
    & \xmark & \xmark & \xmark & \cmark \\

    ChartQA~\cite{masry2022chartqa}
    & \xmark & 1 & 1.0k
    & \xmark & \xmark & \cmark & \xmark
    & \xmark & \xmark & \xmark & \cmark \\
    SGP-Bench $\oslash$~\cite{qiu2025sgpbench}
    & \xmark & 2 & 3.5k
    & \xmark & \xmark & \notcheckmark & \xmark
    & \xmark & \xmark & \xmark & \cmark \\
    MATHEMETRIC $\oslash$~\cite{sun2025mathblind}
    & \xmark & 3 & 1.2k
    & \xmark & \xmark & \cmark & \xmark
    & \cmark & \xmark & \xmark & \cmark \\

  \midrule\midrule
  \textbf{\ourBench{} (Ours)}
  & \bf \cmark & \bf 6 & \bf 3.7k
  & \cmark & \cmark & \cmark & \cmark
  & \cmark & \cmark & \cmark & \cmark \\

  \bottomrule
  \end{tabular}
  }
  \end{table*}

To assist with the above tasks, MLLMs need strong foundational abilities in perception, coding, domain knowledge, and reasoning (\eg, \emph{how to think}~\cite{seed20pro}). Moreover, in the vibe writing workspace, models also need to know \emph{how to act} (agentic ability~\cite{seed20pro}): searching TikZ for unfamiliar syntax, coding based on the provided visual objects, building on TikZ codes to apply edits, or deciding which of these steps to invoke for task solving. Most existing benchmarks (Table~\ref{tab:comparison}) cover narrow diagram types (predominantly charts) with Python or SVG as the code representation, include a single task (coding, editing, or question answering), and no current benchmark designs agentic settings during evaluation.

\begin{figure}[t]
    \centering
    \includegraphics[width=1\linewidth]{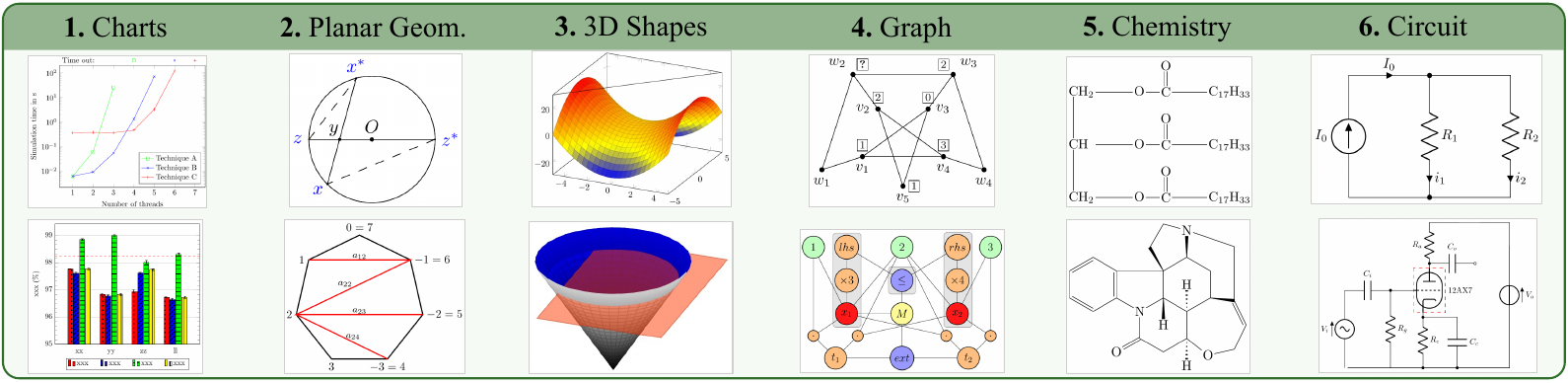}
    \vspace{-1.5em}
    \caption{Demo illustration of the six types of diagrams in \ourBench{}.}
    \label{fig:domain_example}
\end{figure}

We introduce \ourBench{}, a multi-modal benchmark designed to assess MLLMs' ability across diagram-to-code parsing (D2C-P), diagram-to-code editing (D2C-E), and diagram question answering (DQA), alongside agentic evaluation settings. This benchmark comprises a carefully curated dataset sourced from TikZ/\LaTeX{} documentations, with a selection and filtering process to control quality. The final {3,744} unique diagrams paired with {18,305} evaluation samples are cross-validated by 13 graduate students. Specifically, (1)~we adopt TikZ code, which is commonly used in paper writing in \LaTeX{} and integrates directly into Overleaf and Prism, and cover six scientific diagram types including charts, planar geometry, 3D shapes, graphs, chemistry, and circuit diagrams (Fig. ~\ref{fig:domain_example})--for the first time covering chemistry and circuits for diagram-to-code tasks; (2)~for evaluation metrics on diagram-to-code tasks, we use three levels: image-level measuring visual appearance similarity, code-level measuring syntactic correctness, and our object-level F1 scores measuring the basic objects that the code draws; (3)~we design {16} controllable evaluation settings (Table~\ref{tab:context_settings}): 3 for foundational ability and 13 for agentic ability--context utilization, tool use, state management, and planning. Importantly, for tool use we build our TikZ search tool as an MCP server~\cite{anthropic2024mcp}, enabling MLLMs to selectively access relevant references, reducing the noise of web search and the context rot caused by loading full PDF manuals.

Overall, \ourBench{} is the first benchmark to test both foundational and agentic abilities for solving scientific diagram-related tasks. Evaluation on {12} MLLMs (6 closed-source, 6 open-source) reveals the key findings on \emph{foundational ability}: (1) Models can reason well over diagrams (DQA accuracy up to 86\%) but struggle to code them (D2C-P object-level F1 ranges 31--57\%), revealing perception and coding limitations; (2) Models that fail on diagram-to-code parsing also struggle with editing (D2C-E) and answering tasks (DQA); (3) Gemini-3.0 Pro achieves the most balanced profile across three tasks. On the \emph{agentic ability}, (4) agency benefits editing more than parsing, likely because textual editing instructions help guide tool and context use; (5) most models degrade from excessive retrieval or poorly targeted queries in the TikZ search process, while Claude-4.6 Opus has the strongest tool use ability; (6)~planning is the weakest agentic capability, especially on DQA ($-$0.3 to $-$8.8 accuracy degradation). \ourBench{} serves as a pilot evaluation testbed for scientific diagram parsing, editing, and understanding in vibe writing workspaces, and we hope it inspires further works.


\section{Related Works}
\label{sec:rela}
\vspace{-.2em}

Most existing benchmarks focus on charts~\cite{wu2025plot2code,wang2024charxiv,masry2022chartqa,li2026charte,yang2025chartm3} or narrowly cover scientific diagram types~\cite{rodriguez2025starvector,belouadi2024automatikz,belouadi2024detikzify,sun2025mathblind}, with several using synthetic diagrams~\cite{yang2025chartm3,qiu2025sgpbench,sun2025mathblind}. Moreover, benchmarks that adopt Python or SVG as the code representation produce standalone scripts or markup outside the \LaTeX{} ecosystem, limiting direct integration into scientific authoring environments. We compare \ourBench{} with existing diagram benchmarks in Table~\ref{tab:comparison}; extended discussion is provided in Appendix~\S\ref{sec:appx_rela}.

\begin{figure}[b]
    \centering
    \includegraphics[width=1\linewidth]{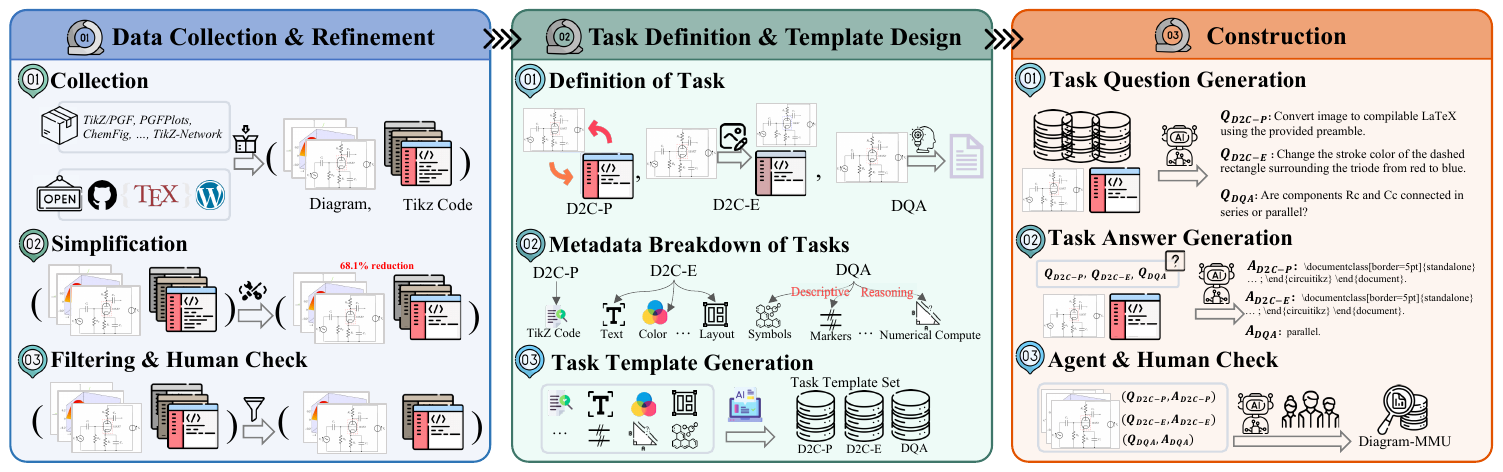}
    \vspace{-1.5em}
    \caption{Task construction pipeline for \ourBench{}.}
    \label{fig:agentic_qa}
    \vspace{-1em}
\end{figure}

\noindent\textbf{Code representation.}
Most coding benchmarks adopt Python, primarily for charts~\cite{yangchartmimic,wu2025plot2code,li2026charte,yang2025chartm3}, or SVG/HTML, for icons, fonts, emojis and web UI~\cite{rodriguez2025starvector,jiang2025viscodex}. Although these languages are common, they are not native to \LaTeX{}: they are rendered by matplotlib/CairoSVG into standalone image files rather than compiled inline within \LaTeX{} authoring environments (Overleaf).  Moreover, those are vector graphics, generating low-level path elements and fail to maintain accurate geometric relations or only generate outputs of limited complexity~\cite{belouadi2024automatikz}.
Due to the expressiveness and diverse packages supporting science in TikZ, \cite{belouadi2024automatikz,belouadi2024detikzify,roberts2024image2struct} build TikZ-based benchmarks, but evaluate only diagram-to-code parsing on limited diagram types. Vibe writing workspaces, \eg, Prism~\cite{prism}, further advance the utilization of TikZ code: even researchers are not familiar with TikZ syntax, vibe writing frameworks can assist them in diagram-to-code parsing, editing, and integrating the output automatically into manuscripts. We thus use TikZ as code representation.  Beyond covering broader diagram types and evaluation metrics, we also build a TikZ search tool as an MCP server to provide TikZ syntax references to advance TikZ integration into vibe writing frameworks.

\noindent\textbf{Task coverage and question types.}
ChartE$^3$~\cite{li2026charte} designs local and global editing templates for charts, and CharXiv~\cite{wang2024charxiv} introduces descriptive and reasoning question types for chart understanding; \ourBench{} extends such designs from charts to all six scientific domains. In \ourBench{}, each diagram is paired with coding, editing, and understanding questions for joint evaluation, and DQA answers are open-ended and graded by an LLM judge, avoiding the selection bias of multiple-choice~\cite{liu2024mmbench,fu2025video} or yes/no formats.



\section{\ourBench{}}
\vspace{-.1em}

\subsection{Data Collection}
\noindent\textbf{Source.}
As shown in the first column of Fig.~\ref{fig:agentic_qa}. We collect the TikZ source code from two primary sources. One source is {official package handbooks}, \textit{i.e.}, the TikZ/PGF manuals: PGFPlots, CircuiTikZ, TKZ-Euclide, ChemFig, and TikZ-Network documentation. 
The other source is {community resources}, including \texttt{texample.net}, 
TeX Stack Exchange, and GitHub \texttt{tikz\_favorites}. 
There are totally 6,849 pieces of 
TikZ code, {3,540} and {3,309} per source. 

\noindent\textbf{Simplification,
filtering,
and categorization.}  
We simplify the TikZ code by removing  redundant and useless codes,
such as preamble packages unrelated to diagrams (\eg, theorem environments).
We check the correctness of
the simplied code
by two ways:
compiling the simplified code
using \texttt{pdflatex}, \texttt{lualatex}, or \texttt{xelatex}
and comparing the rendered images
of the original and simplified code.
We compare the renderd images
by an automatic MLLM model comparison
and a further mannual comparison.
If the compile is not successful
or the rendered images are not the same,
the corresponding code sample is removed.

\begin{table*}[t]
    \caption{Data statistics of \ourBench{}. In Table~\ref{de_stats}, {TikZ.p} indicates TikZ package; basic
      graphical objects are shared across diagrams, while domain-specific objects are summarized in the
      footnote; {\#Q} denotes total questions; {Knowledge} is domain-specific for the DQA task. Figure~\ref{fig:destats} is the metadata breakdown of D2C-E, editing dimensions acorss text, color,
      scope, and layout: scope operates on local transformations and
      layout operates on global transformations (\eg, chart type conversion).}
    \label{tab:data_stats}
    \begin{subfigure}[c]{0.60\textwidth}
      \centering
      \renewcommand{\arraystretch}{1.4}
      \fontsize{8}{20}
      \setlength{\tabcolsep}{0.5pt}
      \resizebox{\linewidth}{!}{%
      \begin{tabular}{l|| cc|| cc}
      \toprule
      & \multicolumn{2}{c||}{\textbf{D2C-P}}
      & \multicolumn{2}{c}{\textbf{DQA}} \\
      \cmidrule(lr){2-3} \cmidrule(lr){4-5}
      \textbf{Domain}
      & \textbf{\#Q} & \textbf{TikZ.p}
      & \textbf{\#Q} & \textbf{Knowledge} \\
      \midrule
      Charts       & 959 & \texttt{pgfplots}         & 1,834 & statistics, trend/extremum analysis \\
      Planar Geom. & 598   & \texttt{tikz/tkz-euclide} & 1,118 & Euclidean notation/formulas \\
      3D Shapes    & 236   & \texttt{pgfplots}         & 455   & solid geometry, cross-sections \\
      Graph        & 1,356 & \texttt{tikz}             & 2,679 & degree, connectivity, paths, tree \\
      Chemistry    & 187   & \texttt{chemfig}          & 366   & bond, atom, functional groups \\
      Circuit      & 403   & \texttt{circuitikz}       & 694   & topology, electrical/power laws \\
      \bottomrule
      \multicolumn{5}{l}{\footnotesize \underline{\textit{Shared objects:}} \texttt{path:open}, \texttt{path:closed}, \texttt{node:rectangle}, \texttt{node:circle}, \texttt{text}, $\cdots$;\;}
   \\
      \multicolumn{5}{l}{\footnotesize \underline{\textit{Domain-specific:}} \texttt{data\_point}/\texttt{axis} (Charts, 3D),\; \texttt{node:vsource}/\texttt{node:resistor} (Circuit).} \\
      \end{tabular}}\vspace{0.1em}\caption{\label{de_stats}}
    \end{subfigure}
    \hfill
    \begin{subfigure}[c]{0.36\textwidth}
      \centering
      \includegraphics[width=1.\linewidth]{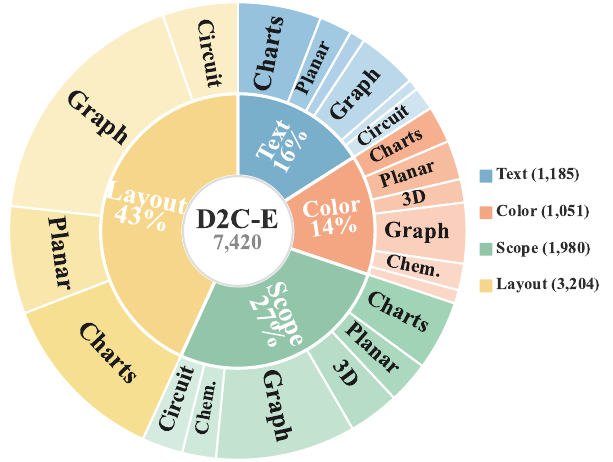}
      \caption{(Updated) \label{fig:destats}}
    \end{subfigure}
    \vspace{-1em}
  \end{table*}

\begin{figure}[t]
    \centering
    \includegraphics[width=1\linewidth]{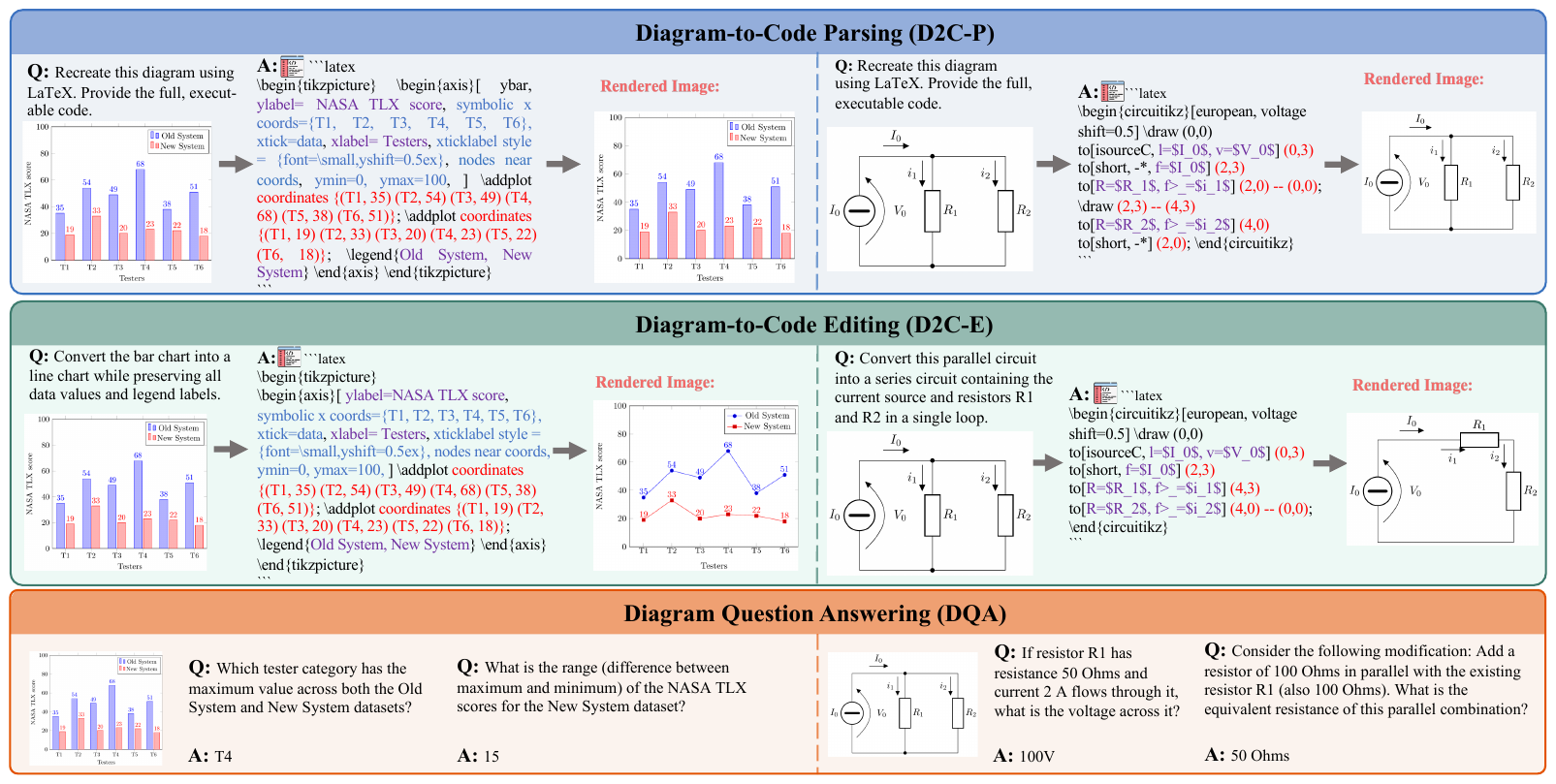}
    \caption{Task overview of \ourBench{}. {Diagram-to-Code Parsing} is to prompt the MLLMs to parse the input diagram into the TikZ code; {Diagram-to-Code Editing} is to prompt the MLLMs to enerate the TiKZ code that can produce an rendered diagram meeting with the modification requirement; and {Diagram Question Answering} is to prompt the MLLMs to answer a question about the input diagram.}
    \label{fig:task_defination}
\end{figure}

\begin{table*}[t]
\centering
\caption{Definition and examples of D2C-Editing Task in \ourBench{}.}
\label{tab:de_tasks}
\vspace{-0.2cm}
\renewcommand{\arraystretch}{1.2}
\setlength{\tabcolsep}{5pt}
\resizebox{\textwidth}{!}{%
\begin{tabular}{l l l l}
\toprule
\textbf{Domain} & \textbf{Editing Dim.} & \textbf{Definition} & \textbf{Example} \\
\midrule

\multirow{4}{*}{Charts} 
& Text   & Modify labels, tick marks, or legend text. 
& Rename label speed to velocity. \\
& Color  & Change fill, stroke, or text color of elements. 
& Change bar color to blue/legend text to red. \\
& Scope  & Add or remove data series or elements. 
& Move the legent to the top-left. \\
& Layout & Modify chart type or filter data. 
& Convert bar chart to line chart/remove values below 10. \\

\midrule

\multirow{4}{*}{Planar Geom.} 
& Text   & Modify point labels or annotations. 
& Change vertices from ABC to DEF. \\
& Color  & Change line or region color. 
& Change the line AB to red. \\
& Scope  & Add, delete, or rotate geometric elements. 
& Add midpoint $M$ on segment AB. \\
& Layout & Add construction lines or auxiliary geometry. 
& Construct the circumcircle of triangle ABC. \\

\midrule
\multirow{4}{*}{3D Shapes}
& Text   & Modify labels of vertices or faces.
& Rename edge AB to MN. \\
& Color  & Change the color of edges or regions.
& Highlight the cross-section polygon in green. \\
& Scope  & Add/delete solid elements.
& Add the altitude from A to plane BCD. \\
& Layout & --
& -- \\

\midrule

\multirow{4}{*}{Graph} 
& Text   & Modify node or edge labels. 
& Rename node A to Start. \\
& Color  & Change node or edge color. 
& Highlight shortest path edges in red. \\
& Scope  & Add or delete nodes or edges. 
& Add an edge between node B and C. \\
& Layout & Rearrange layout or extract subgraph. 
& Re-layout graph using circular layout. \\

\midrule

\multirow{4}{*}{Chemistry} 
& Text   & Modify atom labels, charges, or subscripts. 
& Change CH$_3$ to CH$_2$. \\
& Color  & Change atom or bond color. 
& Highlight oxygen atoms in red. \\
& Scope  & Add or delete bonds or atoms. 
& Add a double bond between C and O. \\
& Layout & --
& -- \\

\midrule

\multirow{4}{*}{Circuit} 
& Text   & Modify component labels or values. 
& Rename the label from V\_in to V\_Source. \\
& Color  & Change wire or component color. 
& Highlight power supply line in red. \\
& Scope  & Swap or reposition components. 
& Move capacitor C1 to the right of R2. \\
& Layout & Modify circuit topology or branch structure.
& Convert a series RLC circuit to a parallel RLC circuit. \\
\bottomrule
\end{tabular}
}
\end{table*}

We then manually remove (near-)duplicate and trivial diagrams (\eg, single arrows, borders, or logos). We use the MLLM,
Qwen3-VL-235B-A22B,
to classify the diagram and the code
into 6 domains:
charts, planar geometry, 3D shapes, graph structures, chemical expressions, and circuit.
Then, we mannually check the classification correctness
and make adjustment if necessary.  Finally, we have 
\textbf{3,744} diagrams. 
Fig.~\ref{fig:domain_example}
presents the examples of the diagrams
of the 6 domains.
The statistics is given in
Table~\ref{tab:data_stats}.


\subsection{Task Definition}
\textbf{Diagram-to-Code Pasing (D2C-P).} The task is to prompt the 
MLLM to  parse the input diagram into 
the TikZ code. An example of the prompt is ``Recreate this diagram using LaTeX. Provide the full, executable code. Start with \Verb|\|\Verb|documentclass[tikz,border=5pt]{standalone}\usepackage{pgfplots}...| ''. It contains instruction conditioned on a provided preamble (document class, required packages, and libraries). 
The top row of Fig.~\ref{fig:task_defination} illustrates this task.

\noindent\textbf{Diagram-to-Code Editing (D2C-E).} This task 
is to prompt the MLLM 
to generate
the TiKZ code
that can produce the rendered
diagram 
meeting with the modification requirement.
An example of the prompt is ``Convert the diagram in the image into a complete, executable LaTeX code block, applying the following modification: convert the line chart to a scatter plot by removing the lines connecting the data markers. Start with \Verb|\|\Verb|documentclass[tikz,border=5pt]{standalone}\usepackage{pgfplots}...|''
The second row of Fig.~\ref{fig:task_defination} gives an example.

\noindent\textbf{Diagram question answering (DQA).}
This task is to prompt the MLLM
to answer a question
about the input diagram.
An example of the prompt is
``Which tester category has the maximum value across both the OldSystem and New System datasets?''
and the expected answer is ``T4''. The third row of Fig.~\ref{fig:task_defination} illustrates this task. 


\subsection{Task Construction}
\noindent\textbf{Overview.} For D2C-P, we use fixed task templates with placeholders instantiated directly from the original diagrams (the first row of Fig.~\ref{fig:task_defination}). For D2C-E and DQA task generation, we design an agentic pipeline, as shown in the third column of Fig.~\ref{fig:agentic_qa}. Given diagrams, TikZ source codes, and task-specific templates as inputs, an agent generates corresponding tasks. 

The generation pipeline for D2C-E includes: (1) Template predefinition
for each domain;
(2) Given diagram,
agent select two tasks
from the templates of the corresponding domain;
(3) Generate the answer
from models;
(4) Agent-check and modification;
(5) Human-check and modification.
The DQA task generation pipeline is similar to that of D2C-E; only task templates differ.  

We provide the detailed agentic pipeline in Appendix~\S\ref{sec:appx_pipeline}, and all templates in Appendix~\S\ref{sec:appx_d2ce_templates} \& ~\S\ref{sec:appx_dqa_templates}. To  guarantee quality, we conduct a rigorous manual review: all samples are manually reviewed by 13 graduate students, where each annotator's assigned samples are cross-validated by another annotator to ensure that the language is clear and unambiguous, and that the question is answerable with logically consistent options and a correct designated answer.

\noindent\textbf{Question types.} 
\label{sec:quesion_define}
D2C-P has one standard question type following its task definition: parsing the diagram into the corresponding TikZ code. D2C-E also follows a standard editing format: prompting models to generate the corresponding TikZ code conditioned on the changes, but we cover four editing dimensions, \eg, text, color, scope, and layout. Example illustrations are shown in Table~\ref{tab:de_tasks}.


We construct two types of questions in DQA: {descriptive} and {reasoning}, with a total of 60 templates across the 6 domains (see Appendix~\S\ref{sec:appx_dqa_templates} for details). {Descriptive questions} assess the model's capability in extracting and aggregating basic information from diagrams, where the correct answer must strictly adhere to domain-specific knowledge, \eg, two nodes connected by a line represent a geometric segment in planar geometry but a single bond between atoms in chemistry. {Reasoning questions} evaluate the model's ability to perform numerical computation, where the correct answer requires applying domain-specific formulas and laws, \eg, Ohm's law for circuit analysis, Euclidean formulas for geometric area computation, or solid geometry formulas for polyhedron volume.

\emph{Descriptive questions}, constructed from 23 templates, require:  {(1)}~identifying symbols and markers per domain (\eg, zigzag symbols as resistors in circuits, bond types and functional groups in chemistry, right-angle squares in geometry);  {(2)}~aggregating diagram information to count elements (\eg, vertices in geometry, ticks/legends in charts, atoms/functional groups in chemistry) and to recognize typology and data patterns (\eg, node degree in graphs, increase/decrease trends in charts). See Appendix~\S\ref{sec:appx_dqa_descriptive_examples} for data illustrations.

\emph{Reasoning questions} include standard questions (formed by 18 templates) and {what-if} questions (formed by 19 templates).  The {what-if} questions require the model to predict answer conditioned on a specific element modification, \eg, ``if the data value changes to X, what is the mean of the data series?'' All questions require: {(1)}~numerical computation by applying domain-specific formulas (\eg, min, max, and median in charts; area and perimeter in geometry); {(2)}~visual reasoning over diagram (\eg, path length and height of tree in graphs; molecular properties from bond structure in chemistry). Our diagram QAs require only domain semantics/laws and numerical computation, rather than advanced expert-level knowledge~\cite{lu2024mathvista,zhang2024mathverse} (see Appendix~\S\ref{sec:appx_dqa_reasoning_examples} for per-domain examples). 





\subsection{Evaluation Metrics}
\label{eval_primitive}

\noindent\textbf{Diagram-to-Code Parsing and Editing.}
We evaluation the diagram-to-code parsing and editing tasks
from the three aspects:
image,
code,
and object.

\noindent\emph{Image-based.} 
We compile both generated and ground-truth TikZ code into images and compare them with four classical metrics, following papers~\cite{roberts2024image2struct, li2026charte}:

\noindent {(1)} SSIM~\cite{wang2004ssim} compares brightness, contrast, structure patterns (higher is better);

\noindent{(2)} CLIP Score~\cite{radford2021learning} measures semantic similarity between two images in a shared vision-language space (higher is better);

\noindent{(3)} LPIPS~\cite{zhang2018unreasonable} uses deep features from a VGG network to measure perceptual distance in a way that matches human judgement (lower is better);

\noindent{(4)} FID~\cite{heusel2017gans} compares the distribution of generated and original images using inception network features (lower is better). 

Together, these metrics assess whether the rendered output looks visually similar to the input diagram.

\noindent\emph{Code-based.}
We use CrystalBLEU~\cite{eghbali2022crystalbleu} to measure the similarity between generated and ground-truth TikZ code, as in~\cite{belouadi2024automatikz}. This focuses on how well the model reproduces the correct coding syntax and structure.

\noindent\emph{Object-based.} We compile both generated and ground-truth TikZ code into DVI and convert them to SVG, from which we extract a Semantic Object Model (SOM)---a structured list of typed, attributed elements (\eg, nodes, paths, text labels, data series). We then organize them into type, text, color, and BBox (see Appendix~\S\ref{sec:appx_som} for detailed extraction process). Finally, we compute four F1 scores between the generated and ground-truth SOMs, as follows:

\noindent{(1)} ${F1}_{{type}}$ measures whether the generated diagram contains the correct types of elements (\eg, node:circle and node:rectangle);

\noindent{(2)} $\mathrm{F1}_{{text}}$ evaluates exact string matching of text labels and numeric values;

\noindent{(3)} ${F1}_{{color}}$ compares element colors via permutation-based assignment using CIEDE2000 perceptual distance~\cite{yangchartmimic};

\noindent{(4)} ${F1}_{{bbox}}$ assesses spatial positioning accuracy via Intersection-over-Union (IoU $\geq 0.3$) between element bounding boxes.

Together, these four dimensions evaluate whether the model faithfully reconstructs the semantic structure, content and layout of diagram objects. The formal formulation of F1 score per dimension is provided in Appendix~\S\ref{sec:appx_f1_formular}. ${F1}_{{avg}}$ is the average across the four dimensions: $\tfrac{1}{4}\left(\mathrm{F1}_{{type}} + \mathrm{F1}_{{text}} + \mathrm{F1}_{{color}} + \mathrm{F1}_{{bbox}}\right)$.



For the editing task, we further split the code- and object-based metrics into two parts: {preserve-only}, computed over elements that should stay the same, and {edit-only}, computed over elements that the instruction asks to change.


\begin{table}[t]
\centering
\definecolor{dccolor}{HTML}{4A6FD4}
\definecolor{decolor}{HTML}{B35A83}
\definecolor{ducolor}{HTML}{9B4FCF}
\caption{Evaluation settings on \ourBench{}. TikZ search tool is built as an MCP server and prompts are abbreviated (full versions in Appendix~\S\ref{sec:appx_prompts}).}
\vspace{-0.5em}
\label{tab:context_settings}
\renewcommand{\arraystretch}{1.15}
\setlength{\tabcolsep}{4pt}
\resizebox{\textwidth}{!}{%
\begin{tabular}{cl l l}
\toprule
\textbf{ID} & \textbf{Settings} & \textbf{Abbreviated Prompts} & \textbf{Capability} \\
\midrule
\rowcolor{basetxt!12}
{\color{basetxt}\textbf{S1}} & Diagram-to-code parsing & Direct coding & Foundational \\
{\color{ctxtxt}\textbf{S2}} & + Objects & Use the provided perception data for coding & Agentic \\
{\color{tooltxt}\textbf{S3}} & + TikZ search tool & Use search tool for TikZ syntax and examples when necessary & Agentic \\
{\color{statetxt}\textbf{S4}} & + Objects (model-gen.) & First perceive basic obejcts in diargam, then  generate code & Agentic \\
{\color{plantxt}\textbf{S5}} & + S2\&S3 & Plan where to call tool in synergy with perception data & Agentic \\
\midrule
\rowcolor{basetxt!12}
{\color{basetxt}\textbf{S6}} & Diagram-to-code editing & Direct editing &  Foundational  \\
{\color{ctxtxt}\textbf{S7}} & + Objects & Use the provided perception data for editing & Agentic  \\
{\color{tooltxt}\textbf{S8}} & + TikZ search tool & Use search tool for TikZ syntax and examples when necessary & Agentic  \\
{\color{statetxt}\textbf{S9}} & + TikZ codes (required) & First generate TikZ code of diagram, then edit code & Agentic  \\
{\color{plantxt}\textbf{S10}} & + TikZ codes (optional) & Plan which diagrams require TikZ code generation for editing &Agentic  \\
{\color{plantxt}\textbf{S11}} & + S7\&S9 & Plan which diagrams require TikZ code and where to call tool for editing & Agentic  \\
\midrule
\rowcolor{basetxt!12}
{\color{basetxt}\textbf{S12}} & Diagram question answering & Direct answering &  Foundational  \\
{\color{ctxtxt}\textbf{S13}} & + Objects & Use the provided perception data for answering & Agentic  \\
{\color{statetxt}\textbf{S14}} & + TikZ codes (required) & First generate TikZ code of diagram, then answer question & Agentic \\
{\color{plantxt}\textbf{S15}} & + TikZ codes (optional)  & Plan which diagrams require TikZ code generation for answering & Agentic \\
{\color{plantxt}\textbf{S16}} & + TikZ search tool\&S14 & Plan  which diagrams require TikZ code and where to call tool for answering & Agentic \\
\bottomrule
\end{tabular}
}\vspace{-0.5em}
\end{table}

\noindent\textbf{Diagram Question Answering.} 
Following~\cite{zhou2026worldvqameasuringatomicworld}, we evaluate diagram question answering using accuracy. We use Qwen3-Next-80B-A3B-Instruct to extract the answer and assign binary scores (0 for incorrect, 1 for correct). The grading prompt is provided in Appendix~\S\ref{sec:appx_dqa_grading}.

\subsection{Evaluation Scheme} 
\label{sec:eval_set}
In real scientific writing workflows, MLLMs working with diagrams need foundational abilities, \eg, perception, coding, domain knowledge, and reasoning, but also the ability to know when and how to act ($\textit{a.k.a}$ agentic capability): they parsing a diagram may first consult TikZ documentation for unfamiliar syntax (tool use), leverage perceptual objects in the diagram to write TikZ code (context utilization), build on generated diagram code to edit (state management), or decide which of these steps to invoke or in what order (planning).  \ourBench{} is, to our knowledge, the first benchmark to provide 16 flexible control settings evaluating both foundational and agentic capability of current MLLMs (Table~\ref{tab:context_settings}). Specifically, the four levels of agentic ability are defined as: {(1)}~{Context utilization}: whether models can leverage task-relevant information from context; {(2)}~{Tool use}: knowing when to invoke tools, what to query, and how to incorporate the results; {(3)}~{State management}: whether models can build on prior outputs incrementally toward the final goal; {(4)}~{Planning}:  deciding which ability is needed and how to combine them for task solving~\cite{yao2022react, buyya2026agentic,hartung2025ai,tang2025human}.

For tool use evaluation, we build our TikZ search tool as an MCP server. Na\"ive approaches such as web search or loading full PDF manuals introduce noise: browsers may return forum threads and advertisements, while complete manuals (\eg, the PGFPlots reference spans ${\sim}$560 pages) cause context rot. We use Mintlify\footnote{\url{https://mintlify.com}} to generate an MCP server from curated TikZ documentation, enabling selective access to relevant references instead of ingesting entire documents.  MCP serves as a unified tool interface, allowing any MLLM to query syntax references without model-specific implementations (details in Appendix~\S\ref{sec:appx_mcp}). 

\begin{figure*}
    \centering
    \includegraphics[width=\linewidth]{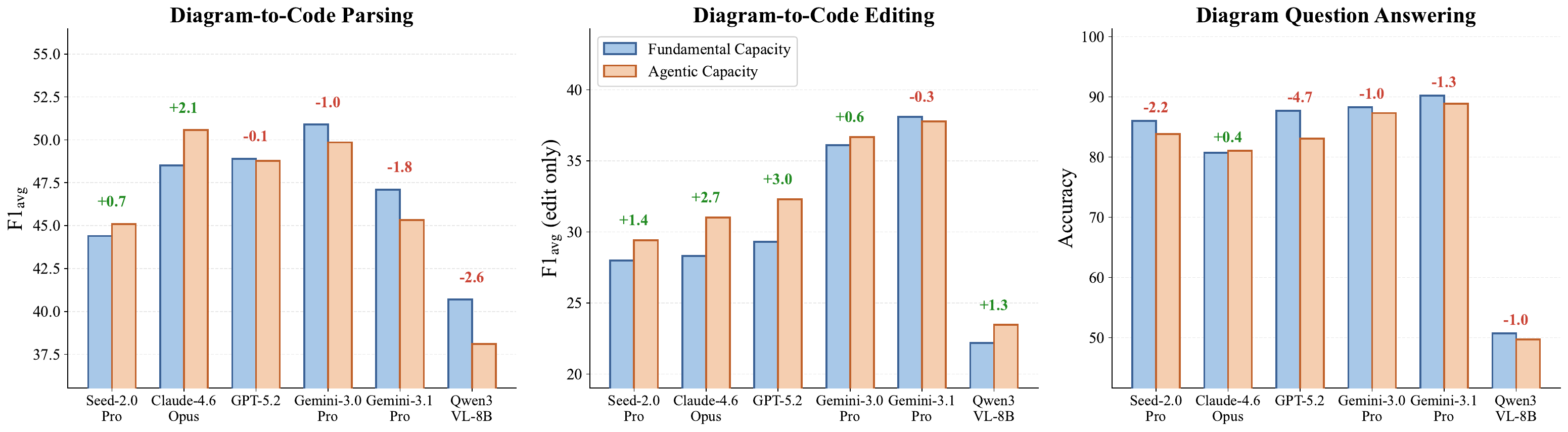}
    \caption{Foundational \textit{vs.} agentic performance across three tasks. Green/red annotations indicate gains/drops from foundational to agentic. Current models reason well over diagrams but struggle with visual perception and coding under foundational evaluation. Agency improves editing more than parsing, likely because textual instructions guide tool and context use, but nearly all models degrade on DQA. }  
    \label{fig:found_vs_agent}
\end{figure*}

We do not claim \ourBench{} evaluates the full spectrum of agentic ability; as a pilot benchmark for evaluating agentic ability in vibe writing \wrt diagram-to-code and understanding tasks, our coverage targets only certain aspects. We discuss limitations in Appendix~\S\ref{sec:appx_limitations} and hope to inspire future efforts to advance more thorough agentic evaluation dimensions in the vibe writing workspaces.

\noindent\textbf{Results.} Fig.~\ref{fig:found_vs_agent} reports F1$_{{avg}}$ for D2C-P, F1$_{{avg}}$ (edit-only) for  D2C-E, and accuracy for DQA (see \S\ref{sec:experiments} for detailed results and analysis.)

\emph{Foundational capacity}: current models show stronger capability on reasoning but weaker ability on perception and coding, revealing an interesting asymmetry: they excel in knowledge-intensive evaluations but still struggle to link visual objects to coding skills. In short, models can reason deeply over diagrams but fail to code over them--a visual perception weakness aligned with the findings of BabyVision~\cite{chen2026babyvision}. Overall, Gemini-3.0 Pro achieves the most balanced profile.

\emph{Agentic capacity}: Across the coding tasks, closed-source models show stronger agency on D2C-E than D2C-P. For example, GPT-5.2 drops -0.1 on D2C-P but gains +3.0 on D2C-E; we hypothesize that the textual editing  instructions help guide models to know when to use tools and what context is useful for target editing. For DQA, unlike direct answering (input diagram $\to$ answer), in the agentic setting the model plans to invoke diagram-to-code parsing when needed (input diagram $\to$ TikZ code $\to$ answer) and decides what to query in the TikZ search tool. Except for Claude-4.6 Opus (+0.4), all models degrade on DQA, exposing weak planning ability for complex tasks, but more critically, revealing that models cannot effectively map codes to high-level reasoning.

\section{Experiments}
\label{sec:experiments}

\noindent\textbf{Evaluated Models.}
We evaluate 12 closed-source and open-source MLLMs on \ourBench{}.
Closed-source (6): Gemini-3.1 Pro~\cite{gemini31pro}, Gemini-3.0 Pro~\cite{gemini3pro}, Gemini-3.0 Flash~\cite{gemini3flash}, GPT-5.2~\cite{gpt52}, Claude-4.6 Opus~\cite{claude46opus}, and Seed-2.0 Pro~\cite{seed20pro};
Open-source (6): five general-purpose models, \eg, Qwen3.5-397B-A17B~\cite{qwen3.5}, Qwen3-VL-235B-A22B~\cite{qwen3vltechnicalreportbai2025}, Kimi-K2.5~\cite{kimi25}, Qwen3-VL-8B~\cite{qwen3vltechnicalreportbai2025}, and InternVL3-38B~\cite{zhu2025internvl3exploringadvancedtraining}, and one specialist TikZero+\,10B~\cite{belouadi2025tikzero} which is fine-tuned on 456K diagram--TikZ code pairs.
TikZero+\,10B is not trained on editing or question answering samples, so we evaluate it only on the diagram-to-code parsing task.
Generation configurations of models are provided in Table~\ref{tab:appx_model_config}.

\subsection{Main Results}
\label{sec:main_results}
\begin{table*}[t]
\centering
\caption{Main results on foundational capability. Object-level metric is F1$_{\text{avg}}$ (\%), with preserve-only (p) and edit-only (e) split for diagram-to-code editing. Code-level metric is  CrystalBLEU (\%). Image-level is measured by the average of SSIM and CLIP (SC, \%) and the average of FID and LPIPS (FL, \%); $\downarrow$ means lower is better. \textit{All} averages F1$_{\text{avg}}$,   
  CBLEU, and SC (higher is better). Diagram question answering (DQA) uses accuracy (Acc., \%). \textbf{Bold} is best in class; \underline{underline} is second best.}  
\label{tab:main_results}
\renewcommand{\arraystretch}{1.15}
\setlength{\tabcolsep}{4pt}
\resizebox{\textwidth}{!}{%
\begin{tabular}{l|ccccc|ccccccc|c}
\toprule
\multirow{2}{*}{\textbf{Model}}
& \multicolumn{5}{c|}{\textbf{Diagram-to-Code Parsing (D2C-P)}}
& \multicolumn{7}{c|}{\textbf{Diagram-to-Code Editing (D2C-E)}}
& \textbf{DQA} \\
\cmidrule(lr){2-6} \cmidrule(lr){7-13} \cmidrule(lr){14-14}
& \cellcolor{gray!15}\textit{All}
& {F1$_{\text{avg}}$}
& {CBLEU}
& \multicolumn{2}{c|}{(SC\;/\;FL${\downarrow}$)}
& \cellcolor{gray!15}\textit{All}
& \multicolumn{2}{c}{F1$_{\text{avg}}$ (p\,/\,e)}
& \multicolumn{2}{c}{CBLEU (p\,/\,e)}
& \multicolumn{2}{c|}{(SC\;/\;FL${\downarrow}$)}
& {Acc.} \\
\midrule
\multicolumn{14}{c}{\textbf{Close-Source Multimodal Large Language Models}} \\
\midrule
Gemini-3.1 Pro & \cellcolor{gray!15}48.38 & 49.94 & \textbf{32.88} & (62.33 & \textbf{4.05}) & \cellcolor{gray!15}\underline{51.22} & (60.60 & \textbf{40.84}) & (42.02 & \textbf{2.98}) & (66.47 & \textbf{2.52})  & \underline{86.29} \\
Gemini-3.0 Pro & \cellcolor{gray!15}\textbf{53.28} & \textbf{54.64} & \underline{32.76} & (72.44 & \underline{4.61}) & \cellcolor{gray!15}\textbf{53.60} & (\textbf{65.12} & \underline{40.24}) & (\textbf{44.77} & \underline{2.34}) & (\textbf{72.38} & 3.12)  & \textbf{86.46} \\
Gemini-3.0 Flash & \cellcolor{gray!15}50.97 & 51.19 & 30.67 & (71.04 & 4.79) & \cellcolor{gray!15}49.01 & (59.29 & 37.24) & (42.21 & 2.17) & (67.44 & \underline{2.98})  & 84.07 \\
GPT-5.2 & \cellcolor{gray!15}51.88 & 51.35 & 28.81 & (\textbf{75.48} & 6.56) & \cellcolor{gray!15}46.15 & (55.59 & 31.15) & (38.75 & 1.15) & (65.33 & 4.55)  & 81.67 \\
Claude-4.6 Opus & \cellcolor{gray!15}\underline{52.41} & \underline{52.23} & 31.27 & (\underline{73.73} & 7.11) & \cellcolor{gray!15}51.18 & (\underline{62.56} & 31.95) & (\underline{43.59} & 1.42) & (\underline{71.78} & 4.61)  & 68.75 \\
Seed-2.0 Pro & \cellcolor{gray!15}50.89 & 47.57 & 32.56 & (72.54 & 7.55) & \cellcolor{gray!15}43.73 & (53.03 & 30.14) & (37.49 & 1.41) & (61.86 & 5.34)  & 75.90 \\
\midrule
\multicolumn{14}{c}{\textbf{Open-Source Multimodal Large Language Models}} \\
\midrule
Qwen3.5-397B-A17B & \cellcolor{gray!15}52.72 & 51.38 & 32.92 & (73.84 & \underline{5.73}) & \cellcolor{gray!15}50.82 & (62.75 & \underline{36.35}) & (\textbf{43.66} & \underline{1.91}) & (\textbf{70.43} & \underline{3.76})  & \textbf{83.42} \\
Qwen3-VL-235B-A22B & \cellcolor{gray!15}\underline{55.98} & \underline{55.11} & \textbf{37.92} & (\underline{74.90} & 6.69) & \cellcolor{gray!15}\textbf{52.25} & (\textbf{63.79} & 33.38) & (\underline{43.05} & 1.85) & (\underline{70.07} & 4.57)  & 63.60 \\
Kimi-K2.5 & \cellcolor{gray!15}\textbf{56.97} & \textbf{57.48} & \underline{36.13} & (\textbf{77.31} & \textbf{5.20}) & \cellcolor{gray!15}\underline{52.13} & (\underline{63.12} & \textbf{36.52}) & (42.11 & \textbf{2.02}) & (69.88 & \textbf{3.17})  & \underline{79.74} \\
Qwen3-VL-8B & \cellcolor{gray!15}48.26 & 44.66 & 33.31 & (66.81 & 8.23) & \cellcolor{gray!15}19.00 & (21.26 & 9.95) & (16.29 & 0.51) & (26.94 & 9.50)  & 47.12 \\
InternVL3-38B & \cellcolor{gray!15}41.13 & 31.47 & 30.91 & (61.02 & 10.14) & \cellcolor{gray!15}38.85 & (44.72 & 22.69) & (33.13 & 1.29) & (57.95 & 7.02)  & 56.39 \\
TikZero+ 10B & \cellcolor{gray!15}25.30 & 15.43 & 17.19 & (43.29 & 12.67) & \multicolumn{7}{c|}{--}  & -- \\
\bottomrule
\end{tabular}}
\end{table*}

\begin{figure}
    \centering
    \includegraphics[width=\linewidth]{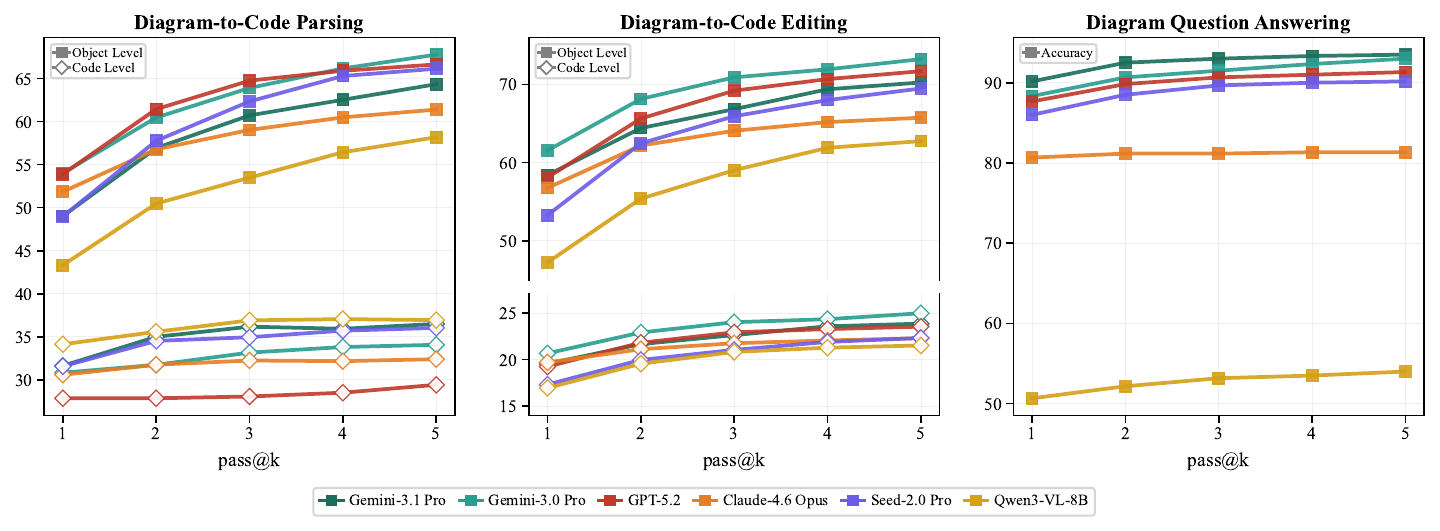} \vspace{-2em}
    \caption{Pass@k evaluation across three tasks: performance improves with $k$, with gains peaking from pass@1 to pass@2 and saturating beyond $k=4$.}
    \label{fig:passk}
    \vspace{-2.em}
\end{figure}

\noindent\textbf{Fundamental Capacity.} Table~\ref{tab:main_results} presents the main results of fundamental capabilities under pass@1 evaluation (average over 6 types of diagrams; per-type results in Appendix~\S\ref{sec:appx_per_domain}. Pass@$k$ evaluation across three tasks is shown in Fig.~\ref{fig:passk}, where models improve with increasing $k$, with the largest relative gain from pass@1 to pass@2, and saturate when $k$ exceeds 4.
Table~\ref{tab:main_results} reports F1$_{\text{avg}}$ for D2C-P and D2C-E (per object F1 score in Fig.~\ref{fig:appx_obj_radar} and more analysis in Appendix \S~\ref{sec:appx_object_f1}).

\noindent\textbf{(1) A larger performance gap among open-source models in object perception.}
D2C-P requires models to perceive basic objects and integrate them into code. Image-based metrics measure visual appearance similarity; all general-purpose models score highest here, and the gap between closed- and open-source models is small (\eg, closed-source models achieve 62.33--75.48 and open-source models 61.02--77.31).
At coding syntax and structure evaluation (CrystalBLEU), generic models score lowest, with a similarly small gap (closed-source: 28.81--32.88; open-source: 30.91--37.92). However, a larger gap emerges at object perception: closed-source F1$_{\text{avg}}$ clusters within 47.57--54.64, while open-source spans 31.47--57.48, and the specialist TikZero+ trails far behind at only 15.43.
This means a generated diagram may appear visually plausible at image level yet contain incorrect identification at object level.

\noindent\textbf{(2) Models follow editing instructions but struggle to produce matching code. }
In the editing task, models reach only 0.51--2.98 CrystalBLEU on the edit partitions, yet achieve relatively higher object F1 scores, indicating they may generate the correct objects in line with the editing instructions but fail to place them at the correct positions or formats to match the ground truth code.

\begin{figure}
    \centering
    \includegraphics[width=\linewidth]{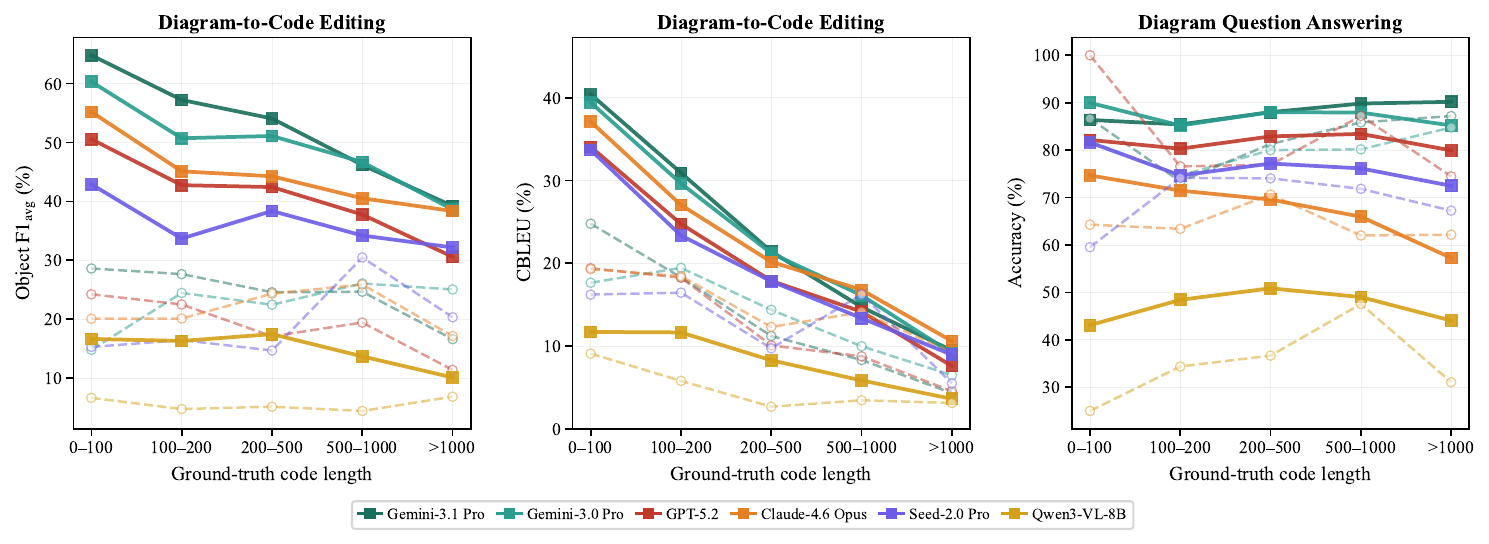}\vspace{-0.5em}
    \caption{(Updated) Model performance on D2C-E/DQA across two diagram sets split by D2C-P output: successful rendering (solid lines) \textit{vs.} failed rendering (dashed lines). D2C-E is evaluated by object F1$_{\text{avg}}$ (left) and CrystalBLEU (middle), and DQA by accuracy (right). The $x$-axis denotes ground-truth code length. Models that fail D2C-P also score lower on D2C-E and DQA.}
    \label{fig:split_d2cp}
    \vspace{-.1em}
\end{figure}

\noindent\textbf{(3) Models failing on D2C-P also struggle with D2C-E and DU.} 
We split diagrams into two sets conditioned on whether D2C-P produces executable TikZ code (Fig.~\ref{fig:split_d2cp}). On diagrams where D2C-P succeeds, models consistently achieve higher D2C-E scores than on D2C-P failures: the object F1$_{\text{avg}}$ (average of preserve-only\&edit-only) gap between the two sets reaches 20--40 points across most models and code lengths. The similar pattern holds at code-level metrics. 
Even models can produce executable TikZ code on D2C-P failed-rendering diagrams during editing; this does not yield accurate editing. We hypothesize that such executable TikZ results from textual editing instructions, which guide code generation, \eg, leveraging the model's stronger text-to-code ability~\cite{kimik2openagentickimiteam2026,glm5team2026glm5vibecodingagentic}.

\begin{figure}[!tbp]
\centering
\includegraphics[width=0.96\linewidth]{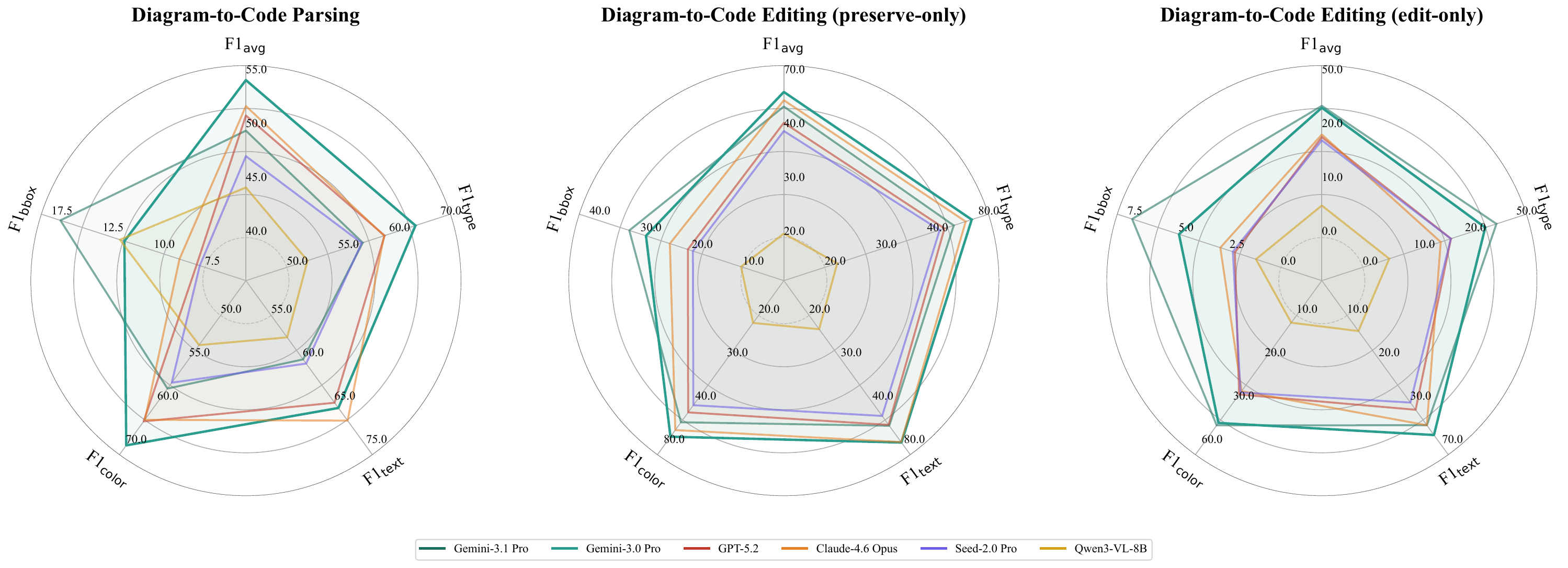}
\caption{(Updated) Breakdown of F1 scores. Each panel shows five axes (avg, type, bbox, color, text), each with its own scale. F1$_{\text{bbox}}$ drops sharply in Diagram-to-Code Parsing, especially for GPT-5.2, Claude-4.6~Opus, and Seed-2.0~Pro.}
\label{fig:appx_obj_radar}
\end{figure}

\noindent\textbf{(4) Models struggle with fine-grained spatial grounding.}
All six models achieve relatively high performance on text extraction and color recognition (Fig.~\ref{fig:appx_obj_radar}). However, spatial grounding (bbox) is consistently the weakest perception dimension. While models can recognize what elements exist, they struggle to encode where they are located.
This deficit is sharpest for GPT-5.2 and Claude-4.6~Opus, where both score 62--71 on type, text, and color yet only 8.0 and 9.2 on bbox. Editing instructions in D2C-E compensate for this weakness only on preserved elements, not on newly edited ones: Claude-4.6~Opus's F1$_{\text{bbox}}$ rises from 9.2 to 24.3 on preserve, and GPT-5.2 from 8.0 to 20.8; however, edit-only F1$_{\text{bbox}}$ drops to 3.0 and 2.3, respectively, since placing newly introduced elements is harder than retaining existing ones. Qwen3-VL-8B, however, falls behind all models across every perception dimension.

\noindent\textbf{(5)  Models struggle most with 3D shapes across all three tasks.}
3D shapes pose the greatest challenge for both closed- and open-source models across all three tasks, likely due to the scarcity of TikZ-3D training samples in pretraining corpora. For D2C-P and D2C-E, charts are the second most challenging domain after 3D, as they contain a higher density of small-scale elements (\eg, tick labels, data markers). Model-specific DQA weaknesses diverge beyond 3D: Seed-2.0 Pro drops notably on charts; Claude-4.6 Opus underperforms on graph structures; and Gemini-3.0 Pro shows relatively lower accuracy on circuits (see Table~\ref{tab:appx_dqa_domain} in Appendix~\S\ref{sec:appx_dqa_domain}).

\noindent\textbf{Agentic Capacity.}
\label{sec:agentic_eval}
We select a \textbf{mini} split of 300 diagrams (50 per domain) with 1,500 QA instances (300/600/600 for D2C-P/D2C-E/DQA), evaluated on 6 representative models: Seed-2.0 Pro, Claude-4.6 Opus, GPT-5.2, Gemini-3.0 Pro, Gemini-3.1 Pro, and Qwen3-VL-8B.

\begin{table*}[t]
\centering
\caption{Main results on agentic evaluation settings across D2C-P (S1--S5) and DQA (S12--S16). {\color{basetxt}S1/S12} uses {\color{basetxt}direct coding/answering}; {\color{ctxtxt}\textbf{Context utilization}: S2/S13} provide diagram objects as context; {\color{tooltxt}\textbf{Tool use}: S3} provides TikZ search tool;
{\color{statetxt}\textbf{State management}: S4} perceives objects first, then builds upon them to code; {\color{statetxt}S14} requires answering building upon the coding states;
{\color{plantxt}\textbf{Planning}: S5} coordinates objects and tool; {\color{plantxt}S15} optionally generates code before answering; {\color{plantxt}S16} adds tool access to S15 (Table~\ref{tab:context_settings}). $\Delta$ (\textcolor{ForestGreen}{$\uparrow$}\,gain, $\downarrow$\,drop) over direct coding/answering {\color{basetxt}S1/S12}.
}
\label{tab:ablation_d2c_dqa}
\vspace{-0.1cm}
\small
\renewcommand{\arraystretch}{1.2}
\setlength{\tabcolsep}{3pt}
\resizebox{\textwidth}{!}{%
\begin{tabular}{l|cc|cc|cc|cc|cc||c|c|c|c|c}
\toprule
\multirow{2}{*}{\textbf{Model}}
& \multicolumn{2}{c|}{\textcolor{basetxt}{\textbf{S1}}}
& \multicolumn{2}{c|}{\textcolor{ctxtxt}{\textbf{S2}}}
& \multicolumn{2}{c|}{\textcolor{tooltxt}{\textbf{S3}}}
& \multicolumn{2}{c|}{\textcolor{statetxt}{\textbf{S4}}}
& \multicolumn{2}{c||}{\textcolor{plantxt}{\textbf{S5}}}
& \textcolor{basetxt}{\textbf{S12}} & \textcolor{ctxtxt}{\textbf{S13}} & \textcolor{statetxt}{\textbf{S14}} & \textcolor{plantxt}{\textbf{S15}} & \textcolor{plantxt}{\textbf{S16}} \\
\cmidrule(lr){2-3}\cmidrule(lr){4-5}\cmidrule(lr){6-7}\cmidrule(lr){8-9}\cmidrule(lr){10-11}\cmidrule(lr){12-16}
& {F1$_{\text{avg}}$} & CBLEU & {F1$_{\text{avg}}$} & CBLEU & {F1$_{\text{avg}}$} & CBLEU & {F1$_{\text{avg}}$} & CBLEU & {F1$_{\text{avg}}$} & CBLEU
& Acc. & Acc. & Acc. & Acc. & Acc. \\
\midrule
Seed-2.0 Pro    & 44.4 & 31.6 & {\color{ForestGreen}$\uparrow$0.2}  & $\downarrow$0.2  & {\color{ForestGreen}$\uparrow$1.2}  & {\color{ForestGreen}$\uparrow$1.4}  & $\downarrow$1.2   & $\downarrow$1.2  & {\color{ForestGreen}$\uparrow$2.6}  & {\color{ForestGreen}$\uparrow$1.5}  & 86.0 & {\color{ForestGreen}$\uparrow$1.0}  & {\color{ForestGreen}$\uparrow$0.8}  & $\downarrow$1.8  & $\downarrow$8.8 \\
Claude-4.6 Opus & 48.5 & 30.6 & {\color{ForestGreen}$\uparrow$1.2}  & {\color{ForestGreen}$\uparrow$0.3}  & {\color{ForestGreen}$\uparrow$2.2}  & {\color{ForestGreen}$\uparrow$1.9}  & $\downarrow$0.2   & $\downarrow$0.9  & {\color{ForestGreen}$\uparrow$5.1}  & {\color{ForestGreen}$\uparrow$2.8}  & 80.7 & {\color{gray}0.0}                   & {\color{ForestGreen}$\uparrow$3.8}  & $\downarrow$2.0  & $\downarrow$0.3 \\
GPT-5.2         & 48.9 & 27.9 & {\color{ForestGreen}$\uparrow$2.9}  & $\downarrow$2.9  & $\downarrow$0.3   & $\downarrow$0.5   & $\downarrow$1.5   & $\downarrow$0.8  & $\downarrow$1.6   & $\downarrow$4.1   & 87.7 & $\downarrow$7.3   & $\downarrow$1.3   & $\downarrow$4.3  & $\downarrow$5.7 \\
Gemini-3.0 Pro  & 50.9 & 30.8 & $\downarrow$2.0   & $\downarrow$3.2  & $\downarrow$1.0   & $\downarrow$0.6   & $\downarrow$1.3   & $\downarrow$1.2  & {\color{ForestGreen}$\uparrow$0.1}  & $\downarrow$0.5   & 88.3 & {\color{ForestGreen}$\uparrow$1.2}  & $\downarrow$2.5   & $\downarrow$2.3  & $\downarrow$0.3 \\
Gemini-3.1 Pro  & 47.1 & 31.6 & {\color{ForestGreen}$\uparrow$1.9}  & {\color{ForestGreen}$\uparrow$1.4}  & $\downarrow$4.3   & $\downarrow$3.0   & {\color{ForestGreen}$\uparrow$0.3}  & $\downarrow$0.8  & $\downarrow$5.0   & $\downarrow$4.5   & 90.2 & {\color{ForestGreen}$\uparrow$0.5}  & $\downarrow$3.2   & $\downarrow$2.7  & {\color{gray}0.0} \\
Qwen3-VL-8B    & 40.7 & 34.1 & $\downarrow$6.1   & $\downarrow$5.6  & $\downarrow$0.1   & $\downarrow$1.2   & {\color{ForestGreen}$\uparrow$1.4}  & $\downarrow$0.2  & $\downarrow$5.6   & $\downarrow$6.4   & 50.7 & {\color{ForestGreen}$\uparrow$3.0}  & $\downarrow$1.3   & $\downarrow$2.3  & $\downarrow$3.3 \\
\bottomrule
\end{tabular}}
\vspace{-0.3em}
\end{table*}

\noindent\textbf{(1) Context utilization.}
We provide diagram objects as context across all three tasks (Tables~\ref{tab:ablation_d2c_dqa} and~\ref{tab:ablation_de}). On D2C-E, objects improve all models consistently: F1$_{\text{avg}}$ gains 4.1--10.6 on preserve and 3.3--7.1 on edit; thus objects may help models locate which elements to keep and which to modify. On D2C-P, results are mixed: four models gain, while Gemini-3.0~Pro (-2.0) and Qwen3-VL-8B (-6.1) degrade. At code level, only Claude-4.6~Opus (+0.3) and Gemini-3.1~Pro (+1.4) gain, showing objects aid element perception more than code synthesis. On DQA, most models benefit, while GPT-5.2 drops sharply (-7.3), indicating it cannot map low-level objects to semantic reasoning.

\noindent\colorbox{blue!8}{\parbox{\dimexpr\columnwidth-2\fboxsep}{%
\textbf{Takeaway:} Most models can use objects to ground local edits but struggle to integrate them into syntax or domian knowledege reasoning.}}

\noindent\textbf{(2) Tool use.}
Most models exhibit weak tool-calling ability.  The sharpest failure is Gemini-3.1~Pro (-4.3 F1$_{avg}$ on D2C-P). Manual inspection of trajectory reveals that model iteratively queries without a clear stopping criterion, causing context rot until terminated by the token-length limit (tool-call counts and failure rates in Appendix~\S\ref{sec:appx_tool_stats}); Qwen3-VL-8B trajectory issues poorly targeted queries, lacking ability of what to  query in the search tool. Claude-4.6~Opus performs best, with consistent gains on both DC (+2.2 F1$_{avg}$) and DE (+4.4/+2.2).

\noindent\colorbox{blue!8}{\parbox{\dimexpr\columnwidth-2\fboxsep}{%
\textbf{Takeaway:} Only Claude-4.6~Opus consistently benefits from tool access; Gemini-3.1~Pro suffers from excessive retrieval; Qwen3-VL-8B fails at querying.}}

\noindent\textbf{(3) State management.}
We require models to first generate the diagram's TikZ code, then build upon it for editing (D2C-E) or answering (DQA). On D2C-E, most models degrade on preserve F1$_{\text{avg}}$, indicating the limitation of managing intermediate code states for precise editing. Claude-4.6~Opus (+2.5/+2.6) and GPT-5.2 (+2.5/+3.5) are exceptions, successfully building on their generated code. On DQA, Claude-4.6~Opus gains substantially (+3.8) and Seed-2.0~Pro gains marginally (+0.8), while all other models degrade (-1.3 to -3.2).

\noindent\colorbox{blue!8}{\parbox{\dimexpr\columnwidth-2\fboxsep}{%
\textbf{Takeaway:} Most models fail to manage intermediate code states across steps; only Claude-4.6~Opus maintains coherence from coding to editing\&reasoning.}}

\noindent\textbf{(4) Planning.}
We test whether models can coordinate multiple agentic abilities: combining objects with tool use for D2C-P, optionally generating code before editing  or answering, and coordinating tool use with optional code generation (D2C-E and DU). On D2C-P, Claude-4.6~Opus (+5.1/+2.8) and Seed-2.0~Pro (+2.6/+1.5) coordinate both resources successfully, while (-1.6/-4.1), Gemini-3.1~Pro (-5.0/-4.5), and Qwen3-VL-8B (-5.6/-6.4) degrade. On D2C-E, most models show marginal or negative changes under optional code generation; adding tool access recovers Claude-4.6~Opus (+4.1/+3.0) and GPT-5.2 (+2.6/+1.9), but degrades Gemini-3.1~Pro (-4.4/-0.9). On DQA, all six models degrade under optional code generation, and adding tool access amplifies degradation for Seed-2.0~Pro (-8.8) and GPT-5.2 (-5.7).

\noindent\colorbox{blue!8}{\parbox{\dimexpr\columnwidth-2\fboxsep}{%
\textbf{Takeaway:} Planning is the weakest agentic capability; no model reliably composes multiple abilities as task complexity grows from D2C-P to DQA.}}



\begin{table*}[t]
\centering
\caption{Agentic evaluation results across D2C-E (S6--S11) with preserve\,(p)\,/\,edit\,(e) split. {\color{basetxt}S6} uses {\color{basetxt}direct editing}; {\color{ctxtxt}\textbf{Context utilization}: S7} provides diagram objects as context; {\color{tooltxt}\textbf{Tool use}: S8} provides TikZ search tool; {\color{statetxt}\textbf{State management}: S9} requires editing building upon the coding states;
{\color{plantxt}\textbf{Planning}: S10} optionally generates code for editing; {\color{plantxt}S11} adds tool access to S10 (Table~\ref{tab:context_settings}). $\Delta$ (\textcolor{ForestGreen}{$\uparrow$}\,gain, $\downarrow$\,drop) over direct editing.
}
\label{tab:ablation_de}
\vspace{-0.15cm}
\small
\renewcommand{\arraystretch}{1.3}
\setlength{\tabcolsep}{2.5pt}
\resizebox{\textwidth}{!}{%
\begin{tabular}{l|cc|cc|cc|cc|cc|cc}
\toprule
\multirow{2}{*}{\textbf{Model}}
& \multicolumn{2}{c|}{\textcolor{basetxt}{\textbf{S6}}}
& \multicolumn{2}{c|}{\textcolor{ctxtxt}{\textbf{S7}}}
& \multicolumn{2}{c|}{\textcolor{tooltxt}{\textbf{S8}}}
& \multicolumn{2}{c|}{\textcolor{statetxt}{\textbf{S9}}}
& \multicolumn{2}{c|}{\textcolor{plantxt}{\textbf{S10}}}
& \multicolumn{2}{c}{\textcolor{plantxt}{\textbf{S11}}} \\
\cmidrule(lr){2-3}\cmidrule(lr){4-5}\cmidrule(lr){6-7}\cmidrule(lr){8-9}\cmidrule(lr){10-11}\cmidrule(lr){12-13}
& {F1$_{\text{avg}}$(p/e)} & {CBLEU(p/e)}
& {F1$_{\text{avg}}$(p/e)} & {CBLEU(p/e)}
& {F1$_{\text{avg}}$(p/e)} & {CBLEU(p/e)}
& {F1$_{\text{avg}}$(p/e)} & {CBLEU(p/e)}
& {F1$_{\text{avg}}$(p/e)} & {CBLEU(p/e)}
& {F1$_{\text{avg}}$(p/e)} & {CBLEU(p/e)} \\
\midrule
Seed-2.0 Pro
  & 50.6/28.0 & 34.3/1.5
  & {\color{ForestGreen}$\uparrow$9.3}/{\color{ForestGreen}$\uparrow$7.1} & {\color{ForestGreen}$\uparrow$6.6}/{\color{ForestGreen}$\uparrow$0.5}
  & {\color{ForestGreen}$\uparrow$0.7}/{\color{ForestGreen}$\uparrow$0.2} & {\color{ForestGreen}$\uparrow$0.7}/{\color{ForestGreen}$\uparrow$0.1}
  & $\downarrow$2.3/$\downarrow$0.4 & $\downarrow$1.3/{\color{ForestGreen}$\uparrow$0.1}
  & {\color{gray}0.0}/$\downarrow$0.4 & {\color{ForestGreen}$\uparrow$0.1}/$\downarrow$0.3
  & {\color{ForestGreen}$\uparrow$0.6}/{\color{ForestGreen}$\uparrow$0.6} & {\color{ForestGreen}$\uparrow$0.1}/{\color{gray}0.0} \\
Claude-4.6 Opus
  & 55.9/28.3 & 39.3/1.3
  & {\color{ForestGreen}$\uparrow$8.9}/{\color{ForestGreen}$\uparrow$6.0} & {\color{ForestGreen}$\uparrow$6.0}/{\color{gray}0.0}
  & {\color{ForestGreen}$\uparrow$4.4}/{\color{ForestGreen}$\uparrow$2.2} & {\color{ForestGreen}$\uparrow$4.7}/{\color{ForestGreen}$\uparrow$0.5}
  & {\color{ForestGreen}$\uparrow$2.5}/{\color{ForestGreen}$\uparrow$2.6} & {\color{ForestGreen}$\uparrow$2.5}/{\color{gray}0.0}
  & $\downarrow$1.5/$\downarrow$0.2 & $\downarrow$0.4/{\color{ForestGreen}$\uparrow$0.1}
  & {\color{ForestGreen}$\uparrow$4.1}/{\color{ForestGreen}$\uparrow$3.0} & {\color{ForestGreen}$\uparrow$5.6}/{\color{ForestGreen}$\uparrow$0.7} \\
GPT-5.2
  & 54.9/29.3 & 38.6/1.2
  & {\color{ForestGreen}$\uparrow$10.6}/{\color{ForestGreen}$\uparrow$5.9} & {\color{ForestGreen}$\uparrow$7.2}/{\color{ForestGreen}$\uparrow$0.1}
  & {\color{ForestGreen}$\uparrow$1.3}/{\color{ForestGreen}$\uparrow$2.6} & {\color{ForestGreen}$\uparrow$2.4}/{\color{gray}0.0}
  & {\color{ForestGreen}$\uparrow$2.5}/{\color{ForestGreen}$\uparrow$3.5} & {\color{ForestGreen}$\uparrow$2.7}/{\color{ForestGreen}$\uparrow$0.1}
  & {\color{ForestGreen}$\uparrow$0.9}/{\color{ForestGreen}$\uparrow$1.1} & {\color{ForestGreen}$\uparrow$1.6}/{\color{ForestGreen}$\uparrow$0.3}
  & {\color{ForestGreen}$\uparrow$2.6}/{\color{ForestGreen}$\uparrow$1.9} & {\color{ForestGreen}$\uparrow$2.9}/{\color{ForestGreen}$\uparrow$0.4} \\
Gemini-3.0 Pro
  & 58.8/36.1 & 40.9/1.9
  & {\color{ForestGreen}$\uparrow$4.1}/{\color{ForestGreen}$\uparrow$3.3} & {\color{ForestGreen}$\uparrow$2.4}/{\color{ForestGreen}$\uparrow$0.4}
  & {\color{ForestGreen}$\uparrow$2.2}/$\downarrow$0.1 & $\downarrow$0.1/{\color{ForestGreen}$\uparrow$0.5}
  & $\downarrow$1.1/$\downarrow$0.6 & $\downarrow$2.0/{\color{ForestGreen}$\uparrow$0.2}
  & $\downarrow$0.4/$\downarrow$0.5 & $\downarrow$1.7/{\color{ForestGreen}$\uparrow$0.2}
  & {\color{ForestGreen}$\uparrow$2.8}/{\color{ForestGreen}$\uparrow$0.8} & {\color{ForestGreen}$\uparrow$0.7}/{\color{ForestGreen}$\uparrow$0.5} \\
Gemini-3.1 Pro
  & 57.6/38.1 & 38.0/2.5
  & {\color{ForestGreen}$\uparrow$6.6}/{\color{ForestGreen}$\uparrow$4.8} & {\color{ForestGreen}$\uparrow$4.6}/{\color{ForestGreen}$\uparrow$0.6}
  & $\downarrow$8.5/$\downarrow$3.5 & $\downarrow$5.3/{\color{ForestGreen}$\uparrow$0.8}
  & $\downarrow$5.2/$\downarrow$3.7 & $\downarrow$2.9/{\color{ForestGreen}$\uparrow$0.2}
  & {\color{ForestGreen}$\uparrow$0.6}/{\color{ForestGreen}$\uparrow$1.7} & {\color{ForestGreen}$\uparrow$0.5}/{\color{ForestGreen}$\uparrow$0.7}
  & $\downarrow$4.4/$\downarrow$0.9 & $\downarrow$2.9/{\color{ForestGreen}$\uparrow$1.4} \\
Qwen3-VL-8B
  & 47.0/22.2 & 34.2/0.8
  & {\color{ForestGreen}$\uparrow$8.4}/{\color{ForestGreen}$\uparrow$3.3} & {\color{ForestGreen}$\uparrow$4.3}/{\color{ForestGreen}$\uparrow$0.2}
  & $\downarrow$2.0/$\downarrow$0.2 & $\downarrow$2.2/{\color{ForestGreen}$\uparrow$0.2}
  & $\downarrow$1.0/{\color{ForestGreen}$\uparrow$2.5} & $\downarrow$1.4/{\color{ForestGreen}$\uparrow$0.4}
  & $\downarrow$3.3/$\downarrow$1.0 & $\downarrow$3.0/{\color{gray}0.0}
  & {\color{gray}0.0}/{\color{ForestGreen}$\uparrow$1.7} & $\downarrow$0.9/{\color{ForestGreen}$\uparrow$0.3} \\
\bottomrule
\end{tabular}} 
\vspace{-0.5em}
\end{table*}

\section{Conclusion}
Scientific diagrams are an important visual medium for expressing abstract ideas, and their TikZ code representation can be natively compiled inline within \LaTeX{} during scientific paper writing. Vibe writing platforms provide MLLM-assisted paper writing, and one important feature is converting scientific diagrams directly into TikZ code. We present \ourBench{}, a TikZ-based benchmark with 16 evaluation settings across diagram-to-code parsing, editing, and understanding alongside agentic settings per task. We also build a TikZ search tool as an MCP server that any model can selectively query TikZ syntactic references without any model-specific implementations. Comprehensive evaluation of 12 MLLMs reveals their weaknesses in coding diagrams into TikZ, and stronger agentic abilities, \eg, context utilization and tool use, could help. We hope our analysis can inspire further research to build more thorough evaluation settings and develop methods to improve MLLM-assisted vibe writing workspaces.

\section*{Acknowledgements}
We thank all 13 annotators for their contributions to data quality assurance. Collectively, they cross-validated over 2,000 scientific diagrams across six domains. Three annotators are co-authors. The ten non-author contributors include one research scientist from NetMind.ai, eight PhD students from Nanjing University of Science and Technology (NJUST), and one PhD student from Westlake University. Kindly refer to \S~\ref{sec:author_state} for details of their contributions.

\clearpage
\bibliographystyle{unsrt}
\bibliography{main}

\appendix

\renewcommand{\theHsection}{appendix.\thesection}
\renewcommand{\theHsubsection}{appendix.\thesubsection}

\setcounter{table}{0}
\setcounter{figure}{0}
\renewcommand{\thetable}{\thesection.\arabic{table}}
\renewcommand{\thefigure}{\thesection.\arabic{figure}}
\renewcommand{\theHtable}{appendix.\thetable}
\renewcommand{\theHfigure}{appendix.\thefigure}
\makeatletter
\@addtoreset{table}{section}
\@addtoreset{figure}{section}
\makeatother

\setcounter{topnumber}{4}
\setcounter{bottomnumber}{2}
\setcounter{totalnumber}{6}
\renewcommand{\topfraction}{0.95}
\renewcommand{\bottomfraction}{0.90}
\renewcommand{\textfraction}{0.05}
\renewcommand{\floatpagefraction}{0.88}
\setlength{\textfloatsep}{7pt plus 1pt minus 2pt}
\setlength{\floatsep}{6pt plus 1pt minus 1pt}
\setlength{\intextsep}{7pt plus 1pt minus 2pt}
\captionsetup{skip=2pt}

\titlecontents{lsection}[0em]{\smallskip\bfseries}
  {\thecontentslabel\quad}{}{~\dotfill~\contentspage}
\titlecontents{lsubsection}[2.3em]{}
  {\thecontentslabel\quad}{}{~\dotfill~\contentspage}
\thispagestyle{empty}
\section*{Contents}
\label{sec:appx_overview}
\makeatletter
\makeatother
\startcontents[appendix]
\printcontents[appendix]{l}{1}[2]{\providecommand{\authcount}[1]{}}

\newpage
\raggedbottom
\section{Related Works}
\label{sec:appx_rela}


\subsection{Diagram Parsing}
Diagram parsing extracts the topology of a diagram, \eg, the basic elements and their relationships. This requires symbolic perception instead of pixel-level understanding. A geometric primitive carries meaning as a symbol: its semantics come from its type, path and spatial relationships, not from its pixel values. For example, a circle remains a circle regardless of its color, size or line thickness. This contrasts with natural images, where each pixel contributes semantic content through color, texture, and intensity.  This distinction explains why MLLMs that perform well on natural image tasks struggle with diagrams~\cite{sun2025mathblind}.
%
A common surrogate is Optical Character Recognition (OCR), where models extract text annotations to interpret diagrams. However, OCR captures only textual content while ignoring geometric topology, thus failing at structural relationship extraction \cite{yoshida2026nodes,li2025chainofregion}. Consequently, error rates remain high for domain-specific diagrams like scientific charts and Entity-Relationship (ER) schemas \cite{sami2025challenges}. To address this, recent work integrates visual parsing with domain ontologies \cite{thangaraj2025ontology} or uses Region Decomposition \cite{shiinoki2025overcoming}. These methods capture cross-node dependencies, transforming pixel-level features into high-level semantics. An alternative line of work represents diagrams as code, called diagram-to-code parsing, which is the focus of this paper. We review these methods in \S\ref{sec:rela}\&Tabel~\ref{tab:comparison}.

\subsection{Diagram Editing}
Diagram editing is shifting from manual graphical manipulation to intent-driven editing using natural language \cite{yu2020flowsense,wei2025words}. Unlike general image editing via diffusion models, editing technical diagrams requires strict global logical coherence. Models must modify local attributes without causing structural damage \cite{wang2025editclip}. Recent frameworks solve this by parsing instructions and modifying vector graphics or code directly \cite{tan2025sketchagent}. For example, DiagramAgent uses a multi-agent system. It decomposes user instructions into explicit plans, allowing specialized agents to collaboratively edit and verify \LaTeX/TikZ or DOT representations \cite{xingchen2026davinci}. Other methods combine visual representation learning with autoregressive frameworks, such as VAREdit \cite{mao2025visual}. These approaches bridge the gap between language intents and vector graphics, safely handling complex topological changes.

\subsection{Vibe Writing}
With vibe coding~\cite{meske2025vibecoding}, researchers provide a diagram and describe the desired changes in natural language, and an MLLM generates the compilable code; a workflow that fundamentally depends on strong diagram-to-code generation. In the vide writing environment, human creators no longer need to write tedious macros or calculate coordinates. By providing simple logical intents or vibes, users guide multimodal agents to autonomously handle structural inference, code synthesis, and visual rendering. This radically reshapes how researchers create academic literature and diagrams.

Generating highly precise TikZ/\LaTeX\ code is a major research focus \cite{belouadi2024automatikz}. However, high-quality diagram-code pairs are scarce. To address this, models like TikZero use zero-shot architectures to decouple visual and textual features \cite{belouadi2025tikzero}. Similarly, DeTikZify optimizes generated TikZ code using Monte Carlo Tree Search (MCTS) and compiler feedback \cite{belouadi2024detikzify}. Furthermore, TikZilla improves visual fidelity using Reinforcement Learning (RL) and inverse graphics rewards \cite{greisinger2026tikzilla}, while MathCoder-VL synthesizes massive aligned datasets using a ``model-in-the-loop'' strategy \cite{wang2025mathcoder}.

\section{Benchmark Details}
\label{sec:appx_benchmark}

\subsection{Data Statistics}
\label{sec:appx_taxonomy}

\ourBench{} covers six scientific diagram types, each associated with one or more dedicated TikZ packages. The full dataset contains {3,744} diagrams paired with {18,305} evaluation instances (1 D2C-P + 2 D2C-E + 2 DQA per diagram), with a balanced \textbf{mini split} of {300} diagrams (50 per domain). Table~\ref{tab:appx_full_stats} summarizes the per-domain source distribution and task instance counts. We describe each type of diagram, as follows: 

\vspace{0.5em}
\noindent\textbf{(1) Diagram types:}
\begin{itemize}[leftmargin=*,nosep,topsep=2pt]
  \item \textbf{Charts} (960 diagrams). Statistical and data visualizations rendered primarily with \texttt{pgfplots}, including line plots, bar charts, scatter plots, pie charts, and histograms. Domain knowledge includes statistics, trend analysis, and extremum identification.

  \item \textbf{Planar Geometry} (601 diagrams). Two-dimensional geometric constructions drawn with \texttt{tikz} and \texttt{tkz-euclide}, including triangles, circles, polygons, and angle/distance annotations. Domain knowledge covers Euclidean notation, area/perimeter formulas, and geometric theorems.

  \item \textbf{3D Shapes} (237 diagrams). Three-dimensional solid geometry visualizations rendered with \texttt{pgfplots} and \texttt{tikz}, including prisms, pyramids, spheres, and cross-section illustrations. Domain knowledge includes solid geometry, surface area, and volume computation.

  \item \textbf{Graph Structures} (1,356 diagrams). Combinatorial and network-theoretic diagrams drawn with \texttt{tikz} and \texttt{tikz-network}, including directed/undirected graphs, trees, automata, and flow networks. Domain knowledge includes degree, connectivity, shortest paths, and tree properties.

  \item \textbf{Chemistry} (187 diagrams). Molecular and chemical structure diagrams rendered with \texttt{chemfig}, including organic molecules, functional groups, and reaction schemes. Domain knowledge covers bond types, atom labels, and functional group identification.

  \item \textbf{Circuit Diagrams} (403 diagrams). Electrical and electronic circuit schematics drawn with \texttt{circuitikz}, including resistive networks, RC/RLC circuits, and logic gates. Domain knowledge includes circuit topology, Ohm's law, and series/parallel analysis.
\end{itemize}

\begin{table}[!tbp]
\centering
\caption{Per-domain statistics of the \ourBench{}. ``Official'' denotes diagrams sourced from TikZ package documentation; ``Community'' denotes diagrams from community resources. Each diagram is paired with up to 5 evaluation questions (1 D2C-P + 2 D2C-E + 2 DQA); a small number of instances were removed during human cross-validation, so per-task totals are slightly below the nominal 1$\times$/2$\times$/2$\times$ diagram count.}
\label{tab:appx_full_stats}
\renewcommand{\arraystretch}{1.10}
\setlength{\tabcolsep}{4.mm}{
\resizebox{0.95\linewidth}{!}{%
\begin{tabular}{l r c rr rrr r rr}
\toprule
& & & \multicolumn{2}{c}{\textbf{Data Source}} & \multicolumn{3}{c}{\textbf{Task Instances}} & & \multicolumn{2}{c}{\textbf{Code Stats}} \\
\cmidrule(lr){4-5} \cmidrule(lr){6-8} \cmidrule(lr){10-11}
\textbf{Domain} & \textbf{\#Diag.} & \textbf{Primary TikZ Pkg.} & \textbf{Official} & \textbf{Community} & \textbf{D2C-P} & \textbf{D2C-E} & \textbf{DQA} & \textbf{Total} & \textbf{Avg.~Lines} & \textbf{Avg.~Chars} \\
\midrule
Charts               & 960 & \texttt{pgfplots} & 579 & 381 & 959 & 1,912 & 1,834 & 4,705 & 34.6 & 1,098 \\
Planar Geom.         & 601 & \texttt{tikz/tkz-euclide} & 253 & 348 & 598 & 1,162 & 1,118 & 2,878 & 26.9 & 1,074 \\
3D Shapes            & 237 & \texttt{pgfplots} & 146 & 91 & 236 & 471 & 455 & 1,162 & 29.6 & 988 \\
Graph Struct.        & 1,356 & \texttt{tikz} & 299 & 1,057 & 1,356 & 2,699 & 2,679 & 6,734 & 34.9 & 1,452 \\
Chemistry            & 187 & \texttt{chemfig} & 185 & 2 & 187 & 373 & 366 & 926 & 18.5 & 412 \\
Circuit              & 403 & \texttt{circuitikz} & 387 & 16 & 403 & 803 & 694 & 1,900 & 21.2 & 660 \\
\midrule
\textbf{Total} & \textbf{3,744} & -- & \textbf{1,849} & \textbf{1,895} & \textbf{3,739} & \textbf{7,420} & \textbf{7,146} & \textbf{18,305} & 30.9 & 1,134 \\
\bottomrule
\end{tabular}}
}
\end{table}

\vspace{0.5em}
\noindent\textbf{(2) D2C-E editing dimensions.}
Each D2C-E sample is annotated with one of four editing dimensions: \emph{text} (label/annotation modification), \emph{color} (fill/stroke color change), \emph{scope} (local element addition, deletion, or transformation), and \emph{layout} (global structural change such as chart type conversion or circuit topology modification). We exclude layout editing from 3D shapes and chemistry. Table~\ref{tab:appx_d2ce_dims} shows the per-domain breakdown.

\begin{table}[!tbp]
\centering
\caption{D2C-E editing dimension breakdown for the full dataset and mini split. 3D Shapes and Chemistry have no layout-dimension edits (marked ``--'').}
\label{tab:appx_d2ce_dims}
\renewcommand{\arraystretch}{1.10}
\setlength{\tabcolsep}{4.mm}{
\resizebox{0.95\linewidth}{!}{%
\begin{tabular}{l rrrrr rrrrr}
\toprule
& \multicolumn{5}{c}{\textbf{Full Dataset}} & \multicolumn{5}{c}{\textbf{Mini Split}} \\
\cmidrule(lr){2-6} \cmidrule(lr){7-11}
\textbf{Domain} & \textbf{Text} & \textbf{Color} & \textbf{Scope} & \textbf{Layout} & \textbf{Total} & \textbf{Text} & \textbf{Color} & \textbf{Scope} & \textbf{Layout} & \textbf{Total} \\
\midrule
Charts               & 438 & 194 & 365 & 915 & 1,912 & 21 & 10 & 19  & 50  & 100 \\
Planar Geom.         & 172 & 205 & 221 & 564 & 1,162 & 20 & 17 & 14  & 49  & 100 \\
3D Shapes            & 81 & 123 & 267 & -- & 471 & 14 & 23 & 63  & --  & 100 \\
Graph Struct.        & 315 & 319 & 735 & 1,330 & 2,699 & 10 & 12 & 28  & 50  & 100 \\
Chemistry            & 57 & 139 & 177 & -- & 373 & 9  & 41 & 50  & --  & 100 \\
Circuit              & 122 & 71 & 215 & 395 & 803 & 18 & 7  & 26  & 49  & 100 \\
\midrule
\textbf{Total} & \textbf{1,185} & \textbf{1,051} & \textbf{1,980} & \textbf{3,204} & \textbf{7,420} & \textbf{92} & \textbf{110} & \textbf{200} & \textbf{198} & \textbf{600} \\
\bottomrule
\end{tabular}}
}
\end{table}

\noindent\textbf{(3) DQA question types.}
DQA comprises two question types: descriptive and reasoning. Reasoning questions are further split into standard and \emph{what-if} types (definitions in \S\ref{sec:quesion_define}). Table~\ref{tab:appx_dqa_types} shows the per-domain breakdown. 
\begin{table}[!tbp]
\centering
\caption{DQA question type breakdown for the full dataset and mini split. Standard questions cover both descriptive and reasoning types; what-if questions are reasoning-only and require hypothetical inference conditioned on element modifications.}
\label{tab:appx_dqa_types}
\renewcommand{\arraystretch}{1.10}
\setlength{\tabcolsep}{4.mm}{
\resizebox{0.9\linewidth}{!}{%
\begin{tabular}{l rrrr rrrr}
\toprule
& \multicolumn{4}{c}{\textbf{Full Dataset}} & \multicolumn{4}{c}{\textbf{Mini Split}} \\
\cmidrule(lr){2-5} \cmidrule(lr){6-9}
\textbf{Domain} & \textbf{Standard} & \textbf{What-if} & \textbf{Total} & \textbf{\%\,What-if} & \textbf{Standard} & \textbf{What-if} & \textbf{Total} & \textbf{\%\,What-if} \\
\midrule
Charts               & 1,244 & 590 & 1,834 & 32.2 & 78 & 22 & 100 & 22.0 \\
Planar Geom.         & 780 & 338 & 1,118 & 30.2 & 65 & 35 & 100 & 35.0 \\
3D Shapes            & 344 & 111 & 455 & 24.4 & 75 & 25 & 100 & 25.0 \\
Graph Struct.        & 1,787 & 892 & 2,679 & 33.3 & 72 & 28 & 100 & 28.0 \\
Chemistry            & 250 & 116 & 366 & 31.7 & 74 & 26 & 100 & 26.0 \\
Circuit              & 480 & 214 & 694 & 30.8 & 70 & 30 & 100 & 30.0 \\
\midrule
\textbf{Total} & \textbf{4,885} & \textbf{2,261} & \textbf{7,146} & \textbf{31.6} & \textbf{434} & \textbf{166} & \textbf{600} & \textbf{27.7} \\
\bottomrule
\end{tabular}}
}
\end{table}

\noindent\textbf{(4) DQA answer formats.}
Each DQA question specifies one of three output instruction types for answering: \textsc{OI-NUM} (numerical answer, 5,013 questions), \textsc{OI-TERM} (domain-specific term or label, 1,779 questions), and \textsc{OI-LIST} (list of elements, 354 questions). This distribution reflects that the majority of questions require numerical computation (\eg, calculating area, degree, or resistance values), followed by identification of domain-specific entities.

\vspace{0.5em}
\noindent\textbf{(5) Mini split.}
The mini split is a class-balanced subset of the full dataset, containing exactly {50} diagrams per domain ({300} total) with {1,500} evaluation instances (300 D2C-P + 600 D2C-E + 600 DQA). The per-domain task breakdowns of the mini split are shown in Tables~\ref{tab:appx_d2ce_dims} and~\ref{tab:appx_dqa_types}. 

\subsection{Task Generation Pipeline}
\label{sec:appx_pipeline}

Figure~\ref{fig:agentic_qa} illustrates the overall benchmark construction pipeline. Herein, we detail the task-specific generation procedure:

\noindent\textbf{(1) Diagram-to-Code Parsing (D2C-P).}
D2C-P uses fixed task templates instantiated directly from the original diagrams. The task question asks the model to convert the diagram image into compilable TikZ code, conditioned on a provided preamble (document class, required packages, and libraries). The original TikZ source code serves as the ground-truth answer. No agentic generation is required for this task.

\noindent\textbf{(2) Diagram-to-Code Editing (D2C-E).}
D2C-E employs an agentic pipeline with two main components: a \emph{generation agent} (Gemini-3 Flash) and a \emph{verifier agent} comprising two judge models (GPT-5.2 and Gemini-3 Pro) for cross-validation. The pipeline proceeds as follows:
\begin{itemize}[nosep,topsep=2pt]
    \item \textbf{Template predefinition.} We manually design a pool of editing task templates across four dimensions (text, color, scope, layout) for each of the six domains (Table~\ref{tab:de_tasks}). Each template contains placeholders for diagram-specific elements (\eg, labels, colors, vertices, and edges).

    \item \textbf{Task selection.} Given a diagram and its TikZ source code, the generation agent selects two editing tasks from the template pool of the corresponding domain, constrained to cover distinct editing dimensions to ensure diversity.

    \item \textbf{Answer generation.} The generation agent instantiates the selected templates with specific diagram elements and generates the corresponding modified TikZ code. Multi-engine compilation (\texttt{pdflatex}$\rightarrow$\texttt{lualatex}$\rightarrow$\texttt{xelatex}) is performed to verify executability.

    \item \textbf{Agent verification.} The verifier agents evaluate generated results against predefined criteria: all edits must be visually observable and conform to the selected editing dimension, and unaffected elements must remain unchanged. A case is accepted only when both verifier agents agree.

    \item \textbf{Iterative refinement.} Failed cases, together with verifier feedback, are returned to the generation agent for regeneration, with a maximum of three attempts.

    \item \textbf{Human verification.} All accepted instances are manually reviewed by 13 graduate students. Each annotator's assigned instances are cross-validated by another annotator to ensure language clarity, logical consistency, and answer correctness.
\end{itemize}

\noindent\textbf{(3) Diagram Question Answering (DQA).}
The DQA generation pipeline follows the same agentic workflow as D2C-E; only the task templates differ. The generation agent produces descriptive and reasoning questions based on the diagram, its TikZ source code, and domain-specific templates encoding symbol semantics and domain knowledge (\S\ref{sec:appx_dqa_templates}). Template placeholders are instantiated using diagram elements extracted from the source code. For example, in organic chemistry, functional groups (\eg, COOH or OH) are extracted to generate questions such as ``which functional group is present in this molecule?''. The verifier agents then assess each (diagram, question, answer) triple for logical consistency with both visual evidence and domain knowledge; regeneration with verifier feedback is allowed for up to three attempts, and only validated QA pairs are retained. Human verification follows the same cross-validation protocol as D2C-E.

Each diagram yields five evaluation instances (1 D2C-P + 2 D2C-E + 2 DQA). The template-guided selection enforces dimensional diversity: D2C-E instances cover distinct editing dimensions per diagram, and DQA questions span both descriptive and reasoning types. All templates and rubric designs are detailed in~\S\ref{sec:appx_d2ce_templates} and~\S\ref{sec:appx_dqa_templates}.

\subsection{D2C-P Examples}
\label{sec:appx_d2cp_examples}

Figure~\ref{fig:appx_d2cp_showcase} presents representative D2C-P case studies across diagram domains. Each case shows the input diagram, the task prompt, the ground-truth TikZ code, and the rendered output.

\casegroupheader{d2c-header}{Diagram-to-Code Parsing (D2C-P)}

\begin{tcolorbox}[casebox, colback=d2c-bg, colframe=d2c-header!50]
\textbf{\footnotesize Charts Case 1}\\[2pt]
\begin{minipage}[t]{\caseleftwidth}
  \caseshowcasequestion{Recreate this diagram using \LaTeX. Provide the full, executable code.}
  \centering
  \includegraphics[width=\linewidth,height=3cm,keepaspectratio]{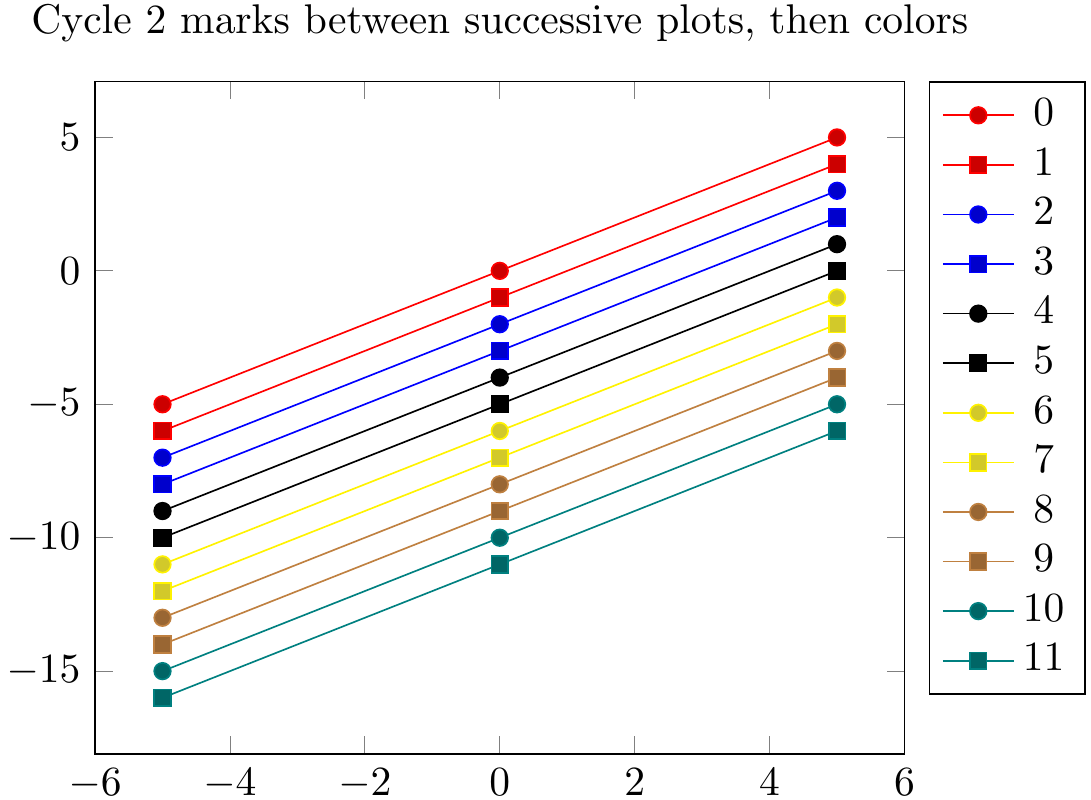}
\end{minipage}%
\hfill
\begin{minipage}[t]{\casecenterwidth}
  \textbf{A:}\\[-2pt]
  \begin{lstlisting}[style=tinycode]
% [1 packages: pgfplots]
\begin{tikzpicture}
\begin{axis}[
title={Cycle 2 marks between successive plots, then colors},
cycle multi list={
color list\nextlist
[2 of]mark list
},
  ...
\addplot {x-8};
\addplot {x-9};
\addplot {x-10};
\addplot {x-11};
\end{axis}
\end{tikzpicture}
  \end{lstlisting}
\end{minipage}%
\hfill
\begin{minipage}[t]{0.24\linewidth}
  {\footnotesize\textbf{Rendered Image:}}\\[3pt]
  \centering
  \includegraphics[width=\linewidth,height=3cm,keepaspectratio]{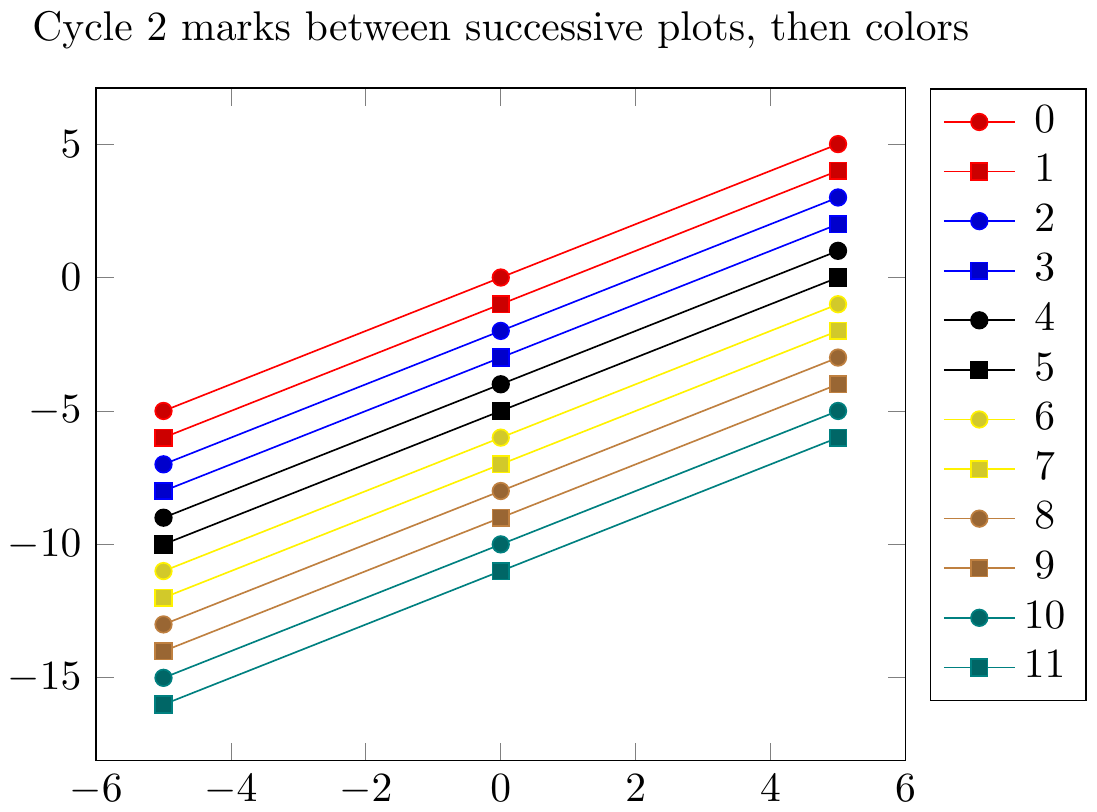}
\end{minipage}
\end{tcolorbox}%
\begin{tcolorbox}[casebox, colback=d2c-bg, colframe=d2c-header!50]
\textbf{\footnotesize Charts Case 2}\\[2pt]
\begin{minipage}[t]{\caseleftwidth}
  \caseshowcasequestion{Recreate this diagram using \LaTeX. Provide the full, executable code.}
  \centering
  \includegraphics[width=\linewidth,height=3cm,keepaspectratio]{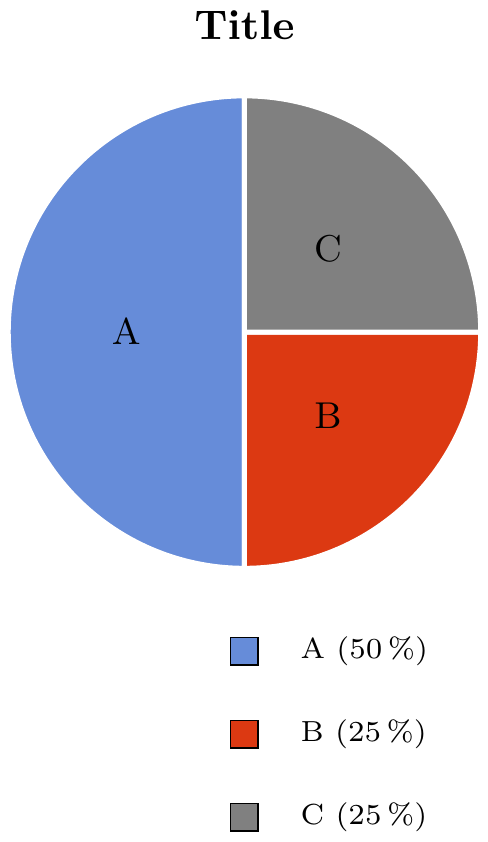}
\end{minipage}%
\hfill
\begin{minipage}[t]{\casecenterwidth}
  \textbf{A:}\\[-2pt]
  \begin{lstlisting}[style=tinycode]
% [1 packages: tikz]
\begin{tikzpicture}[pie chart,slice type={A}{blu},slice type={B}{rosso},slice type={C}{gray},pie values/.style={font=\small},scale=2]
\pie{Title}{50/A,25/B,25/C}\legend[shift={(0,-1cm)}]{{A (50\,\%)}/A,{B (25\,\%)}/B,{C (25\,\%)}/C}
\end{tikzpicture}
  \end{lstlisting}
\end{minipage}%
\hfill
\begin{minipage}[t]{0.24\linewidth}
  {\footnotesize\textbf{Rendered Image:}}\\[3pt]
  \centering
  \includegraphics[width=\linewidth,height=3cm,keepaspectratio]{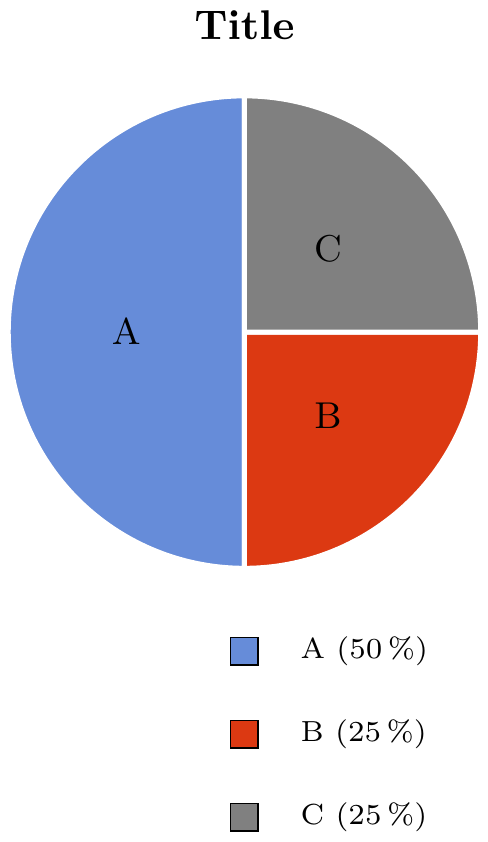}
\end{minipage}
\end{tcolorbox}%
\begin{tcolorbox}[casebox, colback=d2c-bg, colframe=d2c-header!50]
\textbf{\footnotesize Graph Structures Case 1}\\[2pt]
\begin{minipage}[t]{\caseleftwidth}
  \caseshowcasequestion{Recreate this diagram using \LaTeX. Provide the full, executable code.}
  \centering
  \includegraphics[width=\linewidth,height=3cm,keepaspectratio]{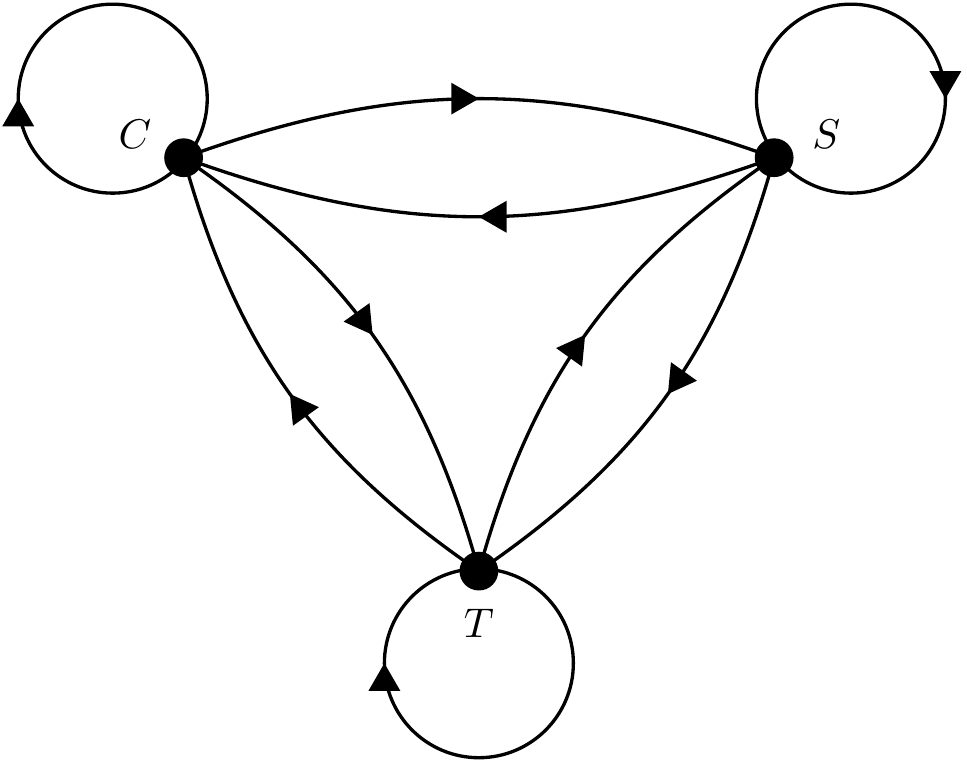}
\end{minipage}%
\hfill
\begin{minipage}[t]{\casecenterwidth}
  \textbf{A:}\\[-2pt]
  \begin{lstlisting}[style=tinycode]
\begin{tikzpicture}[>=triangle 60]
\draw[thick,fill=black] (0,0) circle (1.5mm);
\draw[thick,fill=black] (5,0) circle (1.5mm);
\draw[thick,fill=black] (2.5,-3.5) circle (1.5mm);
\node [left] at (-0.15, 0.2) {$C$};
\node [right] at (5.2,0.2) {$S$};
\node [below] at (2.5,-3.7) {$T$};
  ...
\draw[->-=.5,thick](2.5,-3.5) to [bend left=20](0,0);
\draw[->-=.5,thick](0,0) to [bend left=20](2.5,-3.5);
\draw[yscale=-1,yshift=8.55cm,->-=.5,thick](2.5,-4.27) circle (0.8cm);
\draw[->-=.5,thick](2.5,-3.5) to [bend left=20](5,0);
\draw[->-=.5,thick](5,0) to [bend left=20](2.5,-3.5);
\end{tikzpicture}
  \end{lstlisting}
\end{minipage}%
\hfill
\begin{minipage}[t]{0.24\linewidth}
  {\footnotesize\textbf{Rendered Image:}}\\[3pt]
  \centering
  \includegraphics[width=\linewidth,height=3cm,keepaspectratio]{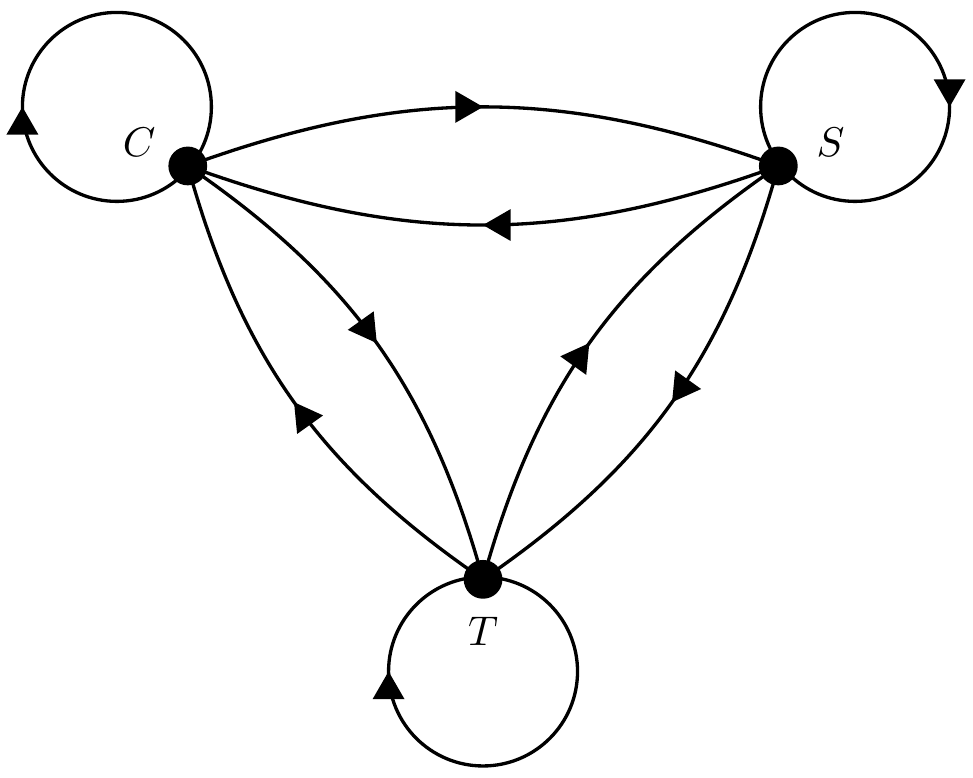}
\end{minipage}
\end{tcolorbox}%
\begin{tcolorbox}[casebox, colback=d2c-bg, colframe=d2c-header!50]
\textbf{\footnotesize Graph Structures Case 2}\\[2pt]
\begin{minipage}[t]{\caseleftwidth}
  \caseshowcasequestion{Recreate this diagram using \LaTeX. Provide the full, executable code.}
  \centering
  \includegraphics[width=\linewidth,height=3cm,keepaspectratio]{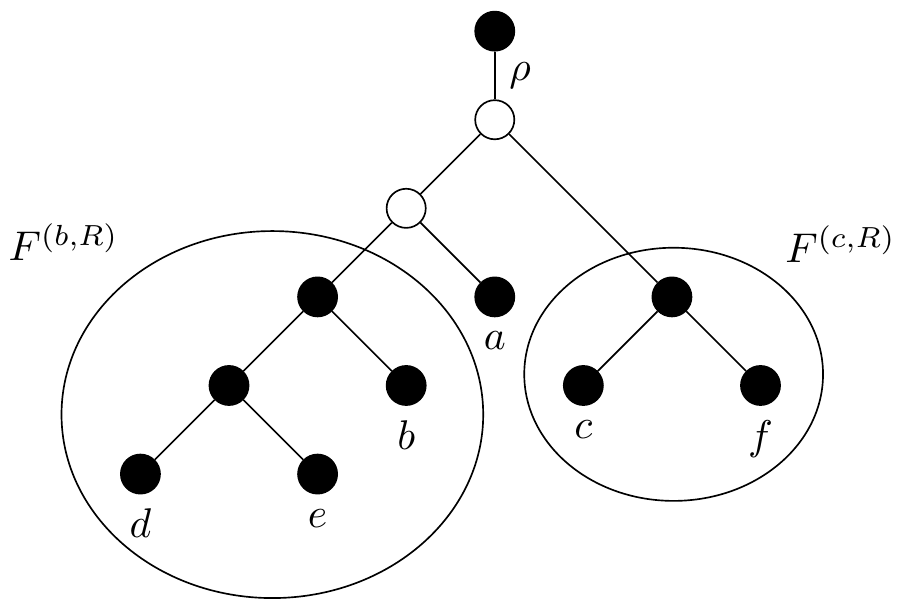}
\end{minipage}%
\hfill
\begin{minipage}[t]{\casecenterwidth}
  \textbf{A:}\\[-2pt]
  \begin{lstlisting}[style=tinycode]
% [5 packages: fontenc, inputenc, lmodern...]
\begin{tikzpicture}
[vertex/.style={minimum size=2pt,fill,draw,circle},
open/.style={fill=none},
sibling distance=1.5cm,level distance=.75cm,
every fit/.style={ellipse,draw,inner sep=-2pt},
leaf/.style={label={[name=#1]below:$#1$}},auto]
\node [vertex] (root) {}
  ...
child { node [vertex,leaf=c] {} }
child { node [vertex,leaf=f] {} } } }
edge from parent node {$\rho$} };
\node [fit=(d) (e) (b) (b's parent),label=above left:$F^{(b,R)}$] {};
\node [fit=(c) (f) (f's parent),label=above right:$F^{(c,R)}$] {};
\end{tikzpicture}
  \end{lstlisting}
\end{minipage}%
\hfill
\begin{minipage}[t]{0.24\linewidth}
  {\footnotesize\textbf{Rendered Image:}}\\[3pt]
  \centering
  \includegraphics[width=\linewidth,height=3cm,keepaspectratio]{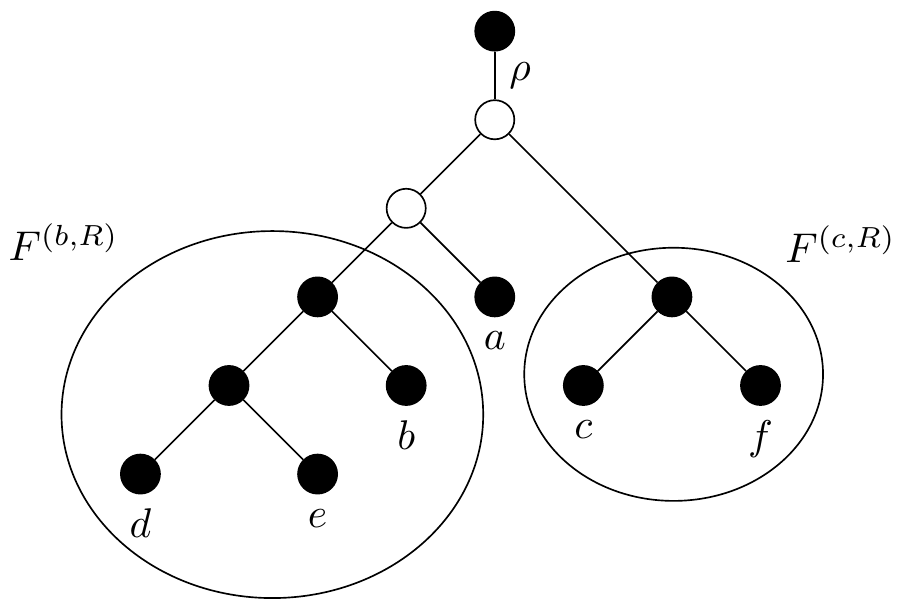}
\end{minipage}
\end{tcolorbox}%
\begin{tcolorbox}[casebox, colback=d2c-bg, colframe=d2c-header!50]
\textbf{\footnotesize Planar Geometry Case 1}\\[2pt]
\begin{minipage}[t]{\caseleftwidth}
  \caseshowcasequestion{Recreate this diagram using \LaTeX. Provide the full, executable code.}
  \centering
  \includegraphics[width=\linewidth,height=3cm,keepaspectratio]{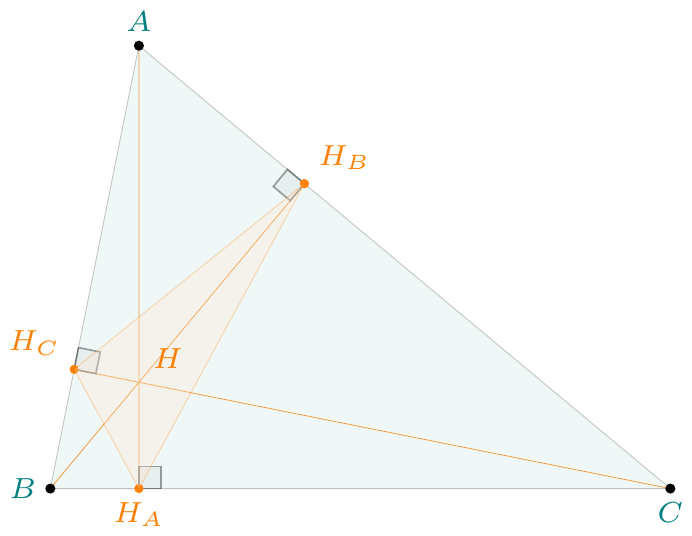}
\end{minipage}%
\hfill
\begin{minipage}[t]{\casecenterwidth}
  \textbf{A:}\\[-2pt]
  \begin{lstlisting}[style=tinycode]
% [3 packages: amsmath,amssymb, xcolor, tkz-euclide]
\begin{tikzpicture}[scale=.75]
\tkzDefPoints{1/5/A,0/0/B,7/0/C}
\tkzDefSpcTriangle[orthic](A,B,C){H_A,H_B,H_C}
\tkzDefTriangleCenter[ortho](B,C,A)
\tkzGetPoint{H}
\tkzDrawSegments[new](A,H_A B,H_B C,H_C)
\tkzMarkRightAngles[fill=gray!20,opacity=.5](A,H_A,C B,H_B,A C,H_C,A)
  ...
\tkzLabelPoints[left](B)
\tkzLabelPoints[above](A)
\tkzLabelPoints[new](H_A)
\tkzLabelPoints[new,above left](H_C)
\tkzLabelPoints[new,above right](H_B,H)
\end{tikzpicture}
  \end{lstlisting}
\end{minipage}%
\hfill
\begin{minipage}[t]{0.24\linewidth}
  {\footnotesize\textbf{Rendered Image:}}\\[3pt]
  \centering
  \includegraphics[width=\linewidth,height=3cm,keepaspectratio]{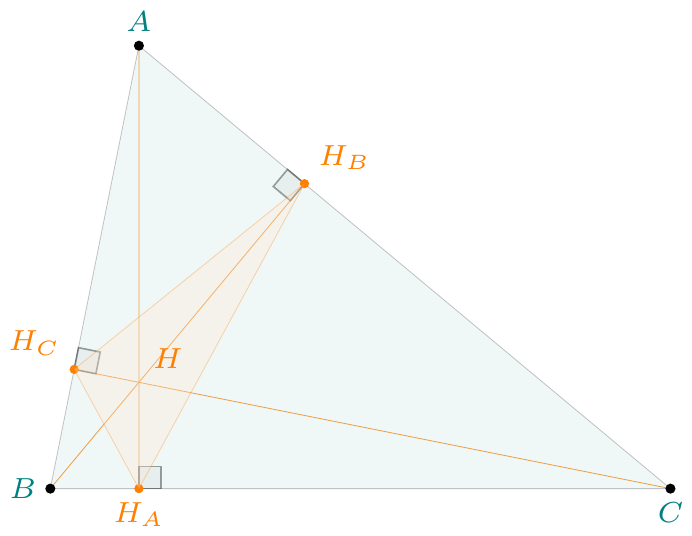}
\end{minipage}
\end{tcolorbox}%
\begin{tcolorbox}[casebox, colback=d2c-bg, colframe=d2c-header!50]
\textbf{\footnotesize Planar Geometry Case 2}\\[2pt]
\begin{minipage}[t]{\caseleftwidth}
  \caseshowcasequestion{Recreate this diagram using \LaTeX. Provide the full, executable code.}
  \centering
  \includegraphics[width=\linewidth,height=3cm,keepaspectratio]{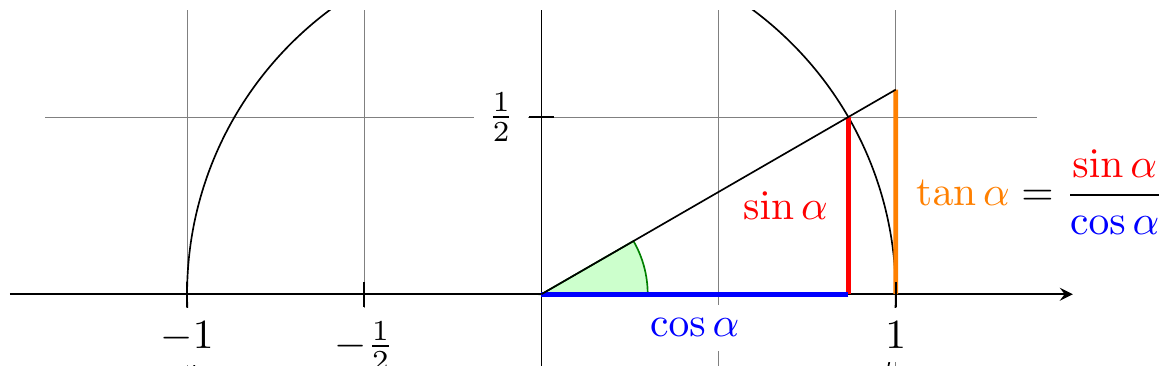}
\end{minipage}%
\hfill
\begin{minipage}[t]{\casecenterwidth}
  \textbf{A:}\\[-2pt]
  \begin{lstlisting}[style=tinycode]
% [1 packages: tikz]
\begin{tikzpicture}[scale=3]
\clip(-2,-0.2)rectangle(2,0.8);
\draw[step=.5cm,gray,very thin](-1.4,-1.4)grid(1.4,1.4);
\filldraw[fill=green!20,draw=green!50!black](0,0)--(3mm,0mm)arc[start angle=0,end angle=30,radius=3mm]--cycle;
\draw[->](-1.5,0)--(1.5,0)coordinate(x axis);
\draw[->](0,-1.5)--(0,1.5)coordinate(y axis);
\draw(0,0)circle[radius=1cm];
  ...
\path[name path=sloped line](0,0)--(30:1.5cm);
\draw[name intersections={of=upward line and sloped line,by=t}][very thick,orange](1,0)--node[right=1pt,fill=white]{$\displaystyle\tan\alpha=\frac{\sin\alpha}{\color{blue}\cos\alpha}$}(t);
\draw(0,0)--(t);
\foreach\x/\xtext in{-1,-0.5/-\frac{1}{2},1}\draw(\x cm,1pt)--(\x cm,-1pt)node[anchor=north,fill=white]{$\xtext$};
\foreach\y/\ytext in{-1,-0.5/-\frac{1}{2},0.5/\frac{1}{2},1}\draw(1pt,\y cm)--(-1pt,\y cm)node[anchor=east,fill=white]{$\ytext$};
\end{tikzpicture}
  \end{lstlisting}
\end{minipage}%
\hfill
\begin{minipage}[t]{0.24\linewidth}
  {\footnotesize\textbf{Rendered Image:}}\\[3pt]
  \centering
  \includegraphics[width=\linewidth,height=3cm,keepaspectratio]{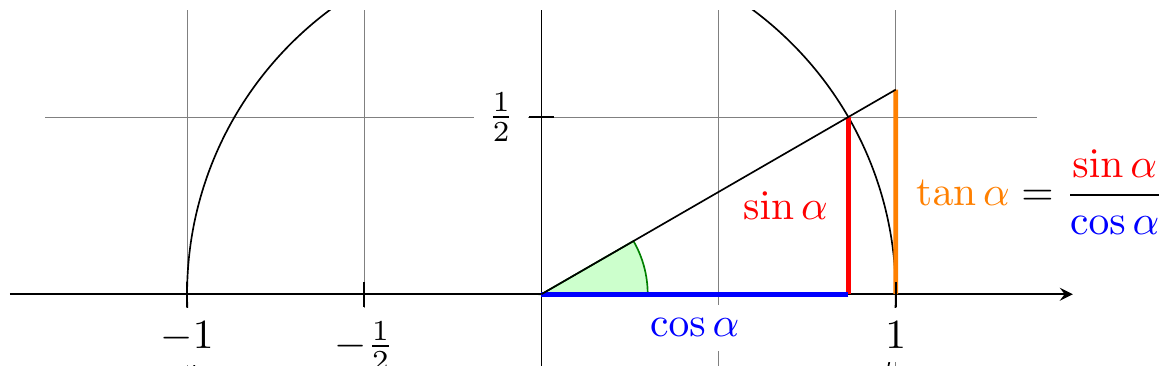}
\end{minipage}
\end{tcolorbox}%
\begin{tcolorbox}[casebox, colback=d2c-bg, colframe=d2c-header!50]
\textbf{\footnotesize 3D Shapes Case 1}\\[2pt]
\begin{minipage}[t]{\caseleftwidth}
  \caseshowcasequestion{Recreate this diagram using \LaTeX. Provide the full, executable code.}
  \centering
  \includegraphics[width=\linewidth,height=3cm,keepaspectratio]{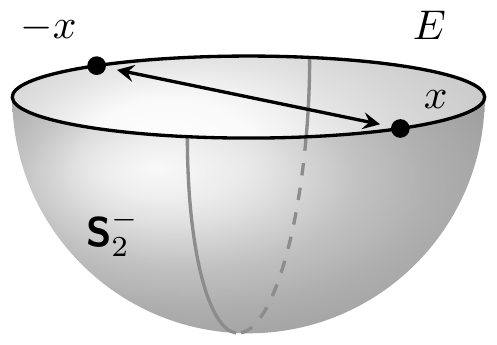}
\end{minipage}%
\hfill
\begin{minipage}[t]{\casecenterwidth}
  \textbf{A:}\\[-2pt]
  \begin{lstlisting}[style=tinycode]
% [2 packages: tikz,bm,pgfplots, tikz-3dplot]
\tdplotsetmaincoords{80}{140}
\begin{tikzpicture}[scale=2,tdplot_main_coords]
\tikzset{dot/.style={circle,fill,minimum size=4.5pt,inner sep=0pt,outer sep=0pt}
,hemispherebehind/.style={ball color=gray!20!white,opacity=0.3},hemispherefront/.style={ball color=gray!65!white,opacity=0.3},circlearc/.style={thick,gray!90},circlearchidden/.style={thick,dashed,gray!90},equator/.style={thick,black},diameter/.style={thick,black,stealth-stealth,shorten <=5pt,shorten >=5pt}}\newcommand{\hemispherefront}{(1cm,0)arc(0:-180:1cm and 1.8mm)arc(180:0:1cm and -1cm)}
\newcommand{\hemispherebehind}{(1cm,0)arc(0:-180:1cm and -1.8mm)arc(-180:0:1cm and 1cm)}
\newcommand{\equator}{(-1,0,0)arc(0:360:-1)}
\tdplotsetthetaplanecoords{35}\path[tdplot_rotated_coords,name path=semicircle](0,-1,0)arc(90:-90:-1);
  ...
\draw[circlearchidden,intersection segments={of=semicircle and hemisphere,sequence={L2}}];
\draw[circlearc,intersection segments={of=semicircle and hemisphere,sequence={L3}}];
\draw[equator]\equator node[pos=1,label={90:$E$}]{};
\draw[diameter](0,-1,0)node[dot,label={120:$-x$}]{}--(0,1,0)node[dot,label={30:$x$}]{};
\node at(.75,0,-.5){$\bm{\mathsf{S}}_2^-$};
\end{tikzpicture}
  \end{lstlisting}
\end{minipage}%
\hfill
\begin{minipage}[t]{0.24\linewidth}
  {\footnotesize\textbf{Rendered Image:}}\\[3pt]
  \centering
  \includegraphics[width=\linewidth,height=3cm,keepaspectratio]{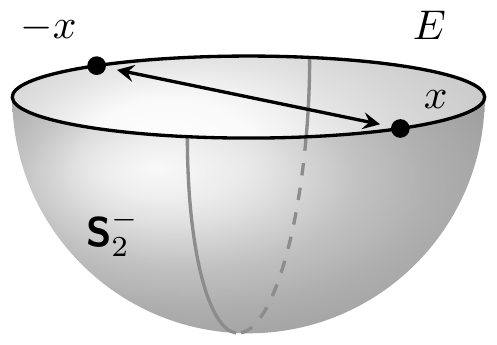}
\end{minipage}
\end{tcolorbox}%
\begin{tcolorbox}[casebox, colback=d2c-bg, colframe=d2c-header!50]
\textbf{\footnotesize 3D Shapes Case 2}\\[2pt]
\begin{minipage}[t]{\caseleftwidth}
  \caseshowcasequestion{Recreate this diagram using \LaTeX. Provide the full, executable code.}
  \centering
  \includegraphics[width=\linewidth,height=3cm,keepaspectratio]{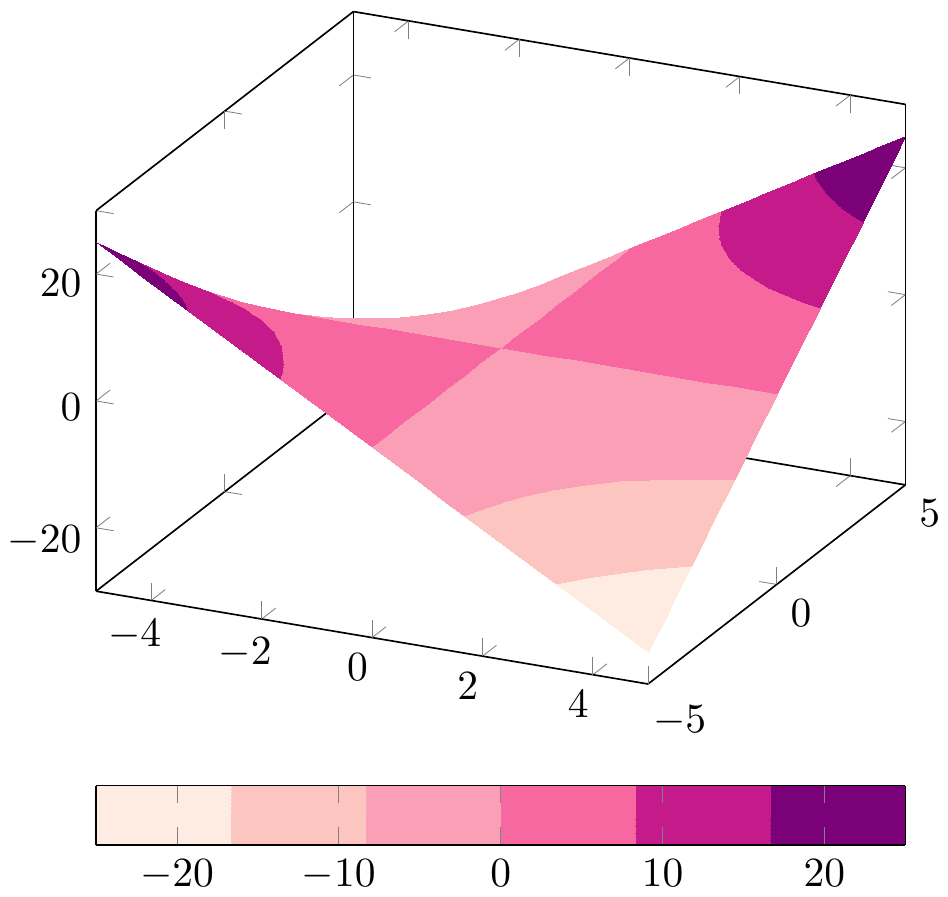}
\end{minipage}%
\hfill
\begin{minipage}[t]{\casecenterwidth}
  \textbf{A:}\\[-2pt]
  \begin{lstlisting}[style=tinycode]
% [1 packages: pgfplots]
\begin{tikzpicture}
\begin{axis}[
colormap access=piecewise constant,
colormap/RdPu-6,
colorbar horizontal,
]
\addplot3 [
surf,
shader=interp,
] {x*y};
\end{axis}
\end{tikzpicture}
  \end{lstlisting}
\end{minipage}%
\hfill
\begin{minipage}[t]{0.24\linewidth}
  {\footnotesize\textbf{Rendered Image:}}\\[3pt]
  \centering
  \includegraphics[width=\linewidth,height=3cm,keepaspectratio]{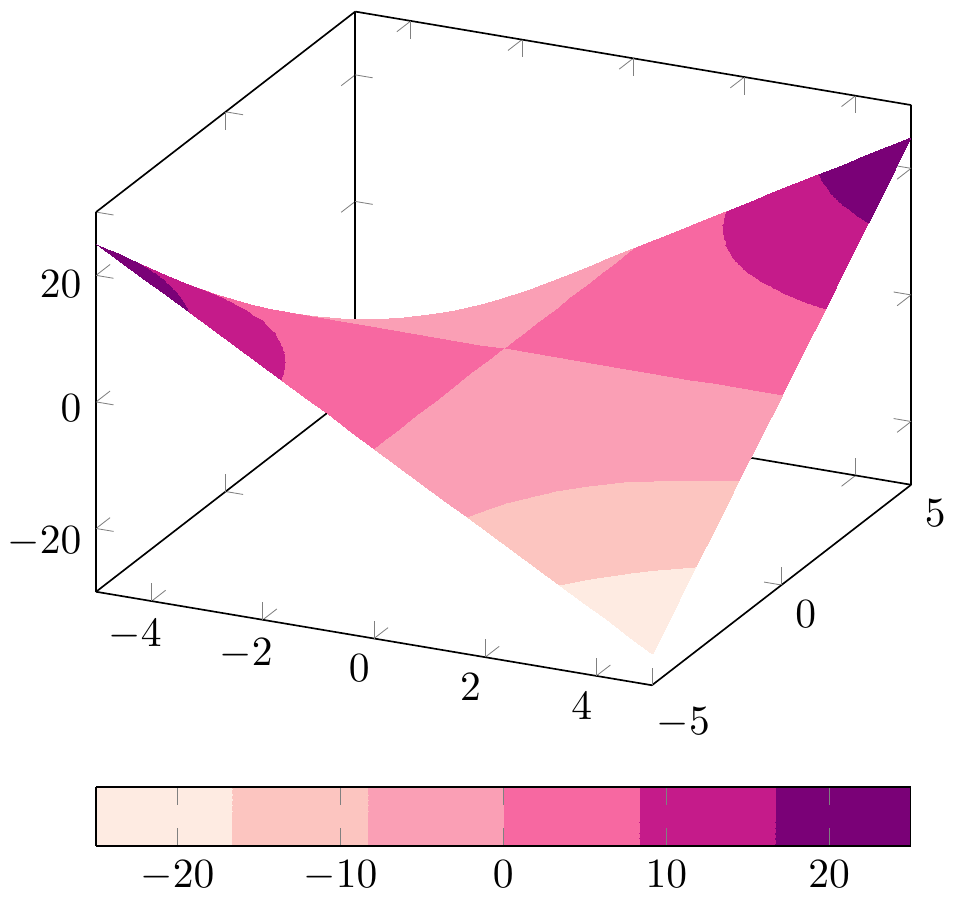}
\end{minipage}
\end{tcolorbox}%
\begin{tcolorbox}[casebox, colback=d2c-bg, colframe=d2c-header!50]
\textbf{\footnotesize Circuit Diagrams Case 1}\\[2pt]
\begin{minipage}[t]{\caseleftwidth}
  \caseshowcasequestion{Recreate this diagram using \LaTeX. Provide the full, executable code.}
  \centering
  \includegraphics[width=\linewidth,height=3cm,keepaspectratio]{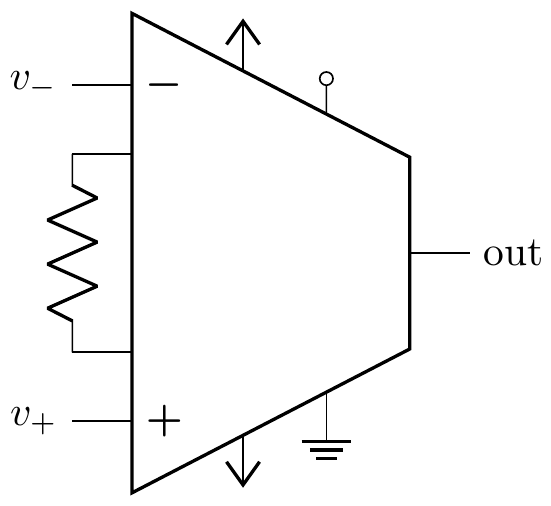}
\end{minipage}%
\hfill
\begin{minipage}[t]{\casecenterwidth}
  \textbf{A:}\\[-2pt]
  \begin{lstlisting}[style=tinycode]
% [3 packages: fontenc, inputenc, circuitikz]
\begin{circuitikz}
\draw
(0,0) node[inst amp ra] (opamp) {}
(opamp.+) node[left] {$v_+$}
(opamp.-) node[left] {$v_-$}
(opamp.out) node[right] {out}
(opamp.up) node[vcc]{}
(opamp.down) node[vee] {}
(opamp.refv down) node[ground]{}
(opamp.refv up) to[short, -o] ++(0,0.3)
(opamp.ra-) to[R] (opamp.ra+);
\end{circuitikz}
  \end{lstlisting}
\end{minipage}%
\hfill
\begin{minipage}[t]{0.24\linewidth}
  {\footnotesize\textbf{Rendered Image:}}\\[3pt]
  \centering
  \includegraphics[width=\linewidth,height=3cm,keepaspectratio]{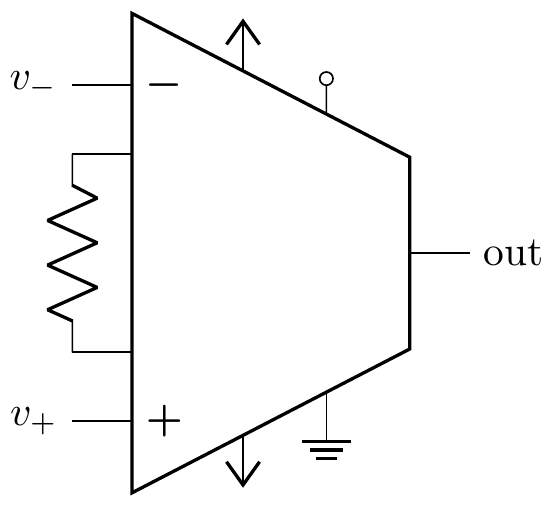}
\end{minipage}
\end{tcolorbox}%
\begin{tcolorbox}[casebox, colback=d2c-bg, colframe=d2c-header!50]
\textbf{\footnotesize Circuit Diagrams Case 2}\\[2pt]
\begin{minipage}[t]{\caseleftwidth}
  \caseshowcasequestion{Recreate this diagram using \LaTeX. Provide the full, executable code.}
  \centering
  \includegraphics[width=\linewidth,height=3cm,keepaspectratio]{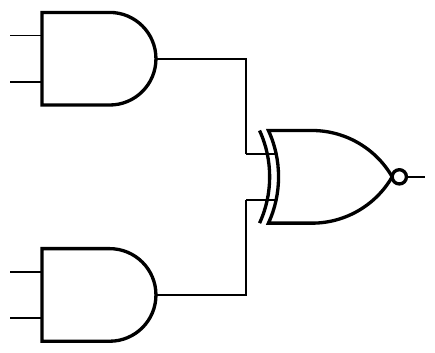}
\end{minipage}%
\hfill
\begin{minipage}[t]{\casecenterwidth}
  \textbf{A:}\\[-2pt]
  \begin{lstlisting}[style=tinycode]
% [4 packages: fontenc, inputenc, lmodern...]
\begin{circuitikz}
\draw
(0,2) node[and port] (myand1) {}
(0,0) node[and port] (myand2) {}
(2,1) node[xnor port] (myxnor) {}
(myand1.out) -| (myxnor.in 1)
(myand2.out) -| (myxnor.in 2);
\end{circuitikz}
  \end{lstlisting}
\end{minipage}%
\hfill
\begin{minipage}[t]{0.24\linewidth}
  {\footnotesize\textbf{Rendered Image:}}\\[3pt]
  \centering
  \includegraphics[width=\linewidth,height=3cm,keepaspectratio]{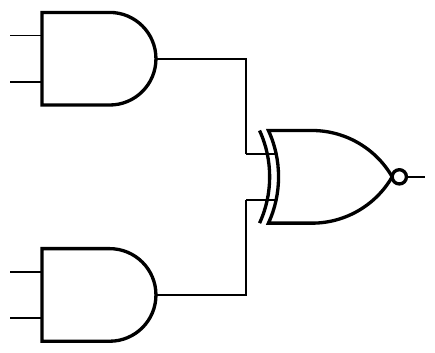}
\end{minipage}
\end{tcolorbox}%
\begin{tcolorbox}[casebox, colback=d2c-bg, colframe=d2c-header!50]
\textbf{\footnotesize Chemistry Case 1}\\[2pt]
\begin{minipage}[t]{\caseleftwidth}
  \caseshowcasequestion{Recreate this diagram using \LaTeX. Provide the full, executable code.}
  \centering
  \includegraphics[width=\linewidth,height=3cm,keepaspectratio]{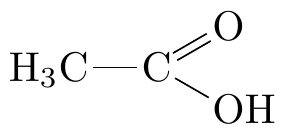}
\end{minipage}%
\hfill
\begin{minipage}[t]{\casecenterwidth}
  \textbf{A:}\\[-2pt]
  \begin{lstlisting}[style=tinycode]
% [1 packages: chemfig]
\chemfig{H_3C-C(=[:30]O)-[:-30]OH}
\vflipnext
\chemfig{H_3C-C(=[:30]O)-[:-30]OH}\medskip
\chemfig{H_3C-C(=[:30]O)-[:-30]OH}
\hflipnext
\chemfig{H_3C-C(=[:30]O)-[:-30]OH}
  \end{lstlisting}
\end{minipage}%
\hfill
\begin{minipage}[t]{0.24\linewidth}
  {\footnotesize\textbf{Rendered Image:}}\\[3pt]
  \centering
  \includegraphics[width=\linewidth,height=3cm,keepaspectratio]{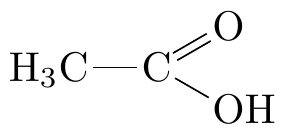}
\end{minipage}
\end{tcolorbox}%
\begin{tcolorbox}[casebox, colback=d2c-bg, colframe=d2c-header!50]
\textbf{\footnotesize Chemistry Case 2}\\[2pt]
\begin{minipage}[t]{\caseleftwidth}
  \caseshowcasequestion{Recreate this diagram using \LaTeX. Provide the full, executable code.}
  \centering
  \includegraphics[width=\linewidth,height=3cm,keepaspectratio]{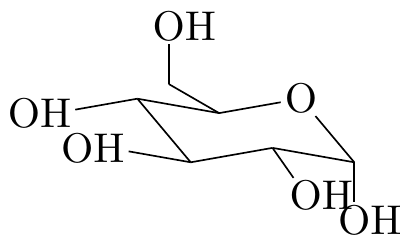}
\end{minipage}%
\hfill
\begin{minipage}[t]{\casecenterwidth}
  \textbf{A:}\\[-2pt]
  \begin{lstlisting}[style=tinycode]
% [3 packages: amsmath,amssymb, xcolor, chemfig]
\chemfig{?(-[:190]OH)-[:-50](-[:170]OH)-[:10](-[:-55,0.7]OH)
-[:-10](-[6,0.7]OH)-[:130]O-[:190]?(-[:150,0.7]-[2,0.7]OH)}
  \end{lstlisting}
\end{minipage}%
\hfill
\begin{minipage}[t]{0.24\linewidth}
  {\footnotesize\textbf{Rendered Image:}}\\[3pt]
  \centering
  \includegraphics[width=\linewidth,height=3cm,keepaspectratio]{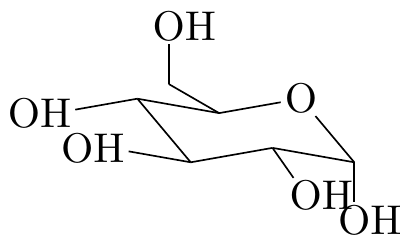}
\end{minipage}
\end{tcolorbox}
\captionof{figure}{Representative D2C-P (Diagram-to-Code Parsing) examples. Given an input diagram, the task requires generating complete, compilable \LaTeX{}/TikZ code that faithfully reproduces the original.}
\label{fig:appx_d2cp_showcase}

\subsection{D2C-E Templates}
\label{sec:appx_d2ce_templates}

The D2C-E employs a pool of \textbf{17 editing task templates} across four evaluation dimensions: \emph{color} (C, 2~templates), \emph{text} (T, 4~templates), \emph{scope} (S, 8~templates), and \emph{layout} (L, 3~templates). Each template specifies a natural-language editing instruction with placeholders (in square brackets) that are instantiated with diagram-specific elements during the agentic pipeline (\S\ref{sec:appx_pipeline}). Table~\ref{tab:appx_d2ce_pool} presents the complete template pool, and Table~\ref{tab:appx_d2ce_apply} shows per-domain applicability.

\vspace{0.3em}

\begin{table}[!tbp]
\centering
\caption{Eediting template pool of D2C-E. 17 templates span four evaluation dimensions. Placeholders in brackets are instantiated with diagram-specific elements during generation.}
\label{tab:appx_d2ce_pool}
\renewcommand{\arraystretch}{1.1}
\setlength{\tabcolsep}{4.mm}{
\resizebox{\linewidth}{!}{%
\begin{tabular}{c l p{10cm}}
\toprule
\textbf{ID} & \textbf{Task Name} & \textbf{Template} \\
\midrule
\rowcolor{softgray} \multicolumn{3}{l}{\textbf{Color}} \\
C1 & Element Color Change & Change the [\textit{color\_attr}] of [\textit{target\_element}] to [\textit{new\_color}]. \\
C2 & Text Color Change & Change the text color of [\textit{target\_text}] to [\textit{new\_color}]. \\
\rowcolor{softgray} \multicolumn{3}{l}{\textbf{Text}} \\
T1 & Label/Annotation Rename & Rename the label [\textit{old\_text}] to [\textit{new\_text}]. \\
T2 & Data Value Modification & Change the data value at [\textit{position}] from [\textit{old\_value}] to [\textit{new\_value}]. \\
T3 & Component Parameter Mod. & Change the parameter of [\textit{component}] from [\textit{old\_param}] to [\textit{new\_param}]. \\
T4 & Title/Legend Text Mod. & Change the [\textit{text\_type}] text from [\textit{old\_text}] to [\textit{new\_text}]. \\
\rowcolor{softgray} \multicolumn{3}{l}{\textbf{Scope}} \\
S1 & Element Position Move & Move [\textit{target\_element}] to [\textit{new\_position}]. \\
S2 & Rotation/Mirror Transform & Rotate [\textit{target}] by [\textit{degrees}]$^\circ$ [\textit{direction}]. / Mirror [\textit{target}] along the [\textit{axis}] axis. \\
S3 & Axis Range Scaling & Change the [\textit{axis}] axis range from [\textit{old\_range}] to [\textit{new\_range}]. \\
S4 & Element Position Swap & Swap the positions of [\textit{element\_A}] and [\textit{element\_B}]. \\
S5 & Element Deletion & Remove [\textit{target\_element}] from the diagram. \\
S6 & Element Addition & Add [\textit{new\_element}] to the diagram. \\
S7 & Node Shape Replacement & Change the shape of [\textit{target\_node}] from [\textit{old\_shape}] to [\textit{new\_shape}]. \\
S8 & Component Type Replacement & Replace [\textit{old\_component}] with [\textit{new\_component\_type}]. \\
\rowcolor{softgray} \multicolumn{3}{l}{\textbf{Layout}} \\
L1 & Structure Type Conversion & Convert [\textit{current\_type}] to [\textit{target\_type}]. \\
L2 & Conditional Filtering & Filter the data to only show [\textit{condition}]. \\
L3 & Reference Element Addition & Add [\textit{reference\_element}] to the diagram. \\
\bottomrule
\end{tabular}}
}
\end{table}

\begin{table}[!tbp]
\centering
\caption{Editing dimensions for each type of diagram. {\cmark} indicates the dimension applies to that domain.}
\label{tab:appx_d2ce_apply}
\renewcommand{\arraystretch}{1.1}
\setlength{\tabcolsep}{4.mm}{
\resizebox{0.8\linewidth}{!}{%
\begin{tabular}{lccccccc}
\toprule
\textbf{Task} & \textbf{Charts} & \textbf{P.\,Geom.} & \textbf{3D} & \textbf{Graph} & \textbf{Chem.} & \textbf{Circuit} & \textbf{\#} \\
\midrule
C1 & \cmark & \cmark & \cmark & \cmark & \cmark & \cmark & 6 \\
C2 & \cmark & \cmark & \cmark & \cmark & \cmark & \cmark & 6 \\
\midrule
T1 & \cmark & \cmark & \cmark & \cmark & \cmark & \cmark & 6 \\
T2 & \cmark & & & & & & 1 \\
T3 & & & & & & \cmark & 1 \\
T4 & \cmark & & & & & & 1 \\
\midrule
S1 & \cmark & \cmark & & \cmark & & \cmark & 4 \\
S2 & & \cmark & \cmark & \cmark & \cmark & & 4 \\
S3 & \cmark & & & & & & 1 \\
S4 & & \cmark & & \cmark & & \cmark & 3 \\
S5 & \cmark & \cmark & \cmark & \cmark & \cmark & \cmark & 6 \\
S6 & \cmark & \cmark & \cmark & \cmark & \cmark & \cmark & 6 \\
S7 & & & & \cmark & & & 1 \\
S8 & & & & & & \cmark & 1 \\
\midrule
L1 & \cmark & & & \cmark & & \cmark & 3 \\
L2 & \cmark & & & \cmark & & \cmark & 3 \\
L3 & \cmark & \cmark & & & & & 2 \\
\midrule
\textbf{Per domain} & \textbf{12} & \textbf{9} & \textbf{6} & \textbf{11} & \textbf{6} & \textbf{11} & \\
\bottomrule
\end{tabular}}
}
\end{table}

\subsection{D2C-E Examples}
\label{sec:appx_d2ce_examples}

Figure~\ref{fig:appx_d2ce_showcase} presents representative D2C-E cases. Each case shows the original diagram, an editing instruction, the modified TikZ code, and the rendered output after editing.

\casegroupheader{d2e-header}{Diagram-to-Code Editing (D2C-E)}

\begin{tcolorbox}[casebox, colback=d2e-bg, colframe=d2e-header!50]
\textbf{\footnotesize Charts Case 1}\\[2pt]
\begin{minipage}[t]{\caseleftwidth}
  \caseshowcasequestion{Convert this grouped bar chart into a line chart, where each series in the legend is represented by a separate line connecting the data points for each agent.}
  \centering
  \includegraphics[width=\linewidth,height=3cm,keepaspectratio]{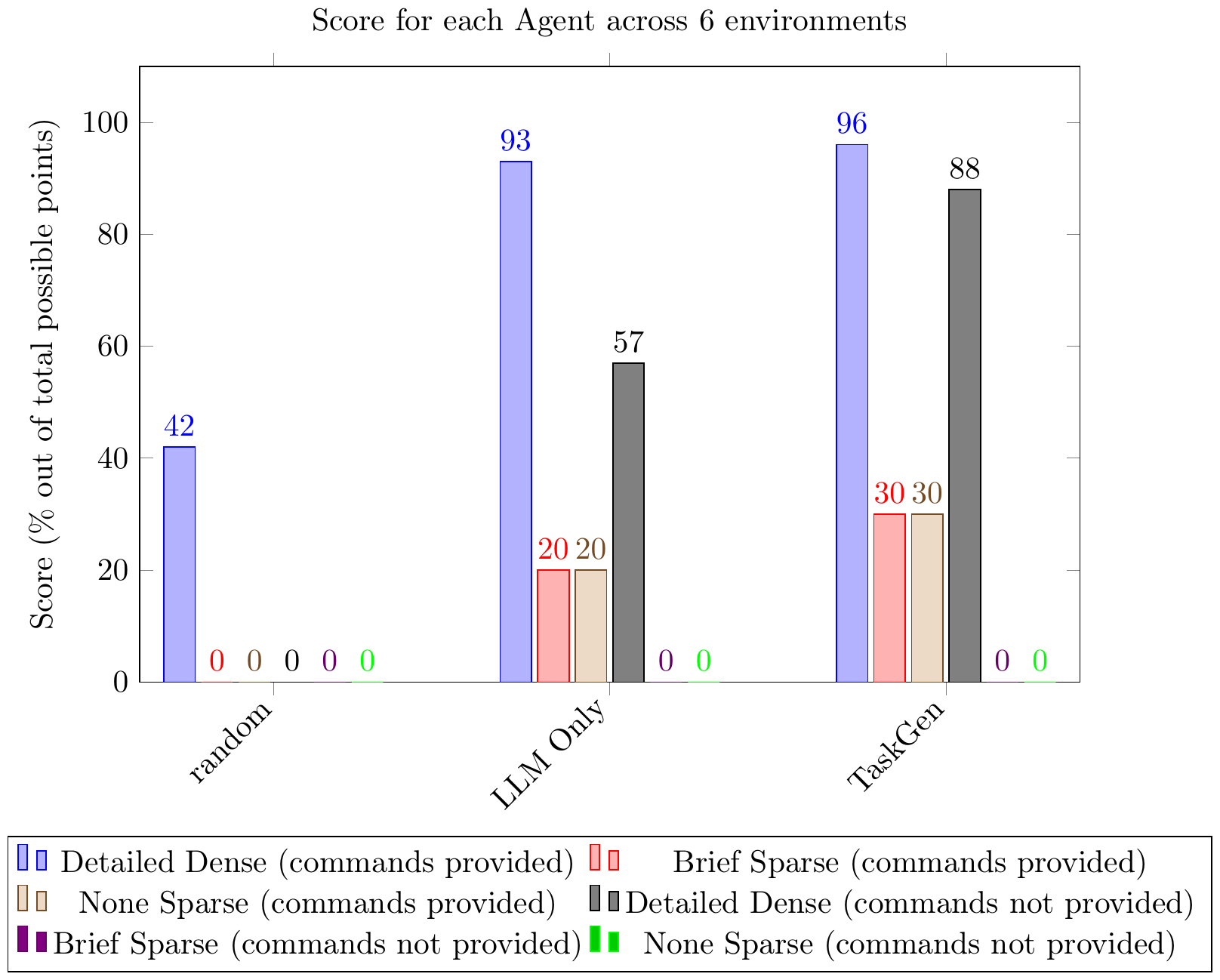}
\end{minipage}%
\hfill
\begin{minipage}[t]{\casecenterwidth}
  \textbf{A:}\\[-2pt]
  \begin{lstlisting}[style=tinycode]
% [5 packages: fontenc, inputenc, xcolor...]
  \begin{tikzpicture}
    \begin{axis}[width=\textwidth, height=0.7\textwidth, xlabel={Agent}, ylabel={Score (\% out of total possible points)}, symbolic x coords={random, LLM Only, TaskGen}, xtick=data, nodes near coords, ymin=0, ymax=110, legend style={at={(0.5,-0.25)}, anchor=north, legend columns=2}, enlarge x limits={abs=1.5cm}, ylabel near ticks, xticklabel style={rotate=45, anchor=east}, title={Score for each Agent across 6 environments}]
      \addplot coordinates {(random,42) (LLM Only,93) (TaskGen,96)};
      \addplot coordinates {(random,0) (LLM Only,20) (TaskGen,30)};
      \addplot coordinates {(random,0) (LLM Only, 20) (TaskGen,30)};
      \addplot coordinates {(random,0) (LLM Only,57) (TaskGen,88)};
      \addplot coordinates {(random,0) (LLM Only,0) (TaskGen,0)};
      \addplot coordinates {(random,0) (LLM Only,0) (TaskGen,0)};
      \legend{Detailed Dense (commands provided), Brief Sparse (commands provided), None Sparse (commands provided), Detailed Dense (commands not provided), Brief Sparse (commands not provided), None Sparse (commands not provided)}
    \end{axis}
  \end{tikzpicture}
  \end{lstlisting}
\end{minipage}%
\hfill
\begin{minipage}[t]{0.24\linewidth}
  {\footnotesize\textbf{Rendered Image:}}\\[3pt]
  \centering
  \includegraphics[width=\linewidth,height=3cm,keepaspectratio]{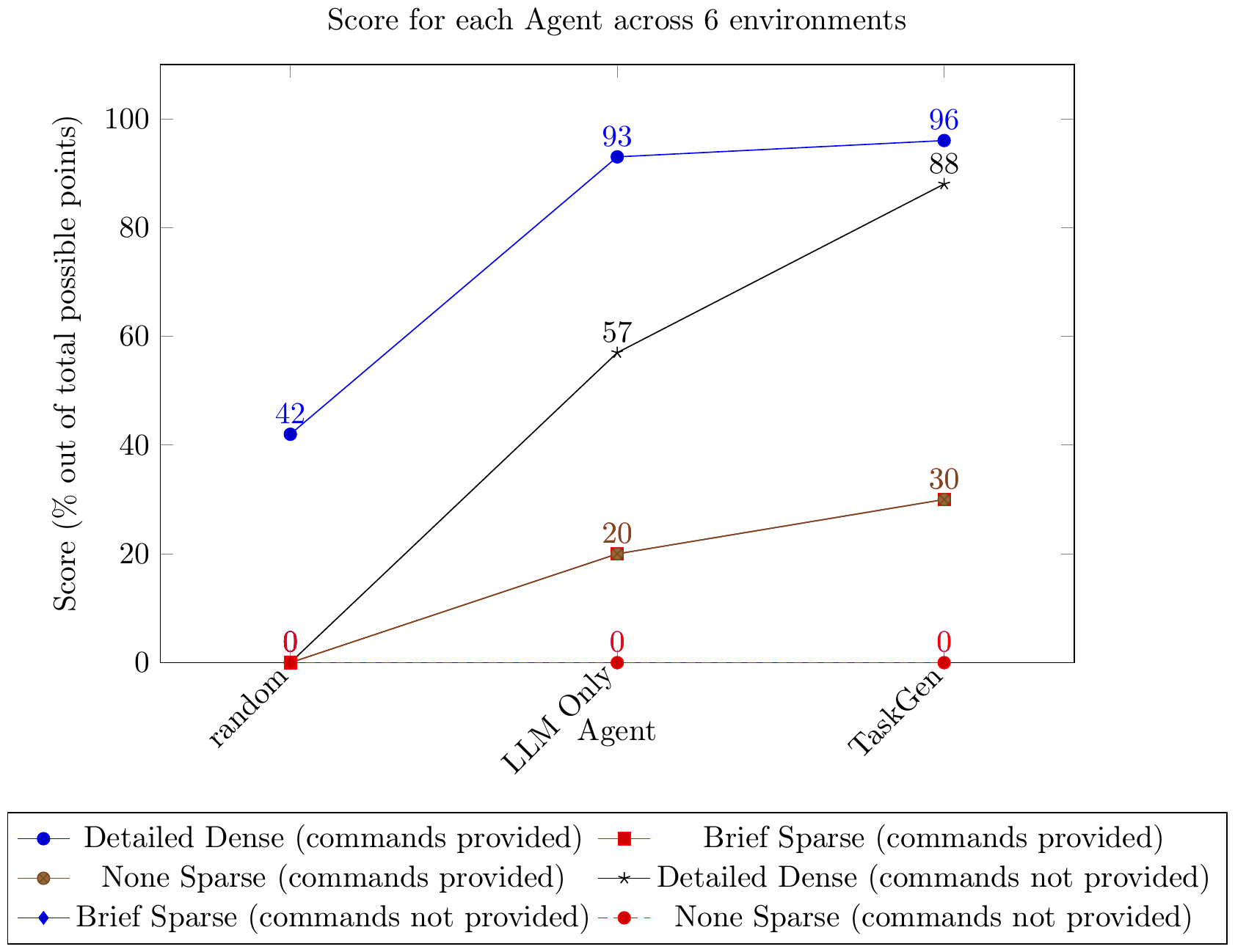}
\end{minipage}
\end{tcolorbox}%
\begin{tcolorbox}[casebox, colback=d2e-bg, colframe=d2e-header!50]
\textbf{\footnotesize Charts Case 2}\\[2pt]
\begin{minipage}[t]{\caseleftwidth}
  \caseshowcasequestion{Change the data value at the third data point of the Benchmark System from (150, 18.73) to (150, 22.00).}
  \centering
  \includegraphics[width=\linewidth,height=3cm,keepaspectratio]{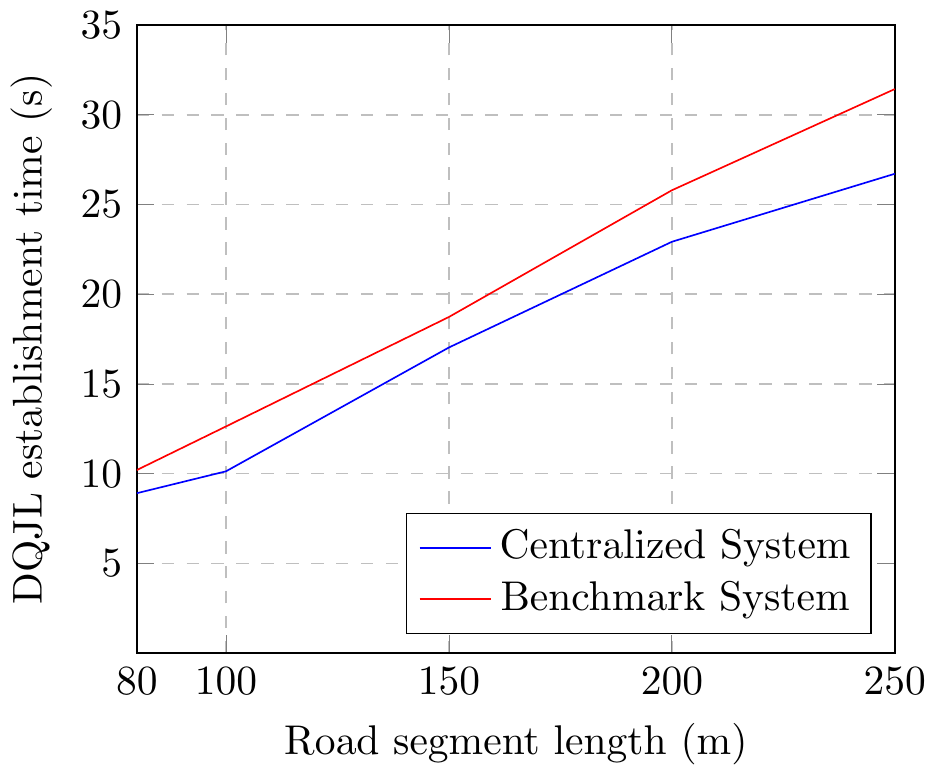}
\end{minipage}%
\hfill
\begin{minipage}[t]{\casecenterwidth}
  \textbf{A:}\\[-2pt]
  \begin{lstlisting}[style=tinycode]
% [6 packages: pgfplots, amsmath, amssymb...]
  \begin{tikzpicture}
    \begin{axis}[xlabel={Road segment length (m)},ylabel={DQJL establishment time (s)},xmin=80,xmax=250,ymin=0,ymax=35,xtick={80,100,150,200,250},ytick={5, 10, 15, 20, 25, 30, 35},legend pos=south east,ymajorgrids=true,xmajorgrids=true,grid style=dashed]\addplot[color=blue]coordinates{(80, 8.91)(100, 10.13)(150, 17.03)(200, 22.92)(250, 26.71)};
      \addplot coordinates{(80, 10.2)(100, 12.63)(150, 22.00)(200, 25.79)(250, 31.43)};
      \legend{Centralized System, Benchmark System}
    \end{axis}
  \end{tikzpicture}
  \end{lstlisting}
\end{minipage}%
\hfill
\begin{minipage}[t]{0.24\linewidth}
  {\footnotesize\textbf{Rendered Image:}}\\[3pt]
  \centering
  \includegraphics[width=\linewidth,height=3cm,keepaspectratio]{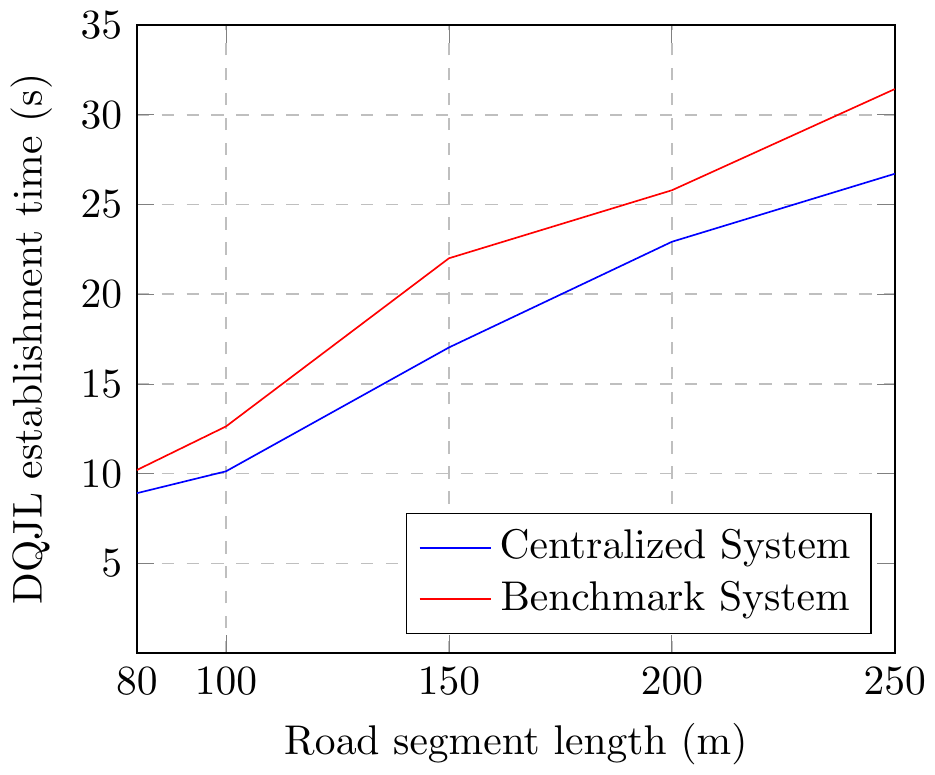}
\end{minipage}
\end{tcolorbox}%
\begin{tcolorbox}[casebox, colback=d2e-bg, colframe=d2e-header!50]
\textbf{\footnotesize Graph Structures Case 1}\\[2pt]
\begin{minipage}[t]{\caseleftwidth}
  \caseshowcasequestion{Convert the current hexagonal cycle layout into a linear vertical layout where the nodes are connected in sequence from 6 at the top to 1 at the bottom.}
  \centering
  \includegraphics[width=\linewidth,height=3cm,keepaspectratio]{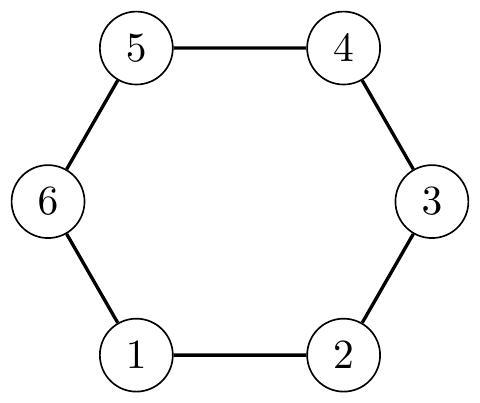}
\end{minipage}%
\hfill
\begin{minipage}[t]{\casecenterwidth}
  \textbf{A:}\\[-2pt]
  \begin{lstlisting}[style=tinycode]
% [3 packages: amsmath, amssymb, tikz,color]
  \begin{tikzpicture}[scale=0.65]
    \node [draw, circle] (1) at (0,0) {$1$};
    \node [draw, circle] (2) at (0,1.5) {$2$};
    \node [draw, circle] (3) at (0,3) {$3$};
    \node [draw, circle] (4) at (0,4.5) {$4$};
    \node [draw, circle] (5) at (0,6) {$5$};
    \node [draw, circle] (6) at (0,7.5) {$6$};
    \draw [thick] (6)--(5)--(4)--(3)--(2)--(1);
  \end{tikzpicture}
  \end{lstlisting}
\end{minipage}%
\hfill
\begin{minipage}[t]{0.24\linewidth}
  {\footnotesize\textbf{Rendered Image:}}\\[3pt]
  \centering
  \includegraphics[width=\linewidth,height=3cm,keepaspectratio]{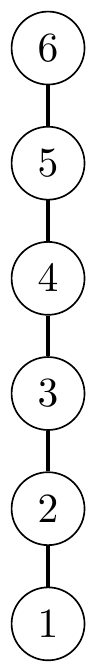}
\end{minipage}
\end{tcolorbox}%
\begin{tcolorbox}[casebox, colback=d2e-bg, colframe=d2e-header!50]
\textbf{\footnotesize Graph Structures Case 2}\\[2pt]
\begin{minipage}[t]{\caseleftwidth}
  \caseshowcasequestion{Change the shape of node 6 from a circle to a rectangle.}
  \centering
  \includegraphics[width=\linewidth,height=3cm,keepaspectratio]{figures/appendix/cases/scidiagram_6198.png}
\end{minipage}%
\hfill
\begin{minipage}[t]{\casecenterwidth}
  \textbf{A:}\\[-2pt]
  \begin{lstlisting}[style=tinycode]
% [3 packages: amsmath, amssymb, tikz,color]
  \begin{tikzpicture}[scale=0.65]
    \node [draw, circle] (1) at (-1.35,0) {$1$};
    \node [draw, circle] (2) at (1.35,0) {$2$};
    \node [draw, circle] (3) at (2.5,2) {$3$};
    \node [draw, circle] (4) at (1.35,4) {$4$};
    \node [draw, circle] (5) at (-1.35,4) {$5$};
    \node [draw, rectangle] (6) at (-2.5,2) {$6$};
    \draw [thick] (1)--(2)--(3)--(4)--(5)--(6)--(1);
  \end{tikzpicture}
  \end{lstlisting}
\end{minipage}%
\hfill
\begin{minipage}[t]{0.24\linewidth}
  {\footnotesize\textbf{Rendered Image:}}\\[3pt]
  \centering
  \includegraphics[width=\linewidth,height=3cm,keepaspectratio]{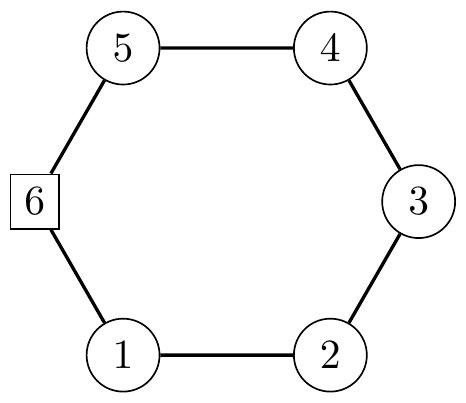}
\end{minipage}
\end{tcolorbox}%
\begin{tcolorbox}[casebox, colback=d2e-bg, colframe=d2e-header!50]
\textbf{\footnotesize Planar Geometry Case 1}\\[2pt]
\begin{minipage}[t]{\caseleftwidth}
  \caseshowcasequestion{Add a vertical diameter passing through the center point O to the diagram.}
  \centering
  \includegraphics[width=\linewidth,height=3cm,keepaspectratio]{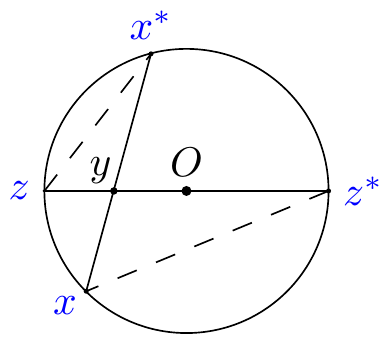}
\end{minipage}%
\hfill
\begin{minipage}[t]{\casecenterwidth}
  \textbf{A:}\\[-2pt]
  \begin{lstlisting}[style=tinycode]
% [1 packages: lmodern,tikz]
  \begin{tikzpicture}[scale=.3]
    \draw(0,0)circle(4.01cm);
    \draw(-1,3.87)--(-2.83,-2.83);
    \draw(4.01,0)--(-4.01,0);
    \draw(0,4.01)--(0,-4.01);
    \draw[dash pattern=on 5pt off 5pt](-4.01,0)--(-1,3.87);
    \draw[dash pattern=on 5pt off 5pt](4.01,0)--(-2.83,-2.83);
  ...
    \fill(4.01,0)circle(2pt)node[blue,right]{$z^*$};
    \fill(-4.01,0)circle(1.5pt)node[blue,left]{$z$};
    \fill(-2.05,0)circle(3pt);
    \node[above left]at(-1.7,-.2){$y$};
    \fill(0,0)circle(4pt)node[above]{$O$};
  \end{tikzpicture}
  \end{lstlisting}
\end{minipage}%
\hfill
\begin{minipage}[t]{0.24\linewidth}
  {\footnotesize\textbf{Rendered Image:}}\\[3pt]
  \centering
  \includegraphics[width=\linewidth,height=3cm,keepaspectratio]{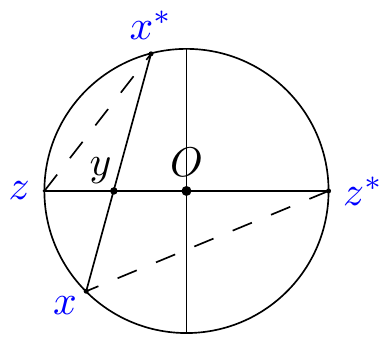}
\end{minipage}
\end{tcolorbox}%
\begin{tcolorbox}[casebox, colback=d2e-bg, colframe=d2e-header!50]
\textbf{\footnotesize Planar Geometry Case 2}\\[2pt]
\begin{minipage}[t]{\caseleftwidth}
  \caseshowcasequestion{Change the stroke color of the red path connecting the tangents and the circle center to blue.}
  \centering
  \includegraphics[width=\linewidth,height=3cm,keepaspectratio]{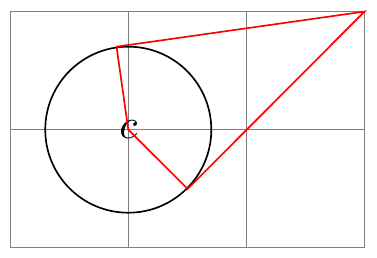}
\end{minipage}%
\hfill
\begin{minipage}[t]{\casecenterwidth}
  \textbf{A:}\\[-2pt]
  \begin{lstlisting}[style=tinycode]
% [1 packages: tikz]
  \begin{tikzpicture}
    \draw[help lines] (0,0) grid (3,2);
    \coordinate (a) at (3,2);
    \node [circle,draw] (c) at (1,1) [minimum size=40pt] {$c$};
    \draw[blue] (a)  -- (tangent cs:node=c,point={(a)},solution=1) --
    (c.center) -- (tangent cs:node=c,point={(a)},solution=2) -- cycle;
  \end{tikzpicture}
  \end{lstlisting}
\end{minipage}%
\hfill
\begin{minipage}[t]{0.24\linewidth}
  {\footnotesize\textbf{Rendered Image:}}\\[3pt]
  \centering
  \includegraphics[width=\linewidth,height=3cm,keepaspectratio]{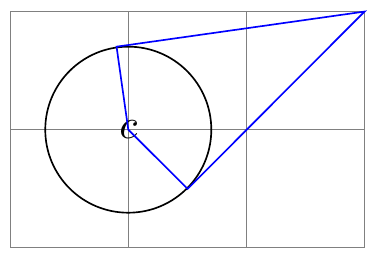}
\end{minipage}
\end{tcolorbox}%
\begin{tcolorbox}[casebox, colback=d2e-bg, colframe=d2e-header!50]
\textbf{\footnotesize 3D Shapes Case 1}\\[2pt]
\begin{minipage}[t]{\caseleftwidth}
  \caseshowcasequestion{Rotate the 3D surface plot by 90 degrees around the vertical axis.}
  \centering
  \includegraphics[width=\linewidth,height=3cm,keepaspectratio]{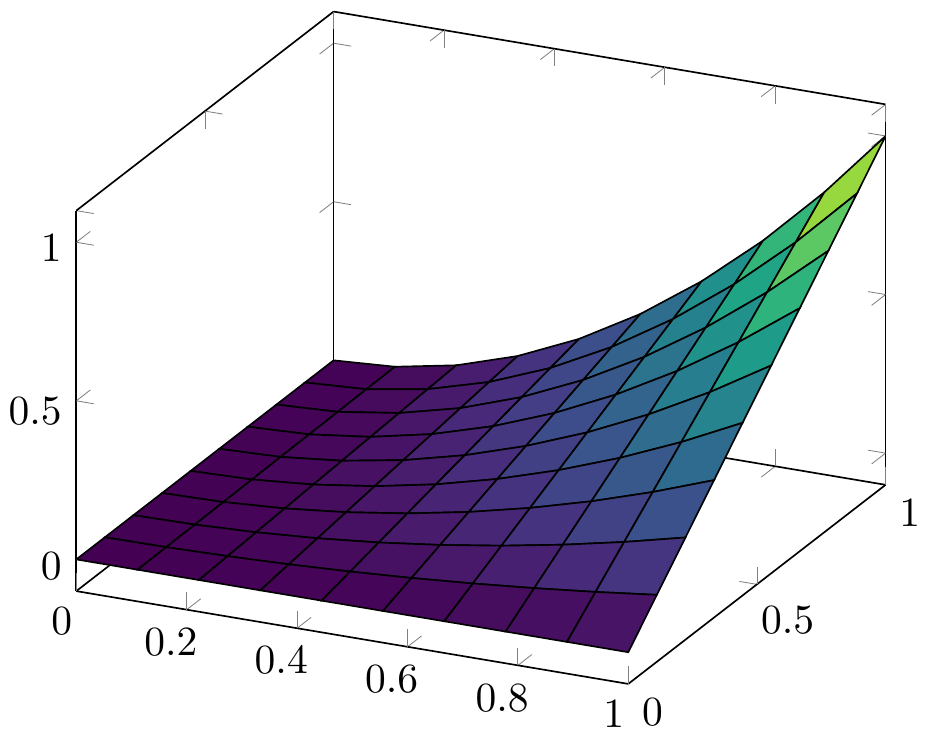}
\end{minipage}%
\hfill
\begin{minipage}[t]{\casecenterwidth}
  \textbf{A:}\\[-2pt]
  \begin{lstlisting}[style=tinycode]
% [1 packages: pgfplots]
  \begin{tikzpicture}
    \begin{axis}[colormap/viridis, view={115}{30}]
      \addplot3 [
      surf,
      shader=flat,
      draw=black,
      samples=10,
      domain=0:1,
      ] {x^2*y};
    \end{axis}
  \end{tikzpicture}
  \end{lstlisting}
\end{minipage}%
\hfill
\begin{minipage}[t]{0.24\linewidth}
  {\footnotesize\textbf{Rendered Image:}}\\[3pt]
  \centering
  \includegraphics[width=\linewidth,height=3cm,keepaspectratio]{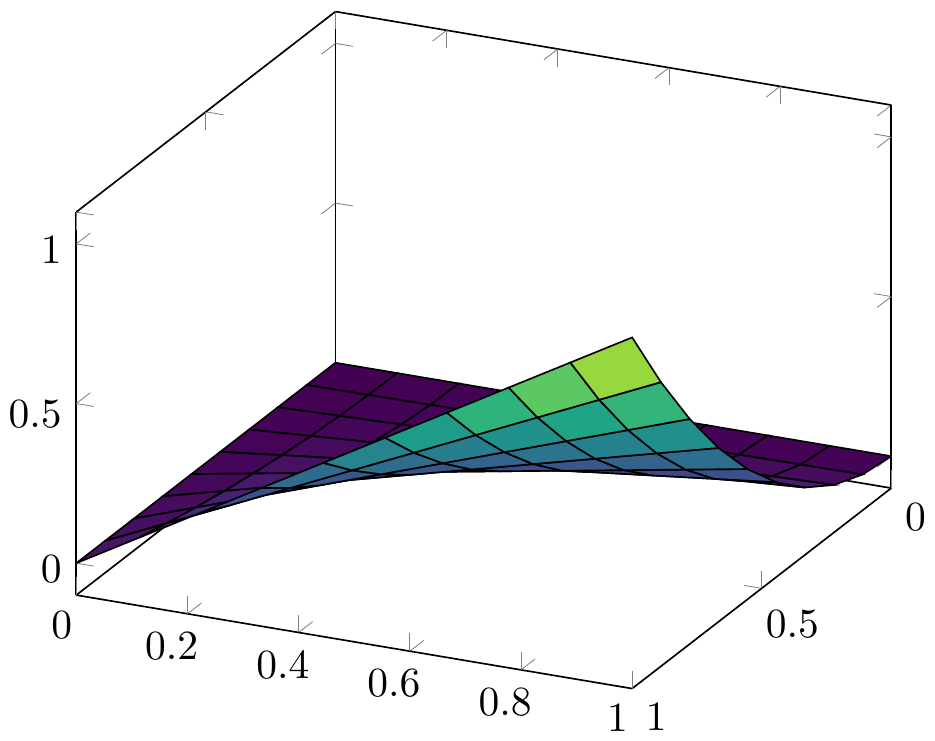}
\end{minipage}
\end{tcolorbox}%
\begin{tcolorbox}[casebox, colback=d2e-bg, colframe=d2e-header!50]
\textbf{\footnotesize 3D Shapes Case 2}\\[2pt]
\begin{minipage}[t]{\caseleftwidth}
  \caseshowcasequestion{Change the interior colormap of the 3D surface from 'hot' to 'plasma'.}
  \centering
  \includegraphics[width=\linewidth,height=3cm,keepaspectratio]{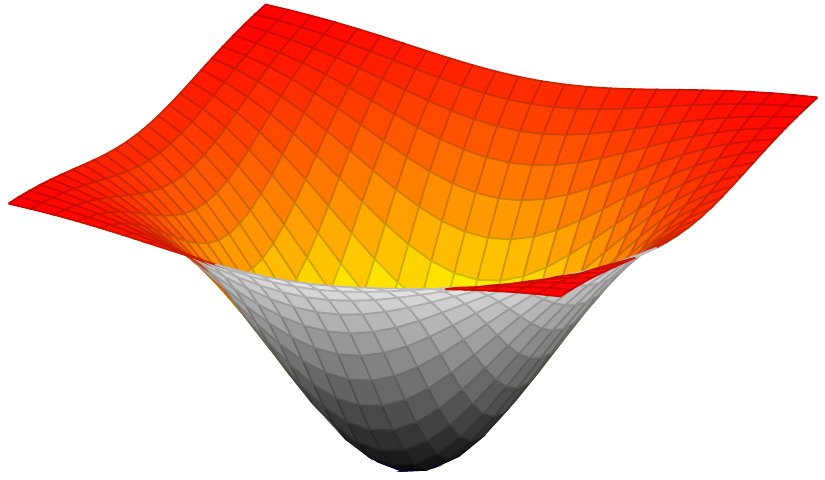}
\end{minipage}%
\hfill
\begin{minipage}[t]{\casecenterwidth}
  \textbf{A:}\\[-2pt]
  \begin{lstlisting}[style=tinycode]
% [1 packages: pgfplots]
  \begin{tikzpicture}
    \begin{axis}[
      hide axis,
      xlabel=$x$,ylabel=$y$,
      mesh/interior colormap name=plasma,
      colormap/blackwhite,
      ]
      \addplot3 [domain=-1.5:1.5,surf]
      {-exp(-x^2-y^2)};
    \end{axis}
  \end{tikzpicture}
  \end{lstlisting}
\end{minipage}%
\hfill
\begin{minipage}[t]{0.24\linewidth}
  {\footnotesize\textbf{Rendered Image:}}\\[3pt]
  \centering
  \includegraphics[width=\linewidth,height=3cm,keepaspectratio]{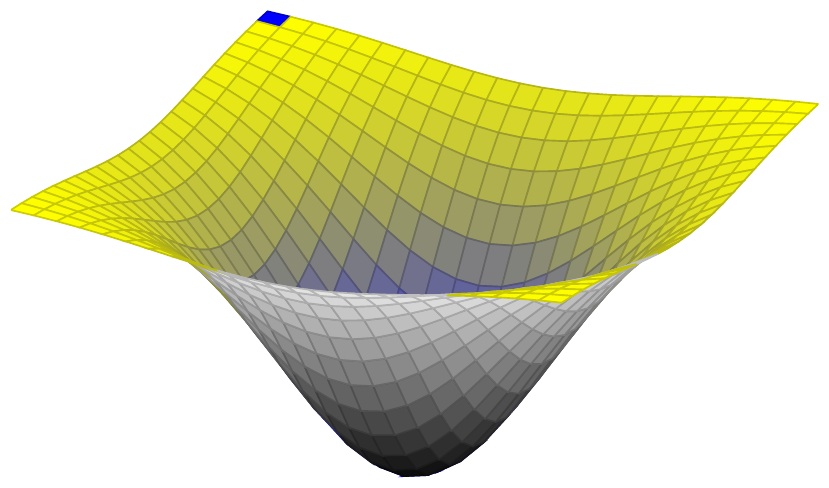}
\end{minipage}
\end{tcolorbox}%
\begin{tcolorbox}[casebox, colback=d2e-bg, colframe=d2e-header!50]
\textbf{\footnotesize Circuit Diagrams Case 1}\\[2pt]
\begin{minipage}[t]{\caseleftwidth}
  \caseshowcasequestion{Replace the resistor connected between the ra- and ra+ pins with a capacitor.}
  \centering
  \includegraphics[width=\linewidth,height=3cm,keepaspectratio]{figures/appendix/cases/scidiagram_1622.png}
\end{minipage}%
\hfill
\begin{minipage}[t]{\casecenterwidth}
  \textbf{A:}\\[-2pt]
  \begin{lstlisting}[style=tinycode]
% [3 packages: fontenc, inputenc, circuitikz]
  \begin{circuitikz}
    \draw
    (0,0) node[inst amp ra] (opamp) {}
    (opamp.+) node[left] {$v_+$}
    (opamp.-) node[left] {$v_-$}
    (opamp.out) node[right] {out}
    (opamp.up) node[vcc]{}
    (opamp.down) node[vee] {}
    (opamp.refv down) node[ground]{}
    (opamp.refv up) to[short, -o] ++(0,0.3)
    (opamp.ra-) to[C] (opamp.ra+);
  \end{circuitikz}
  \end{lstlisting}
\end{minipage}%
\hfill
\begin{minipage}[t]{0.24\linewidth}
  {\footnotesize\textbf{Rendered Image:}}\\[3pt]
  \centering
  \includegraphics[width=\linewidth,height=3cm,keepaspectratio]{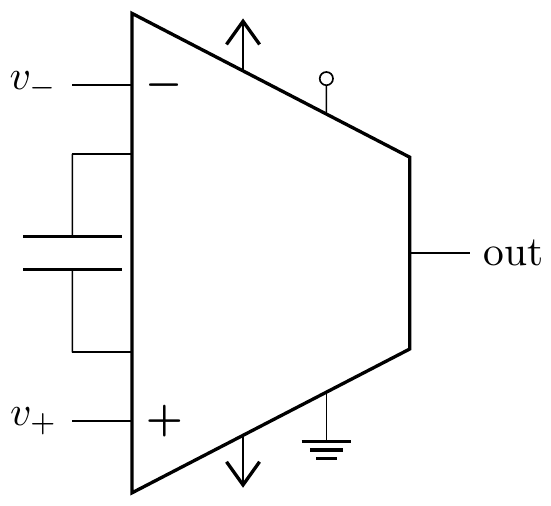}
\end{minipage}
\end{tcolorbox}%
\begin{tcolorbox}[casebox, colback=d2e-bg, colframe=d2e-header!50]
\textbf{\footnotesize Circuit Diagrams Case 2}\\[2pt]
\begin{minipage}[t]{\caseleftwidth}
  \caseshowcasequestion{Change the parameter of the voltage source from 10V to 5V.}
  \centering
  \includegraphics[width=\linewidth,height=3cm,keepaspectratio]{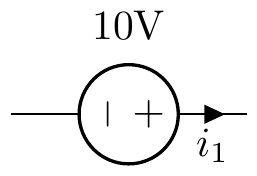}
\end{minipage}%
\hfill
\begin{minipage}[t]{\casecenterwidth}
  \textbf{A:}\\[-2pt]
  \begin{lstlisting}[style=tinycode]
% [4 packages: fontenc, inputenc, lmodern...]
  \begin{circuitikz}[american]
    % Standard independent voltage source with current arrow
    \draw (0,0) to[V=5V, i_=$i_1$] (2,0);
  \end{circuitikz}
  \end{lstlisting}
\end{minipage}%
\hfill
\begin{minipage}[t]{0.24\linewidth}
  {\footnotesize\textbf{Rendered Image:}}\\[3pt]
  \centering
  \includegraphics[width=\linewidth,height=3cm,keepaspectratio]{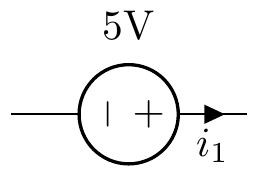}
\end{minipage}
\end{tcolorbox}%
\begin{tcolorbox}[casebox, colback=d2e-bg, colframe=d2e-header!50]
\textbf{\footnotesize Chemistry Case 1}\\[2pt]
\begin{minipage}[t]{\caseleftwidth}
  \caseshowcasequestion{Add a single bond and a methyl group (-CH3) attached to 'C\_3\textasciicircum{}4' to the diagram.}
  \centering
  \includegraphics[width=\linewidth,height=3cm,keepaspectratio]{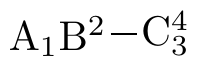}
\end{minipage}%
\hfill
\begin{minipage}[t]{\casecenterwidth}
  \textbf{A:}\\[-2pt]
  \begin{lstlisting}[style=tinycode]
% [1 packages: chemfig]
  \chemfig{A_1B^2-C _ 3 ^ 4-CH_3}
  \end{lstlisting}
\end{minipage}%
\hfill
\begin{minipage}[t]{0.24\linewidth}
  {\footnotesize\textbf{Rendered Image:}}\\[3pt]
  \centering
  \includegraphics[width=\linewidth,height=3cm,keepaspectratio]{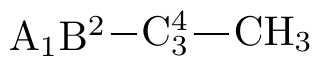}
\end{minipage}
\end{tcolorbox}%
\begin{tcolorbox}[casebox, colback=d2e-bg, colframe=d2e-header!50]
\textbf{\footnotesize Chemistry Case 2}\\[2pt]
\begin{minipage}[t]{\caseleftwidth}
  \caseshowcasequestion{Change the stroke color of the benzene ring to green.}
  \centering
  \includegraphics[width=\linewidth,height=2.2cm,keepaspectratio]{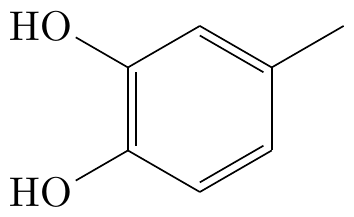}
\end{minipage}%
\hfill
\begin{minipage}[t]{\casecenterwidth}
  \textbf{A:}\\[-2pt]
  \begin{lstlisting}[style=tinycode]
% [1 packages: chemfig]
\setchemfig{atom sep=2em,bond offset=1pt,bond style={green}}
  \chemfig{*6((-HO)-=-(-)=-(-HO)=)}
  \end{lstlisting}
\end{minipage}%
\hfill
\begin{minipage}[t]{0.24\linewidth}
  {\footnotesize\textbf{Rendered Image:}}\\[3pt]
  \centering
  \includegraphics[width=\linewidth,height=2.2cm,keepaspectratio]{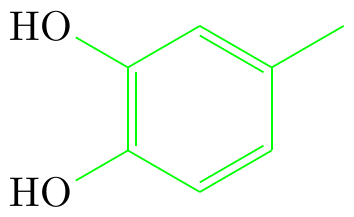}
\end{minipage}
\end{tcolorbox}
\captionof{figure}{Representative D2C-E (Diagram-to-Code Editing) examples. Given an input diagram and an editing instruction, the task requires modifying the TikZ code to produce the desired change while preserving unaffected elements.}
\label{fig:appx_d2ce_showcase}

\subsection{DQA Templates}
\label{sec:appx_dqa_templates}

The DQA uses {60 manually designed question templates} across the six diagram domains, including \emph{descriptive} questions (23~templates) that assess the model's ability to identify domain-specific symbols and extract information; \emph{standard reasoning} questions (18~templates) that require numerical computation with domain-specific formulas; and \emph{what-if reasoning} questions (19~templates) that require predicting answers conditioned on hypothetical element modifications. Each template encodes the relevant domain knowledge. Table~\ref{tab:appx_dqa_tmpl_summary} summarizes the per-domain distribution. We instruct models with the expected output format at inference, and each question specifies one of three output formats:
\begin{itemize}[leftmargin=*,nosep,topsep=2pt]
  \item \textbf{\textsc{OI-NUM}:} A single numerical value with units if applicable (\eg, ``100\,$\Omega$'', ``90\,deg'', ``3.14''). Fallback: \texttt{NOT\_PRESENT} if information is absent from the diagram.
  \item \textbf{\textsc{OI-TERM}:} A standard scientific term or exact label (\eg, ``Parallelogram'', ``Diode'', ``Hydroxyl''). Fallback: \texttt{NOT\_PRESENT}.
  \item \textbf{\textsc{OI-LIST}:} A list of strings sorted alphabetically (\eg, [`Node~A', `Node~C']). Fallback: empty list \texttt{[]}.
\end{itemize}

\begin{center}
\captionof{table}{Question template and instruction counts for DQA across six diagram domains. Templates cover descriptive (Desc.) and reasoning (Reas.) types, where reasoning includes standard (Std.) and what-if sub-types. }
\label{tab:appx_dqa_tmpl_summary}
\renewcommand{\arraystretch}{1.1}
\setlength{\tabcolsep}{4.mm}{
\resizebox{0.8\linewidth}{!}{%
\begin{tabular}{lccccccc}
\toprule
& \multicolumn{3}{c}{\textbf{Template Count}} & & \multicolumn{3}{c}{\textbf{Output Instruction}} \\
\cmidrule(lr){2-4} \cmidrule(lr){6-8}
\textbf{Domain} & \textbf{Desc.} & \textbf{Std./Reas.} & \textbf{What-if/Reas.} & \textbf{Total} & \textbf{NUM} & \textbf{TERM} & \textbf{LIST} \\
\midrule
Charts           & 4 & 4 & 3 & 11 & 9 & 2 & 0 \\
Planar Geom.     & 5 & 3 & 3 & 11 & 6 & 5 & 0 \\
3D Shapes        & 2 & 5 & 4 & 11 & 7 & 4 & 0 \\
Graph Struct.    & 4 & 2 & 3 & 9  & 8 & 0 & 1 \\
Chemistry        & 4 & 2 & 3 & 9  & 4 & 5 & 0 \\
Circuit          & 4 & 2 & 3 & 9  & 5 & 4 & 0 \\
\midrule
\textbf{Total}   & \textbf{23} & \textbf{18} & \textbf{19} & \textbf{60} & \textbf{39} & \textbf{20} & \textbf{1} \\
\bottomrule
\end{tabular}}
}
\end{center}

\noindent\textbf{Question templates.}
(1) Table~\ref{tab:appx_dqa_regular} presents 41 question templates: 23 descriptive and 18 standard reasoning. Descriptive templates assess symbol identification (\eg, recognizing a zigzag as a resistor, a right-angle square as 90$^\circ$), element counting (\eg, vertices, atoms, faces), and pattern recognition (\eg, data trends, series/parallel topology). Standard reasoning templates require applying domain-specific formulas (\eg, $V = IR$ for circuits, $A = \tfrac{1}{2}bh$ for triangles, interior angle sum $= (n{-}2)\times 180^\circ$) or performing visual reasoning (\eg, shortest path length, cross-section shape identification); (2) Table~\ref{tab:appx_dqa_whatif} presents 19 what-if templates for the reasoning questions. Each poses a hypothetical modification to the diagram (\eg, changing a data value, removing an element, scaling a dimension) and asks about the consequence. These modifications are deliberately aligned with D2C-E editing operations: for instance, a what-if question about removing a node (aligned with scope task S5) tests whether the model can reason about connectivity changes, while a question about modifying a resistance value (aligned with text task T3) tests circuit analysis under parameter changes.

\begin{table}[!tbp]
\centering
\caption{Question templates in DQA (Part~1: Graph Structures, Charts, Planar Geometry). ``D'' = descriptive; ``R'' = standard reasoning.}
\label{tab:appx_dqa_regular}
\renewcommand{\arraystretch}{1.1}
\setlength{\tabcolsep}{4.mm}{
\resizebox{\linewidth}{!}{%
\begin{tabular}{c c l p{10cm} c}
\toprule
\textbf{ID} & \textbf{Type} & \textbf{Name} & \textbf{Question Template} & \textbf{OI} \\
\midrule
\rowcolor{softgray} \multicolumn{5}{@{}l@{}}{\textbf{Graph Structures}} \\
Q1 & D & Degree Query & What is the degree of the node labeled [\textit{label}]? & NUM \\
Q2 & D & In-degree Query & What is the in-degree of the node labeled [\textit{label}]? & NUM \\
Q3 & D & Out-degree Query & What is the out-degree of the node labeled [\textit{label}]? & NUM \\
Q4 & D & Neighbors Query & List all nodes that are direct neighbors of [\textit{label}]. & LIST \\
Q5 & R & Path Length & What is the length of the shortest path from [\textit{node\_A}] to [\textit{node\_B}]? & NUM \\
Q6 & R & Tree Height & What is the height of this tree? & NUM \\
\rowcolor{softgray} \multicolumn{5}{@{}l@{}}{\textbf{Charts}} \\
Q1 & D & Max/Min Query & Which data point/category has the maximum (or minimum) value? & TERM \\
Q2 & R & Mean Query & What is the mean (average) value of all visible data points? & NUM \\
Q3 & R & Median Query & What is the median value of the dataset? & NUM \\
Q4 & R & Range Query & What is the range (max $-$ min) of the dataset? & NUM \\
Q5 & D & Monotonicity & In the interval from [\textit{start}] to [\textit{end}], what trend does the data exhibit? & TERM \\
Q6 & R & Extrema Location & At approximately what x-coordinate does the function reach its minimum? & NUM \\
Q7 & D & Heatmap Value & According to the color scale, what is the approximate value at cell ([\textit{row}], [\textit{col}])? & NUM \\
Q8 & D & Polar Plot & At angle $\theta$ = [\textit{angle}]$^\circ$, what is the radial distance $r$? & NUM \\
\rowcolor{softgray} \multicolumn{5}{@{}l@{}}{\textbf{Planar Geometry}} \\
Q1 & D & Right Angle Symbol & What angle measure does the small square symbol at [\textit{location}] indicate? & NUM \\
Q2 & D & Equal Length Marks & Sides [\textit{A}] and [\textit{B}] have identical tick marks. What does this indicate? & TERM \\
Q3 & D & Dashed Line & What does the dashed line in this geometric diagram typically represent? & TERM \\
Q4 & D & Triangle Type & Based on the marked information, what type of triangle is [\textit{label}]? & TERM \\
Q5 & D & Quadrilateral Type & Based on the marked properties, what type of quadrilateral is [\textit{label}]? & TERM \\
Q6 & R & Perimeter & If side [\textit{A}] = [\textit{v1}] and side [\textit{B}] = [\textit{v2}], what is the perimeter of rectangle [\textit{rect}]? & NUM \\
Q7 & R & Area & If base [\textit{b}] = [\textit{v1}] and height [\textit{h}] = [\textit{v2}], what is the area of triangle [\textit{tri}]? & NUM \\
Q8 & R & Interior Angle Sum & What is the sum of interior angles of this [\textit{n}]-sided polygon? & NUM \\
\bottomrule
\end{tabular}}
}
\end{table}

\begin{table}[!tbp]
\centering
\caption{Question templates in DQA (Part~2: Circuit Diagrams, 3D Shapes, Chemistry). ``D'' = descriptive; ``R'' = standard reasoning.}
\label{tab:appx_dqa_regular_2}
\renewcommand{\arraystretch}{1.1}
\setlength{\tabcolsep}{4.mm}{
\resizebox{\linewidth}{!}{%
\begin{tabular}{c c l p{10cm} c}
\toprule
\textbf{ID} & \textbf{Type} & \textbf{Name} & \textbf{Question Template} & \textbf{OI} \\
\midrule
\rowcolor{softgray} \multicolumn{5}{@{}l@{}}{\textbf{Circuit Diagrams}} \\
Q1 & D & Resistor ID & What component is represented by the zigzag symbol labeled [\textit{label}]? & TERM \\
Q2 & D & Capacitor ID & What component is represented by the two parallel lines labeled [\textit{label}]? & TERM \\
Q3 & D & Diode ID & What component is represented by the triangle-with-line symbol labeled [\textit{label}]? & TERM \\
Q4 & D & Series/Parallel & Are components [\textit{comp\_A}] and [\textit{comp\_B}] connected in series or parallel? & TERM \\
Q5 & R & Total Resistance & If [\textit{R1}] = [\textit{v1}]\,$\Omega$ and [\textit{R2}] = [\textit{v2}]\,$\Omega$ are in series, what is the total resistance? & NUM \\
Q6 & R & Ohm's Law & If [\textit{R}] has resistance [\textit{val}]\,$\Omega$ and current [\textit{I}]\,A flows through it, what is the voltage? & NUM \\
\rowcolor{softgray} \multicolumn{5}{@{}l@{}}{\textbf{3D Shapes}} \\
Q1 & D & Solid Identification & What is the name of this 3D solid? & TERM \\
Q2 & R & Face Count & How many faces does this solid have? & NUM \\
Q3 & R & Vertex Count & How many vertices does this solid have? & NUM \\
Q4 & D & Hidden Edges & What do the dashed lines in this 3D diagram represent? & TERM \\
Q5 & R & Cross-section & If this [\textit{solid}] is cut horizontally, what is the shape of the cross-section? & TERM \\
Q6 & R & Volume & If the cylinder has radius [\textit{r}] and height [\textit{h}], what is its volume? (Use $\pi$ = 3.14) & NUM \\
Q7 & R & Surface Area & If the cube has edge length [\textit{e}], what is its surface area? & NUM \\
\rowcolor{softgray} \multicolumn{5}{@{}l@{}}{\textbf{Chemistry}} \\
Q1 & D & Unlabeled Vertex & What atom does the unlabeled vertex at [\textit{location}] represent? & TERM \\
Q2 & D & Bond Type & What type of bond connects atom [\textit{A}] to atom [\textit{B}]? & TERM \\
Q3 & R & Implicit Hydrogens & How many implicit hydrogen atoms are bonded to the carbon at position [\textit{pos}]? & NUM \\
Q4 & D & Functional Group & What functional group is present at [\textit{location}]? & TERM \\
Q5 & D & Carbon Count & How many carbon atoms are in this molecule (including labeled and unlabeled)? & NUM \\
Q6 & R & Molecular Formula & What is the molecular formula of this compound? & TERM \\
\bottomrule
\end{tabular}}
}
\end{table}

\begin{table}[!tbp]
\centering
\caption{DQA what-if question templates. Each template specifies a hypothetical modification and asks about its consequence.}
\label{tab:appx_dqa_whatif}
\renewcommand{\arraystretch}{1.1}
\setlength{\tabcolsep}{4.mm}{
\resizebox{\linewidth}{!}{%
\begin{tabular}{c l p{10.5cm} c}
\toprule
\textbf{ID} & \textbf{Name} & \textbf{Question Template} & \textbf{OI} \\
\midrule
\rowcolor{softgray} \multicolumn{4}{l}{\textbf{Graph Structures}} \\
W1 & Node Removal & Remove node [\textit{target}] and all its edges. How many connected components remain? & NUM \\
W2 & Edge Addition & Add a new edge between [\textit{node\_A}] and [\textit{node\_B}]. What is the new degree of [\textit{node\_A}]? & NUM \\
W3 & Edge Weight Mod. & Change the edge weight from [\textit{A}] to [\textit{B}] from [\textit{old}] to [\textit{new}]. What is the new total weight of the path from [\textit{start}] to [\textit{end}]? & NUM \\
\rowcolor{softgray} \multicolumn{4}{l}{\textbf{Charts}} \\
W1 & Value Modification & Change data point [\textit{label}] from [\textit{old}] to [\textit{new}]. What is the new mean of the dataset? & NUM \\
W2 & Data Removal & Remove data point [\textit{target}]. What is the new maximum value? & NUM \\
W3 & Uniform Shift & Increase all data values by [\textit{$\delta$}]. What is the new range of the dataset? & NUM \\
\rowcolor{softgray} \multicolumn{4}{l}{\textbf{Planar Geometry}} \\
W1 & Side Scaling & Scale side [\textit{label}] by a factor of [\textit{k}]. What is the new perimeter? & NUM \\
W2 & Uniform Scaling & Scale all sides of the [\textit{shape}] by a factor of [\textit{k}]. By what factor does the area increase? & NUM \\
W3 & Angle Modification & Change angle [\textit{A}] from [\textit{old}]$^\circ$ to 90$^\circ$. What type of triangle does [\textit{tri}] become? & TERM \\
\rowcolor{softgray} \multicolumn{4}{l}{\textbf{Circuit Diagrams}} \\
W1 & Resistance Mod. & Change [\textit{R}] from [\textit{old}] to [\textit{new}]\,$\Omega$. If voltage is [\textit{V}]\,V, what is the new current? & NUM \\
W2 & Short Circuit & Short-circuit [\textit{R}] (0\,$\Omega$). What is the current through [\textit{R'}] = [\textit{val}]\,$\Omega$ at [\textit{V}]\,V? & NUM \\
W3 & Parallel Addition & Add [\textit{val}]\,$\Omega$ in parallel with [\textit{R}] (also [\textit{val}]\,$\Omega$). What is the equivalent resistance? & NUM \\
\rowcolor{softgray} \multicolumn{4}{l}{\textbf{3D Shapes}} \\
W1 & Radius Scaling & Scale the cylinder radius by factor [\textit{k}], keeping height constant. By what factor does the volume increase? & NUM \\
W2 & Uniform Scaling & Scale all dimensions of the rectangular prism by factor [\textit{k}]. By what factor does the surface area increase? & NUM \\
W3 & Cone Slicing & Remove the top half of the cone by slicing at mid-height. What is the remaining solid called? & TERM \\
W4 & Stacking & Stack an identical [\textit{shape}] on top. What is the total volume of the combined structure? & NUM \\
\rowcolor{softgray} \multicolumn{4}{l}{\textbf{Chemistry}} \\
W1 & Bond Type Change & Change the bond between [\textit{A}] and [\textit{B}] from single to double. How many implicit H on [\textit{A}]? & NUM \\
W2 & Substituent Repl. & Replace [\textit{old\_group}] at [\textit{pos}] with [\textit{new\_group}]. What is the change in molecular weight? & NUM \\
W3 & Oxygen Insertion & Insert an O atom between [\textit{A}] and [\textit{B}] to form [\textit{A}]-O-[\textit{B}]. What functional group is this? & TERM \\
\bottomrule
\end{tabular}}
}
\end{table}

\noindent\textbf{Domain knowledge.}
Each template encodes domain-specific knowledge required for correct answers. The knowledge spans: graph theory concepts (degree, in-/out-degree, connected components, shortest path) for Graph Structures; statistical measures (mean, median, range, monotonicity) for Charts; Euclidean geometry (shape classification, perimeter, area, interior angle sum) for Planar Geometry; circuit analysis (Ohm's law $V{=}IR$, series/parallel resistance, power) for Circuits; solid geometry (face/vertex counting, volume, surface area, cross-sections) for 3D Shapes; and organic chemistry conventions (implicit hydrogens from valence rules, bond types, functional group identification, molecular weight) for Chemistry. We summarize the domain knowledge per diagram type in Table~\ref{de_stats}. 

\vspace{0.3em}

\FloatBarrier

\subsection{DQA Descriptive Question Examples}
\label{sec:appx_dqa_descriptive_examples}

Figure~\ref{fig:appx_dqa_descriptive_showcase} presents representative DQA descriptive question examples. These questions assess the model's ability to identify domain-specific symbols and extract information directly from the diagram.

\casegroupheader{dqa-header}{Descriptive Questions in Diagram Question Answering}

\begin{tcolorbox}[casebox, colback=dqa-bg, colframe=dqa-header!50]
\textbf{\footnotesize Charts Case 1}\\[2pt]
\begin{minipage}[c]{0.24\linewidth}
  \centering
  \includegraphics[width=\linewidth,height=3cm,keepaspectratio]{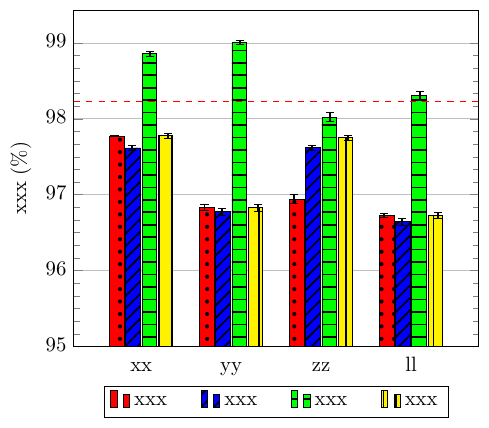}
\end{minipage}%
\hfill
\begin{minipage}[c]{0.73\linewidth}
  \textbf{Q:} Which category and bar style (by color/pattern) has the maximum value?\\[4pt]
  \textbf{A:} green with horizontal lines
\end{minipage}
\end{tcolorbox}%
\begin{tcolorbox}[casebox, colback=dqa-bg, colframe=dqa-header!50]
\textbf{\footnotesize Charts Case 2}\\[2pt]
\begin{minipage}[c]{0.24\linewidth}
  \centering
  \includegraphics[width=\linewidth,height=3cm,keepaspectratio]{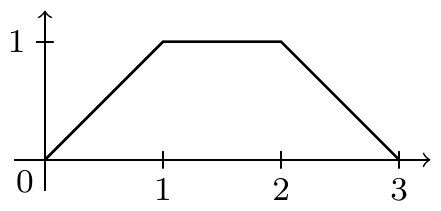}
\end{minipage}%
\hfill
\begin{minipage}[c]{0.73\linewidth}
  \textbf{Q:} In the interval from 0 to 1, what trend does the data exhibit?\\[4pt]
  \textbf{A:} increasing
\end{minipage}
\end{tcolorbox}%
\begin{tcolorbox}[casebox, colback=dqa-bg, colframe=dqa-header!50]
\textbf{\footnotesize Graph Structures Case 1}\\[2pt]
\begin{minipage}[c]{0.24\linewidth}
  \centering
  \includegraphics[width=\linewidth,height=3cm,keepaspectratio]{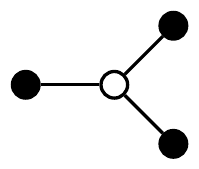}
\end{minipage}%
\hfill
\begin{minipage}[c]{0.73\linewidth}
  \textbf{Q:} What is the degree of the white node?\\[4pt]
  \textbf{A:} 3
\end{minipage}
\end{tcolorbox}%
\begin{tcolorbox}[casebox, colback=dqa-bg, colframe=dqa-header!50]
\textbf{\footnotesize Graph Structures Case 2}\\[2pt]
\begin{minipage}[c]{0.24\linewidth}
  \centering
  \includegraphics[width=\linewidth,height=3cm,keepaspectratio]{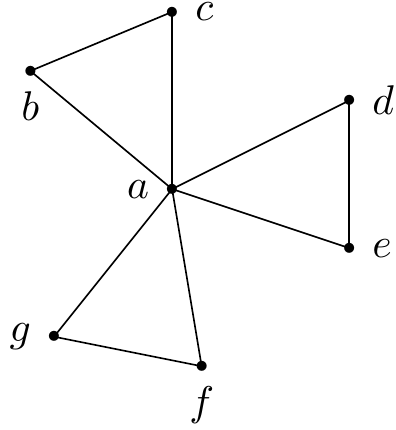}
\end{minipage}%
\hfill
\begin{minipage}[c]{0.73\linewidth}
  \textbf{Q:} List all nodes that are direct neighbors of b.\\[4pt]
  \textbf{A:} a, c
\end{minipage}
\end{tcolorbox}%
\begin{tcolorbox}[casebox, colback=dqa-bg, colframe=dqa-header!50]
\textbf{\footnotesize Planar Geometry Case 1}\\[2pt]
\begin{minipage}[c]{0.24\linewidth}
  \centering
  \includegraphics[width=\linewidth,height=3cm,keepaspectratio]{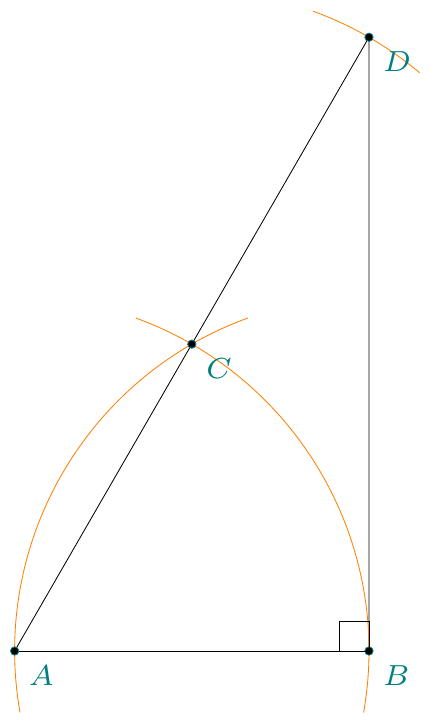}
\end{minipage}%
\hfill
\begin{minipage}[c]{0.73\linewidth}
  \textbf{Q:} What angle measure does the small square symbol at vertex B indicate?\\[4pt]
  \textbf{A:} 90 deg
\end{minipage}
\end{tcolorbox}%
\begin{tcolorbox}[casebox, colback=dqa-bg, colframe=dqa-header!50]
\textbf{\footnotesize Planar Geometry Case 2}\\[2pt]
\begin{minipage}[c]{0.24\linewidth}
  \centering
  \includegraphics[width=\linewidth,height=3cm,keepaspectratio]{figures/appendix/cases/scidiagram_3538.png}
\end{minipage}%
\hfill
\begin{minipage}[c]{0.73\linewidth}
  \textbf{Q:} Based on the marked information, what type of triangle is formed by the segments labeled $\cos\alpha$, $\sin\alpha$, and the radius of the circle?\\[4pt]
  \textbf{A:} right triangle
\end{minipage}
\end{tcolorbox}%
\begin{tcolorbox}[casebox, colback=dqa-bg, colframe=dqa-header!50]
\textbf{\footnotesize 3D Shapes Case 1}\\[2pt]
\begin{minipage}[c]{0.24\linewidth}
  \centering
  \includegraphics[width=\linewidth,height=3cm,keepaspectratio]{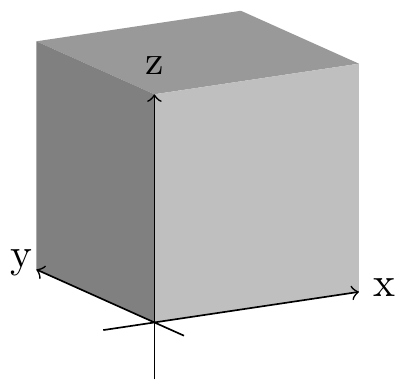}
\end{minipage}%
\hfill
\begin{minipage}[c]{0.73\linewidth}
  \textbf{Q:} What is the name of this 3D solid?\\[4pt]
  \textbf{A:} cube
\end{minipage}
\end{tcolorbox}%
\begin{tcolorbox}[casebox, colback=dqa-bg, colframe=dqa-header!50]
\textbf{\footnotesize 3D Shapes Case 2}\\[2pt]
\begin{minipage}[c]{0.24\linewidth}
  \centering
  \includegraphics[width=\linewidth,height=3cm,keepaspectratio]{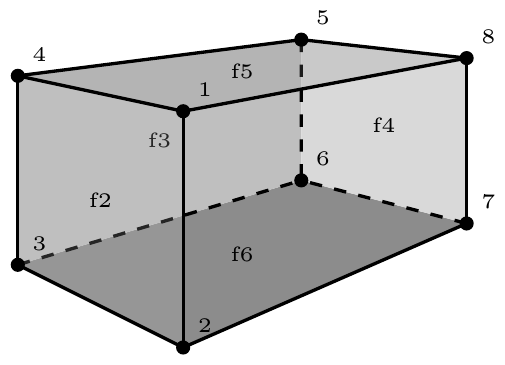}
\end{minipage}%
\hfill
\begin{minipage}[c]{0.73\linewidth}
  \textbf{Q:} What do the dashed lines in this 3D diagram represent?\\[4pt]
  \textbf{A:} hidden edges
\end{minipage}
\end{tcolorbox}%
\begin{tcolorbox}[casebox, colback=dqa-bg, colframe=dqa-header!50]
\textbf{\footnotesize Circuit Diagrams Case 1}\\[2pt]
\begin{minipage}[c]{0.24\linewidth}
  \centering
  \includegraphics[width=\linewidth,height=3cm,keepaspectratio]{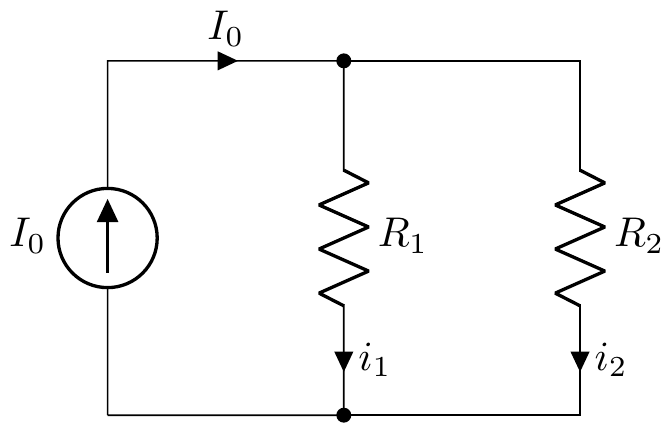}
\end{minipage}%
\hfill
\begin{minipage}[c]{0.73\linewidth}
  \textbf{Q:} What type of component is represented by the zigzag symbol labeled R1?\\[4pt]
  \textbf{A:} resistor
\end{minipage}
\end{tcolorbox}%
\begin{tcolorbox}[casebox, colback=dqa-bg, colframe=dqa-header!50]
\textbf{\footnotesize Circuit Diagrams Case 2}\\[2pt]
\begin{minipage}[c]{0.24\linewidth}
  \centering
  \includegraphics[width=\linewidth,height=3cm,keepaspectratio]{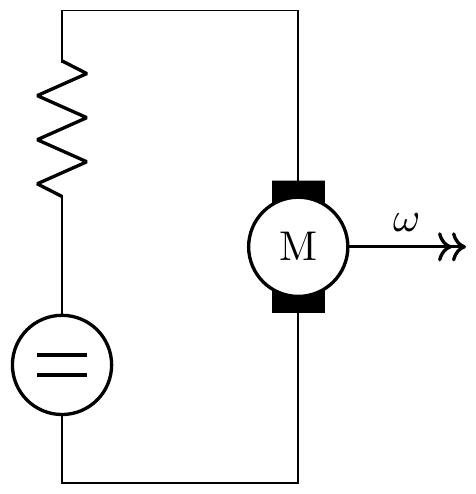}
\end{minipage}%
\hfill
\begin{minipage}[c]{0.73\linewidth}
  \textbf{Q:} Are the resistor and the component labeled M connected in series or parallel?\\[4pt]
  \textbf{A:} series
\end{minipage}
\end{tcolorbox}%
\begin{tcolorbox}[casebox, colback=dqa-bg, colframe=dqa-header!50]
\textbf{\footnotesize Chemistry Case 1}\\[2pt]
\begin{minipage}[c]{0.24\linewidth}
  \centering
  \includegraphics[width=\linewidth,height=3cm,keepaspectratio]{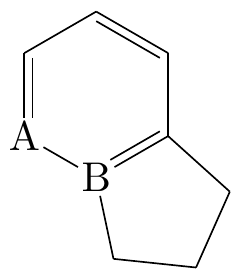}
\end{minipage}%
\hfill
\begin{minipage}[c]{0.73\linewidth}
  \textbf{Q:} What type of bond connects atom A to atom B?\\[4pt]
  \textbf{A:} single bond
\end{minipage}
\end{tcolorbox}%
\begin{tcolorbox}[casebox, colback=dqa-bg, colframe=dqa-header!50]
\textbf{\footnotesize Chemistry Case 2}\\[2pt]
\begin{minipage}[c]{0.24\linewidth}
  \centering
  \includegraphics[width=\linewidth,height=3cm,keepaspectratio]{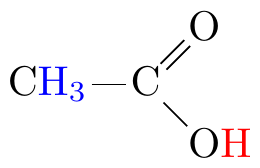}
\end{minipage}%
\hfill
\begin{minipage}[c]{0.73\linewidth}
  \textbf{Q:} What functional group is present at the terminal position on the right?\\[4pt]
  \textbf{A:} carboxyl group
\end{minipage}
\end{tcolorbox}

\captionof{figure}{Representative DQA descriptive question examples. Each case shows a diagram and a question that requires direct observation and information extraction from the visual content.}
\label{fig:appx_dqa_descriptive_showcase}

\subsection{DQA Reasoning Question Examples}
\label{sec:appx_dqa_reasoning_examples}

Figure~\ref{fig:appx_dqa_reasoning_showcase} presents representative DQA reasoning question examples. These questions require numerical computation with domain-specific formulas or predicting outcomes of hypothetical modifications.

\casegroupheader{dqa-header}{Reasoning Questions in Diagram Question Answering}

\begin{tcolorbox}[casebox, colback=dqa-bg, colframe=dqa-header!50]
\textbf{\footnotesize Charts Case 1}\\[2pt]
\begin{minipage}[c]{0.24\linewidth}
  \centering
  \includegraphics[width=\linewidth,height=3cm,keepaspectratio]{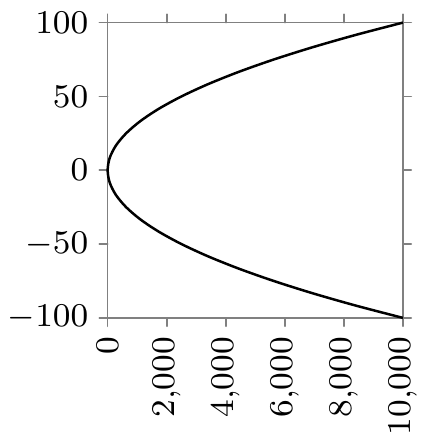}
\end{minipage}%
\hfill
\begin{minipage}[c]{0.73\linewidth}
  \textbf{Q:} What is the range (difference between maximum and minimum) of the vertical y-axis values shown in the dataset?\\[4pt]
  \textbf{A:} 200
\end{minipage}
\end{tcolorbox}%
\begin{tcolorbox}[casebox, colback=dqa-bg, colframe=dqa-header!50]
\textbf{\footnotesize Charts Case 2}\\[2pt]
\begin{minipage}[c]{0.24\linewidth}
  \centering
  \includegraphics[width=\linewidth,height=3cm,keepaspectratio]{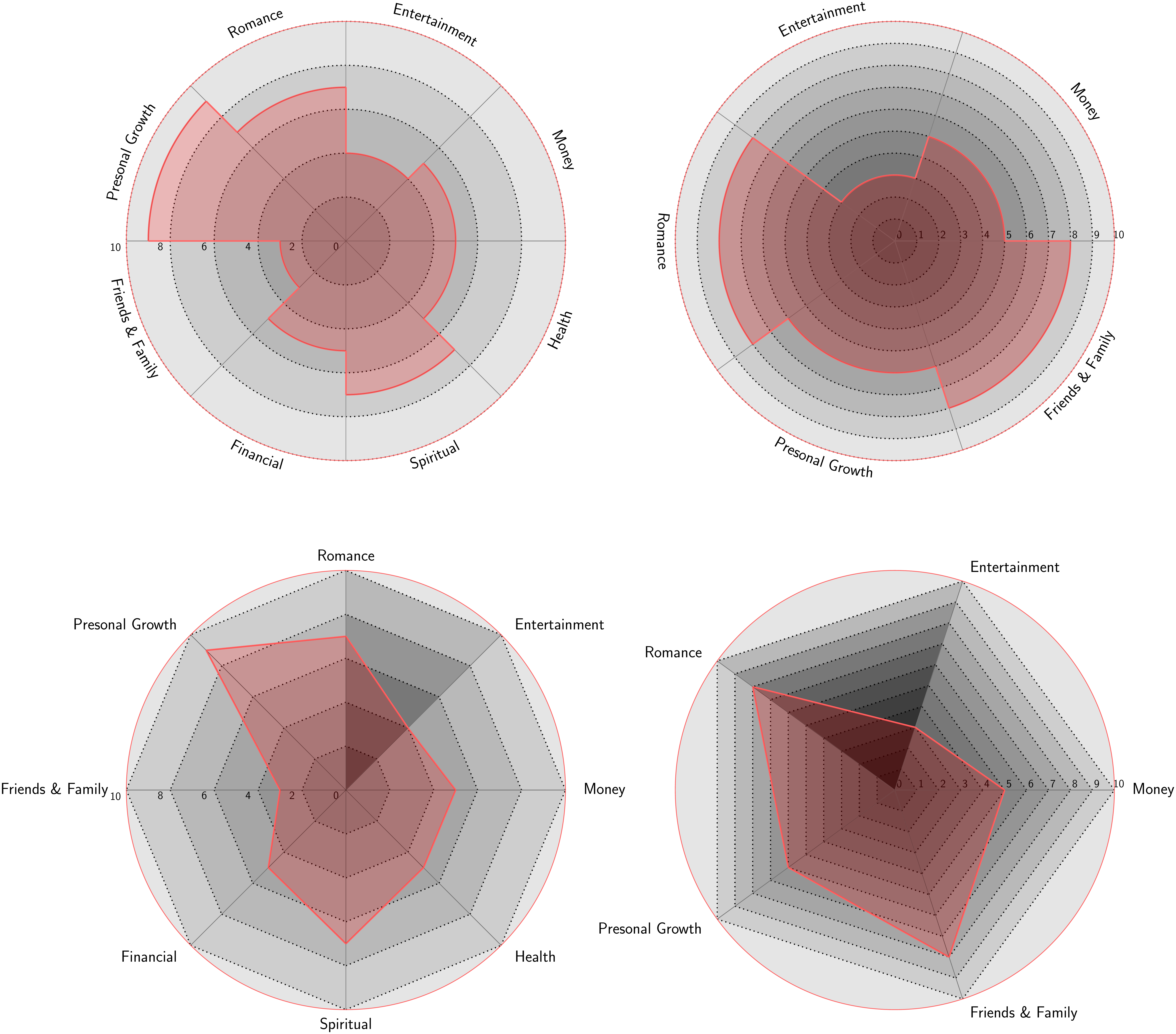}
\end{minipage}%
\hfill
\begin{minipage}[c]{0.73\linewidth}
  \textbf{Q:} Consider the following modification to the dataset in the top-right circular radar chart: Change the value of data point 'Money' from 5 to 10. (Type: T4) What is the new mean of the entire dataset?\\[4pt]
  \textbf{A:} 7
\end{minipage}
\end{tcolorbox}%
\begin{tcolorbox}[casebox, colback=dqa-bg, colframe=dqa-header!50]
\textbf{\footnotesize Graph Structures Case 1}\\[2pt]
\begin{minipage}[c]{0.24\linewidth}
  \centering
  \includegraphics[width=\linewidth,height=3cm,keepaspectratio]{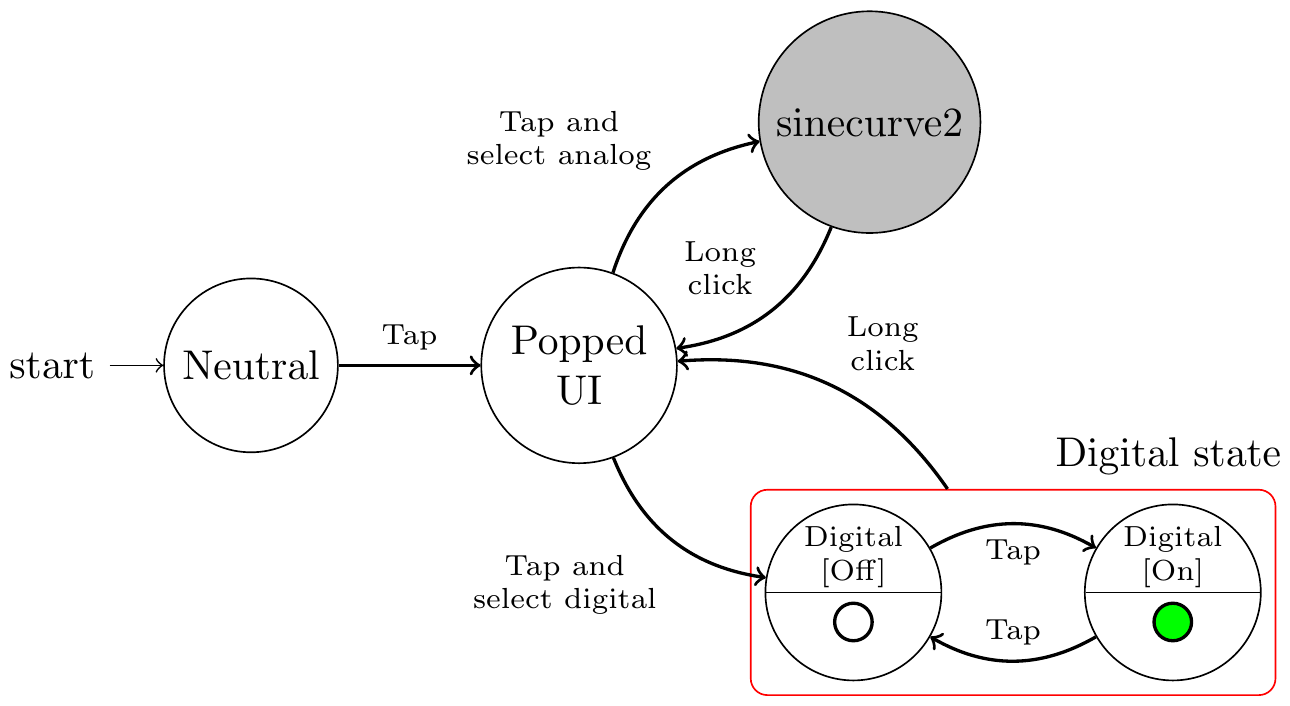}
\end{minipage}%
\hfill
\begin{minipage}[c]{0.73\linewidth}
  \textbf{Q:} What is the length of the shortest path from Neutral to Digital [On]?\\[4pt]
  \textbf{A:} 3
\end{minipage}
\end{tcolorbox}%
\begin{tcolorbox}[casebox, colback=dqa-bg, colframe=dqa-header!50]
\textbf{\footnotesize Graph Structures Case 2}\\[2pt]
\begin{minipage}[c]{0.24\linewidth}
  \centering
  \includegraphics[width=\linewidth,height=3cm,keepaspectratio]{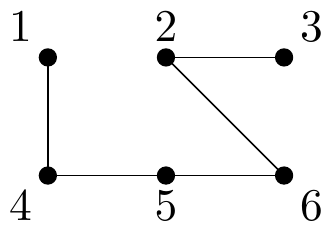}
\end{minipage}%
\hfill
\begin{minipage}[c]{0.73\linewidth}
  \textbf{Q:} Consider the following modification: Remove the node 4 and all its connected edges. (Type: T5) How many connected components remain in the graph?\\[4pt]
  \textbf{A:} 2
\end{minipage}
\end{tcolorbox}%
\begin{tcolorbox}[casebox, colback=dqa-bg, colframe=dqa-header!50]
\textbf{\footnotesize Planar Geometry Case 1}\\[2pt]
\begin{minipage}[c]{0.24\linewidth}
  \centering
  \includegraphics[width=\linewidth,height=3cm,keepaspectratio]{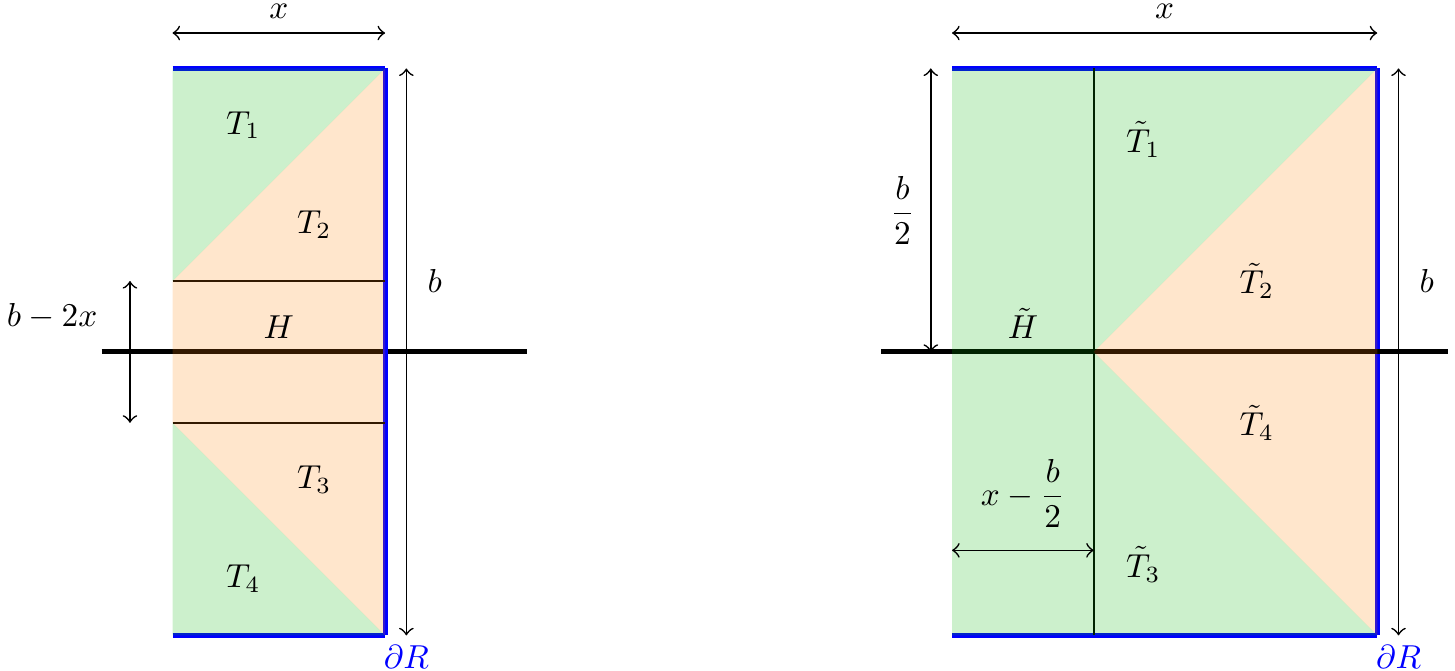}
\end{minipage}%
\hfill
\begin{minipage}[c]{0.73\linewidth}
  \textbf{Q:} In the diagram on the right, if the vertical base of triangle T\textasciitilde{}2 along the right edge is b = 10 units and its horizontal height from the vertex is 5 units, what is the area of triangle T\textasciitilde{}2?\\[4pt]
  \textbf{A:} 12.5
\end{minipage}
\end{tcolorbox}%
\begin{tcolorbox}[casebox, colback=dqa-bg, colframe=dqa-header!50]
\textbf{\footnotesize Planar Geometry Case 2}\\[2pt]
\begin{minipage}[c]{0.24\linewidth}
  \centering
  \includegraphics[width=\linewidth,height=3cm,keepaspectratio]{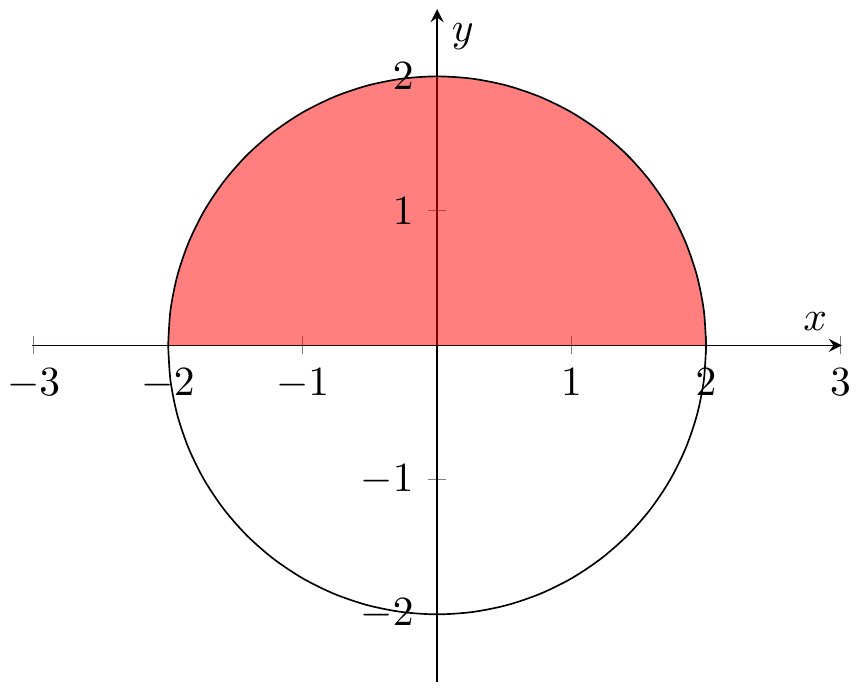}
\end{minipage}%
\hfill
\begin{minipage}[c]{0.73\linewidth}
  \textbf{Q:} Consider the following modification: Scale the radius of the circle (currently 2 units) by a factor of 2. (Type: T2) What is the new perimeter of the full circle?\\[4pt]
  \textbf{A:} 8pi
\end{minipage}
\end{tcolorbox}%
\begin{tcolorbox}[casebox, colback=dqa-bg, colframe=dqa-header!50]
\textbf{\footnotesize 3D Shapes Case 1}\\[2pt]
\begin{minipage}[c]{0.24\linewidth}
  \centering
  \includegraphics[width=\linewidth,height=3cm,keepaspectratio]{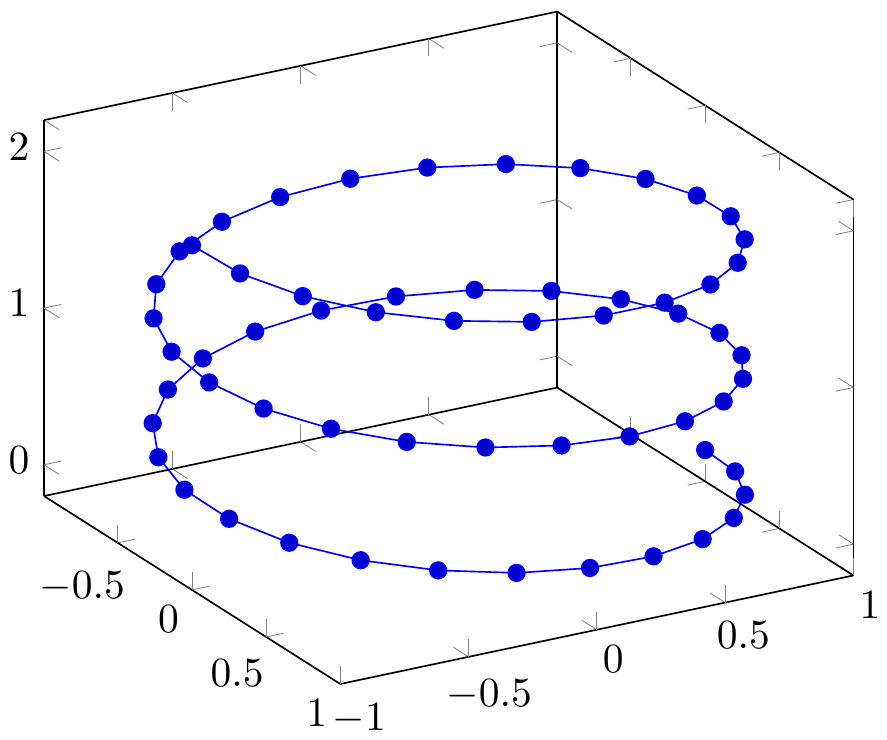}
\end{minipage}%
\hfill
\begin{minipage}[c]{0.73\linewidth}
  \textbf{Q:} If the cylinder that bounds this helix has radius 1 and height 2, what is its volume? (Use pi = 3.14)\\[4pt]
  \textbf{A:} 6.28
\end{minipage}
\end{tcolorbox}%
\begin{tcolorbox}[casebox, colback=dqa-bg, colframe=dqa-header!50]
\textbf{\footnotesize 3D Shapes Case 2}\\[2pt]
\begin{minipage}[c]{0.24\linewidth}
  \centering
  \includegraphics[width=\linewidth,height=3cm,keepaspectratio]{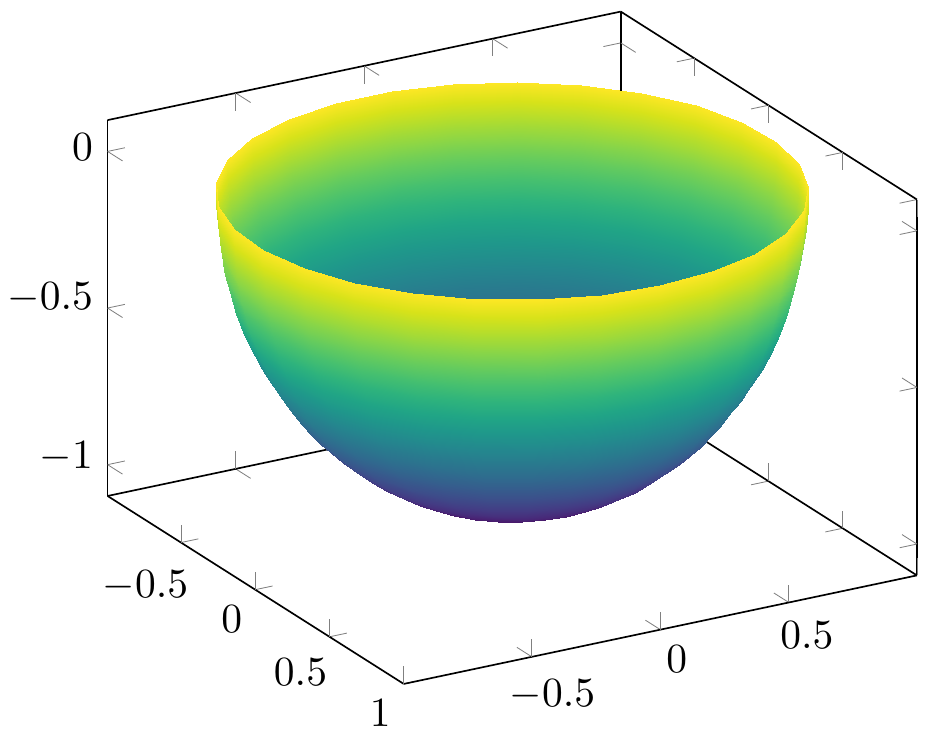}
\end{minipage}%
\hfill
\begin{minipage}[c]{0.73\linewidth}
  \textbf{Q:} If this hemisphere is cut horizontally, what is the shape of the cross-section?\\[4pt]
  \textbf{A:} circle
\end{minipage}
\end{tcolorbox}%
\begin{tcolorbox}[casebox, colback=dqa-bg, colframe=dqa-header!50]
\textbf{\footnotesize Circuit Diagrams Case 1}\\[2pt]
\begin{minipage}[c]{0.24\linewidth}
  \centering
  \includegraphics[width=\linewidth,height=3cm,keepaspectratio]{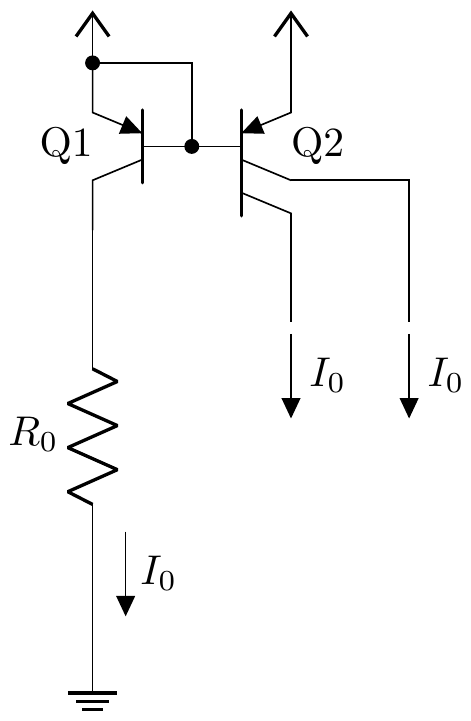}
\end{minipage}%
\hfill
\begin{minipage}[c]{0.73\linewidth}
  \textbf{Q:} If resistor R\_0 has resistance 1000 Ohms and current I\_0 = 0.005 A flows through it, what is the voltage across it?\\[4pt]
  \textbf{A:} 5V
\end{minipage}
\end{tcolorbox}%
\begin{tcolorbox}[casebox, colback=dqa-bg, colframe=dqa-header!50]
\textbf{\footnotesize Circuit Diagrams Case 2}\\[2pt]
\begin{minipage}[c]{0.24\linewidth}
  \centering
  \includegraphics[width=\linewidth,height=3cm,keepaspectratio]{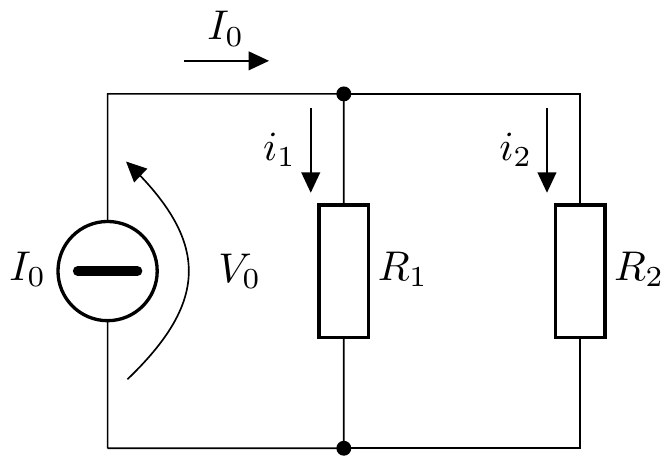}
\end{minipage}%
\hfill
\begin{minipage}[c]{0.73\linewidth}
  \textbf{Q:} Consider the following modification: Add a resistor of 100 Ohms in parallel with the existing resistor R1 (also 100 Ohms). (Type: T9) What is the equivalent resistance of this parallel combination? \\[4pt]
  \textbf{A:} 50 Ohm
\end{minipage}
\end{tcolorbox}%
\begin{tcolorbox}[casebox, colback=dqa-bg, colframe=dqa-header!50]
\textbf{\footnotesize Chemistry Case 1}\\[2pt]
\begin{minipage}[c]{0.24\linewidth}
  \centering
  \includegraphics[width=\linewidth,height=3cm,keepaspectratio]{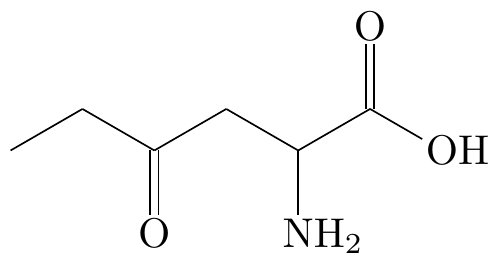}
\end{minipage}%
\hfill
\begin{minipage}[c]{0.73\linewidth}
  \textbf{Q:} Consider the following modification: Change the single bond between the two carbon atoms at the far left of the chain from a single bond to a double bond. (Type: T7) How many implicit hydrogens are now bonded to the terminal carbon atom at the far left?\\[4pt]
  \textbf{A:} 2
\end{minipage}
\end{tcolorbox}%
\begin{tcolorbox}[casebox, colback=dqa-bg, colframe=dqa-header!50]
\textbf{\footnotesize Chemistry Case 2}\\[2pt]
\begin{minipage}[c]{0.24\linewidth}
  \centering
  \includegraphics[width=\linewidth,height=3cm,keepaspectratio]{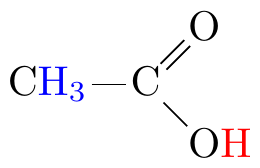}
\end{minipage}%
\hfill
\begin{minipage}[c]{0.73\linewidth}
  \textbf{Q:} Consider the following modification: Replace the -OH group at the bottom right with a -Cl atom. (Type: T9) What is the change in molecular weight (New - Old)?\\[4pt]
  \textbf{A:} 18.44 g/mol
\end{minipage}
\end{tcolorbox}

\captionof{figure}{Representative DQA reasoning question examples. Each case shows a diagram and a question that requires domain-specific reasoning, computation, or hypothetical analysis beyond direct observation.}
\label{fig:appx_dqa_reasoning_showcase}

\section{Evaluation Methodology}
\label{sec:appx_eval_method}

\subsection{Semantic Object Model (SOM) Pipeline}
\label{sec:appx_som}

Our object-based metric evaluates whether the model correctly perceives basic objects that the code draws. We parse both generated and ground-truth TikZ code into a set of graphical objects---their type, text content, color, and bounding box---and compute F1 scores for each dimension. The extraction proceeds through three stages: semantic injection, compilation, and DOM-based extraction.

\vspace{0.5em}
\noindent\textbf{Stage 1: Semantic injection via \path|semantic_spy.sty|.}
Before compilation, we preprocess the TikZ source by injecting a custom \LaTeX{} style file, \path|semantic_spy.sty|, as the \emph{last} loaded package in the preamble. This package hooks into TikZ, pgfplots, and CircuiTikZ commands via \verb|\special{dvisvgm:raw ...}| directives, which embed semantic XML tags directly into the DVI output. Specifically, it installs the following hooks:
\begin{itemize}[leftmargin=*,nosep,topsep=2pt]
  \item \textbf{TikZ core hooks.} Every \verb|\node| is wrapped in \path|<g class="tikz-node">|. The wrapper stores \path|data-id|, \path|data-shape|, and optionally \path|data-text|. Similarly, every \path|tikzpicture| and \path|scope| environment is wrapped with corresponding semantic group tags.
  \item \textbf{pgfplots data hooks.} Every axis environment is wrapped with \path|<g class="pgf-axis">|. Within each \verb|\addplot|, scatter marker hooks inject \path|<g class="data-point">| tags with original coordinates (\path|data-x|, \path|data-y|) and parent-series identifiers. Each plot series is wrapped with \path|<g class="pgf-series">|.
  \item \textbf{CircuiTikZ component hooks.} Every bipole component is wrapped with \path|<g class="circuit-component">|. The wrapper carries a unique component identifier.
\end{itemize}
The preprocessor also handles code standardization: wrapping fragments in standalone documents, detecting and injecting missing packages/libraries, fixing common compilation issues (duplicate packages, xcolor option clashes, pgfplots compatibility version downgrades), and checking DVI compatibility. If the code is incompatible with DVI mode (\eg, uses \path|fontspec| or \path|xeCJK|), semantic injection is skipped and the pipeline falls back to geometry-only extraction.

\vspace{0.5em}
\noindent\textbf{Stage 2: Compilation and SVG conversion.}
The preprocessed code is compiled to DVI format using the \path|latex| engine, which preserves the \verb|\special| directives containing the semantic tags. If DVI compilation fails, the pipeline falls back to PDF-based compilation with multi-engine fallback (\path|pdflatex| $\rightarrow$ \path|lualatex| $\rightarrow$ \path|xelatex|), which still produces geometry-only SVG.
The DVI (or PDF) output is then converted to SVG using \path|dvisvgm| with parameters \path|--font-format=svg| (to preserve text elements), \path|--precision=6|, and \path|--zoom=1| (for 1:1 coordinate mapping). For DVI input, \path|dvisvgm| faithfully translates the injected \verb|\special| tags into SVG group elements with the corresponding CSS classes and data attributes. For PDF input, \path|dvisvgm --pdf| produces geometric SVG without semantic annotations.

\vspace{0.5em}
\noindent\textbf{Stage 3: DOM-based SOM extraction.}
The generated SVG is parsed using an \path|lxml|-based DOM parser. The extractor traverses the SVG tree recursively and classifies each element into one of the following categories:

\begin{itemize}[leftmargin=*,nosep,topsep=2pt]
  \item \emph{Semantic elements}: identified by CSS class attributes injected by \path|semantic_spy.sty|. These include \path|tikz-node| (\textsc{Node}), \path|tikz-path| (\textsc{Path}), \path|pgf-axis| (\textsc{Axis}), \path|pgf-series| (\textsc{DataSeries}), \path|data-point| (\textsc{DataPoint}), and \path|circuit-component| (\textsc{Component}).
  \item \emph{Native SVG shapes}: identified by SVG tag names (\path|rect|, \path|circle|, \path|ellipse|, \path|path|, \path|text|, etc.), mapped to corresponding geometric types.
\end{itemize}

For each element, the extractor records four attributes:
\begin{itemize}[leftmargin=*,nosep,topsep=2pt]
  \item \textbf{Type.} A fine-grained type key that combines the coarse element type with subtype attributes, \eg, \path|node:circle|, \path|node:rectangle|, \path|path:closed|, \path|path:open|, and \path|component:R| (resistor). Structural container types such as \path|picture|, \path|scope|, and \path|axis| are excluded from type-level evaluation because they would inflate the score.
  \item \textbf{Text.} Text content is extracted with the following priority: (1)~the \path|data-text| attribute from semantic injection; (2)~the \path|metadata| child element; (3)~glyph reconstruction from \path|<use>| elements (see below); and (4)~fallback to \path|<text>| or \path|<tspan>| children. For \textsc{DataPoint} elements, the $(x, y)$ data values are formatted as text for matching.
  \item \textbf{Color.} Extracted from \path|fill| and \path|stroke| attributes for path elements, and from \path|fill| for text elements. Colors specified as \path|none| or \path|transparent| are excluded.
  \item \textbf{BBox.} Axis-aligned bounding boxes are computed from SVG geometry: directly from coordinate attributes for basic shapes (\path|rect|, \path|circle|, \path|ellipse|), from path data using \path|svgpathtools| for \path|<path>| elements, and recursively aggregated for group elements. Cumulative SVG transforms (\path|translate|, \path|scale|, \path|matrix|) are tracked through the DOM tree.
\end{itemize}

\vspace{0.5em}
\noindent\textbf{Glyph text reconstruction.}
A key challenge is that \path|dvisvgm| converts text into glyph paths stored in \path|<defs>| and referenced by \path|<use>| elements. Their IDs follow the pattern \path|g{font_id}-{char_code}|. To recover text content, we collect all \path|<use>| elements within a group, sort them by $x$-position (left to right), decode the character codes from glyph IDs, and concatenate the characters. For example, \path|<use href="#g0-65"/>| corresponds to character code 65 (``A''). This reconstruction handles both single-character and multi-character text groups, including colored text whose parent \path|<g>| carries a \path|fill| attribute.

\vspace{0.5em}
\noindent\textbf{Per-domain element mapping.}
The SOM pipeline maps diagrams from different TikZ packages into a unified element schema. Table~\ref{tab:appx_som_domains} summarizes the primary element types extracted per domain.

\begin{table}[!tbp]
\centering
\caption{Primary SOM element types extracted per domain. All domains share basic \textsc{Node}, \textsc{Path}, and \textsc{Text} types; domain-specific types arise from specialized packages.}
\label{tab:appx_som_domains}
\renewcommand{\arraystretch}{1.1}
\setlength{\tabcolsep}{4.mm}{
\resizebox{0.85\linewidth}{!}{%
\begin{tabular}{l l l}
\toprule
\textbf{Domain} & \textbf{Primary TikZ Pkg.} & \textbf{Key SOM Element Types} \\
\midrule
Charts & \texttt{pgfplots} & \textsc{Axis}, \textsc{DataSeries}, \textsc{DataPoint}, \textsc{Node} (legend), \textsc{Text} (tick labels) \\
Graph Struct. & \texttt{tikz} & \textsc{Node} (circle, rectangle, diamond), \textsc{Path} (open/closed), \textsc{Text} (labels) \\
Planar Geom. & \texttt{tikz/tkz-euclide} & \textsc{Node} (point markers), \textsc{Path} (edges, arcs), \textsc{Text} (vertex labels, annotations) \\
Circuit & \texttt{circuitikz} & \textsc{Component} (R, C, L, V, etc.), \textsc{Node} (junctions), \textsc{Text} (value labels) \\
3D Shapes & \texttt{pgfplots/tikz} & \textsc{Axis}, \textsc{Path} (surfaces, wireframes), \textsc{Node} (labels), \textsc{Fill}/\textsc{Filldraw} \\
Chemistry & \texttt{chemfig} & \textsc{Node} (atoms, functional groups), \textsc{Path} (bonds), \textsc{Text} (atom labels) \\
\bottomrule
\end{tabular}}
}
\end{table}

\subsection{Evaluation Metrics}
\label{sec:appx_f1_formular}
\noindent\textbf{Object-based metric computation.} Formally, given TikZ source code $c$ in domain $d$, a domain extractor $\mathcal{E}_d$ maps code to attribute sets:
\begin{equation}
  \mathcal{E}_d(c) \;=\; \bigl\{\, \mathcal{T}_d(c),\; \mathcal{X}_d(c),\; \mathcal{K}_d(c),\; \mathcal{B}_d(c) \,\bigr\}.
\end{equation}
Here,
$\mathcal{T}_d(c)$ is a multiset of primitive {types};
$\mathcal{X}_d(c)$ is a multiset of {textual} elements;
$\mathcal{K}_d(c)$ is a multiset of normalized {color} specifications; and
$\mathcal{B}_d(c)$ is a set of axis-aligned {bounding boxes}
$b_i=(x, y, w, h)$, with $(x, y)$ denoting the center coordinates and $w$ and $h$ representing the width and height.

\noindent\textit{Primitive  F1 Score.}
Given predicted code $\hat{c}$ and ground-truth code $c$, we extract 4 graphic objects via $\mathcal{E}_d$.
Let $M_*$ denote the number of correct matches (defined below), $|\hat{S}|$ the predicted count, and $|S|$ the ground-truth count. The per-dimension F1 is defined as follows, where the subscript * is replaced by each instance in $\{type, text, color, bbox\}$.
\begin{equation}
\mathrm{P}_* = \frac{M_*}{|\hat{S}|},\quad
\mathrm{R}_* = \frac{M_*}{|S|},\quad
\mathrm{F1}_* = \frac{2\,\mathrm{P}_*\,\mathrm{R}_*}{\mathrm{P}_* + \mathrm{R}_*}.
\end{equation}
The four dimensions differ only in how $M_*$ is computed: (1) Count-based multiset matching for type.
Let $\hat{n}(t)$ and $n(t)$ be the counts of type $t \in \mathcal{T}_d$ in predicted and ground-truth code:
\begin{equation}
M_{{type}} = \textstyle\sum_{t \in \mathcal{T}_d} \min\!\left(\hat{n}(t),\, n(t)\right).
\end{equation}
{(2) Greedy one-to-one exact string matching for text.
Each ground-truth text element matches at most one predicted element by exact string equality: $M_{{text}} = |\text{matched pairs}|$. (3) Permutation-based optimal assignment for color~\cite{yangchartmimic}.
Colors are grouped by associated type of element. Within each group $g$, the optimal permutation $\pi$ of predicted colors is selected using CIE2000 perceptual similarity:
\begin{equation}
M_{{color}} = \textstyle\sum_{g}\,\max_{\pi}\sum_{i}\,
\mathrm{sim}\!\left(\pi(\hat{k}_i^g),\, k_i^{g}\right),
\quad
\mathrm{sim}(\hat{k},k) = \max\!\left(0,\; 1 - \frac{\Delta E_{00}(\hat{k},\,k)}{100}\right).
\end{equation}
(4) Greedy one-to-one IoU matching for BBox.
For each ground-truth box $b$ (iterated in order), we search the remaining unmatched predicted boxes and select the first $\hat{b}$ whose Intersection-over-Union (IoU) exceeds a threshold $\tau$=0.3: $\mathrm{IoU}(\hat{b},\,b) = \frac{|\hat{b} \cap b|}{|\hat{b} \cup b|} \ge \tau$. The matched predicted box is then removed from the candidate pool. The final matching count is $M_{{bbox}}=|\text{matched pairs}|$. 

\noindent The primary perception-level metric is the average across the four objects:
\begin{equation}
\mathrm{F1}_{{avg}} = \tfrac{1}{4}\left(\mathrm{F1}_{{type}} + \mathrm{F1}_{{text}} + \mathrm{F1}_{{color}} + \mathrm{F1}_{{bbox}}\right).
\end{equation}

\vspace{0.5em}
\noindent\textbf{Code-based metric computation.}
We use CrystalBLEU~\cite{eghbali2022crystalbleu} to measure token-level similarity between generated and ground-truth TikZ code. CrystalBLEU extends standard BLEU-4 by filtering out trivially shared n-grams---high-frequency, low-information tokens such as \texttt{\{}, \texttt{\}}, \texttt{[}, \texttt{]}, \texttt{,}, \texttt{;}, \texttt{=}, \verb|\begin|, and \verb|\end|---that would inflate scores without reflecting meaningful code similarity. Tokenization is performed using the Pygments \texttt{TexLexer}, which correctly segments \LaTeX{} control sequences, and tokens of type \texttt{Whitespace} and \texttt{Comment} are discarded.

For the editing task, CrystalBLEU is also decomposed into preserve-only and edit-only variants. The edit-only CrystalBLEU performs multiset subtraction of source n-grams from both GT and generated n-grams before computing BLEU, isolating the contribution of newly introduced code. The preserve-only variant keeps only n-grams that intersect with the source (Counter intersection), measuring how faithfully the model retains unchanged code portions.

\vspace{0.5em}
\noindent\textbf{Image-based metric computation.}
We compile both generated and ground-truth TikZ code into PNG images (300 DPI, cropped whitespace) and compare them using four established image similarity metrics. The rendering pipeline uses \texttt{latexmk} with multi-engine fallback (\texttt{pdflatex} $\rightarrow$ \texttt{lualatex} $\rightarrow$ \texttt{xelatex}), followed by PDF cropping (via \texttt{pdfCropMargins}) and \texttt{pdftoppm} conversion to PNG. All images are resized to $256 \times 256$ pixels with white background before metric computation.

\begin{itemize}[leftmargin=*,nosep,topsep=2pt]
  \item \textbf{SSIM}~\cite{wang2004ssim}. Structural Similarity Index Measure, computed via \texttt{torchmetrics} with data range normalized to $[0,1]$. Compares brightness, contrast, and structural patterns between two images (higher is better).

  \item \textbf{CLIP Score}~\cite{radford2021learning}. Cosine similarity between CLIP ViT-B/32 image embeddings of the generated and ground-truth images. Captures high-level semantic similarity in a shared vision-language embedding space (higher is better).

  \item \textbf{LPIPS}~\cite{zhang2018unreasonable}. Learned Perceptual Image Patch Similarity, computed using an AlexNet backbone. Measures perceptual distance using deep features that correlate with human perceptual judgments (lower is better).

  \item \textbf{FID}~\cite{heusel2017gans}. Fr\'{e}chet Inception Distance, a distribution-level metric computed over all generated and ground-truth images within each model's evaluation set using \texttt{torchmetrics} with Inception v3 features (2048-dim). Requires $\geq$2 images per set (lower is better).
\end{itemize}

In the main results (Table~\ref{tab:main_results}), we report two composite image-level scores: SC (the average of SSIM and CLIP, higher is better) and FL (the average of FID and LPIPS, lower is better).

\vspace{0.5em}
\noindent\textbf{Edge cases.}
When both the predicted and ground-truth sets are empty for a given dimension (\eg, neither code contains any text elements), we define $\mathrm{F1} = 1.0$ (perfect match), since the model correctly produces no elements of that type. When the color group contains $\leq 6$ elements, we use exhaustive permutation search; for larger groups ($>6$ elements), we switch to greedy approximation to maintain computational tractability.

\vspace{0.5em}
\noindent\textbf{Preserve/edit splits in D2C-E.}
For the diagram-to-code editing task, a single overall score is insufficient: an edit typically changes only a small portion of the diagram, so a model that ignores the editing instruction and simply reproduces the original diagram could already score high. We therefore split evaluation into two complementary parts: 

(1) {Preserve-only evaluation} measures how well the model retains elements that should remain unchanged after editing. For each dimension, we keep only elements present in both the generated code and the original source code (intersection), then compute F1 score against the ground-truth preserved elements. Formally, for a given dimension:
\begin{itemize}[leftmargin=*,nosep,topsep=2pt]
  \item \emph{GT preserve set}: GT elements that match source elements.
  \item \emph{Gen preserve set}: Generated elements that match source elements.
  \item F1 score is computed between the GT preserve set and the Gen preserve set.
\end{itemize}
(2) {Edit-only evaluation} isolates the quality of the actual edits. For each dimension, elements common between the source and the ground-truth are subtracted from both the ground-truth and the generated code before computing F1 score. This eliminates the contribution of unchanged elements and focuses exclusively on what the editing instruction asks to change. Formally:
\begin{itemize}[leftmargin=*,nosep,topsep=2pt]
  \item \emph{Common set}: Elements present in both source and GT (computed via greedy matching).
  \item \emph{GT edit set}: GT elements after removing the common set.
  \item \emph{Gen edit set}: Generated elements after removing the common set.
  \item F1 score is computed between the GT edit set and the Gen edit set.
\end{itemize}
We apply the same procedure for object- and code-based metric computation on both preserve and edit splits.

\subsection{DQA LLM-as-a-Judge Details}
\label{sec:appx_dqa_grading}

We evaluate DQA responses through a two-stage pipeline: rule-based matching followed by LLM-as-a-judge fallback for ambiguous cases.

\vspace{0.5em}
\noindent\textbf{Stage 1: Answer extraction.}
The model's raw output is processed to extract the final answer. Models are instructed to return JSON \texttt{\{"answer": "...", "reasoning": "..."\}}, but actual outputs may contain markdown code blocks, extra text, or malformed JSON. The extractor tries four strategies in order: (1)~direct JSON parsing; (2)~extraction from markdown \texttt{```json ... ```} code blocks; (3)~finding the outermost \texttt{\{...\}} braces; (4)~regex fallback for \texttt{"answer": "..."} patterns. Thinking tags (\texttt{<think>...</think>}) are stripped before extraction.

\vspace{0.5em}
\noindent\textbf{Stage 2: Rule-based matching.}
The extracted answer is compared against the ground truth using type-specific rules based on the output instruction (OI) type:
\begin{itemize}[leftmargin=*,nosep,topsep=2pt]
  \item \textsc{OI-NUM} (5,013 questions, 70.2\%): Numeric comparison with absolute tolerance ($\leq 10^{-6}$) or relative tolerance ($\leq 1\%$), with unit normalization (\eg, ``degrees'' $=$ ``deg'' $=$ ``$^\circ$''; ``amps'' $=$ ``A'').
  \item \textsc{OI-TERM} (1,779 questions, 24.9\%): Case-insensitive exact string match after normalization (removing quotes, collapsing whitespace, stripping trailing punctuation).
  \item \textsc{OI-LIST} (354 questions, 5.0\%): Order-independent set comparison. Lists are parsed from JSON arrays or CSV format, and items are normalized before set equality check.
\end{itemize}
Each rule produces a confidence score. When confidence is high ($\geq 0.85$), the rule verdict is accepted directly. When confidence is low (\eg, partial containment for terms, or unparseable candidate), the case is escalated to Stage~3.

\vspace{0.5em}
\noindent\textbf{Stage 3: LLM-as-a-judge fallback.}
For cases where rule-based matching yields low confidence, we use an LLM judge. Given a question, the ground-truth answer, and the model's response, we prompt the judge LLM to classify the response as \textsc{Correct}, \textsc{Incorrect}, or \textsc{Unattempted}. The judge prompt is type-aware: it includes OI-specific guidelines (\eg, for \textsc{OI-NUM}, small rounding differences are acceptable; for \textsc{OI-LIST}, extra or missing items make the answer incorrect). The judge outputs a structured two-line response: (1)~an evaluation explaining the reasoning, and (2)~a label. The final binary score is 1 for \textsc{Correct} and 0 otherwise.

The full judge prompt template is shown below:

\begin{tcolorbox}[colback=softgray, colframe=headerblue, title={\small DQA LLM-as-a-Judge Prompt}, fonttitle=\bfseries\small, boxrule=0.5pt, arc=2pt, left=4pt, right=4pt, top=2pt, bottom=2pt, breakable]
\footnotesize
\textbf{Role:} You are an expert judge specialized in evaluating the correctness of answers to scientific diagram understanding questions.

\vspace{0.3em}
\textbf{Task:} Classify the model's response into one of three categories:
\begin{enumerate}[nosep,leftmargin=*]
\item \textbf{Correct}: The model answer contains the core information of the ground truth and is semantically consistent.
\item \textbf{Incorrect}: The model answer contradicts the ground truth.
\item \textbf{Unattempted}: The model explicitly states it cannot answer, or the response is empty.
\end{enumerate}

\vspace{0.3em}
\textbf{Answer Type:} \texttt{\{output\_instruction\}} \\
\texttt{\{OI-specific guidelines\}}

\vspace{0.3em}
\textbf{Output Format:}
\begin{enumerate}[nosep,leftmargin=*]
\item \textbf{Evaluation}: [Brief explanation of reasoning]
\item \textbf{Label}: [Correct / Incorrect / Unattempted]
\end{enumerate}

\vspace{0.3em}
\textbf{Input:}\\
Question: \texttt{\{question\}} \\
Model Answer: \texttt{\{model\_answer\}} \\
Ground Truth Answer: \texttt{\{ground\_truth\_answer\}}
\end{tcolorbox}

\subsection{LLM Judge Human Agreement Validation}
\label{sec:appx_judge_agreement}
To verify the reliability of our LLM-as-a-judge evaluation, we conduct a human agreement study following prior work that validates automated evaluators against independent human annotations and human-majority labels using agreement-based metrics~\cite{badshah2025clev}.

\vspace{0.5em}\noindent\textbf{Sampling.}
We stratify-sample 200 cases from all 7,146 DQA questions along two dimensions: (1)~\emph{evaluation stage}, 70 cases resolved by the LLM judge (Stage~3) and 130 by rule-based matching (Stage~2), which ensures coverage of the most subjective cases; and (2)~\emph{output instruction type}, 133 OI-NUM, 53 OI-TERM, and 14 OI-LIST, approximately following the corpus proportions. We use GPT-5.2 model outputs as the representative responses for annotation.

\vspace{0.5em}\noindent\textbf{Annotation protocol.}
Three graduate-student annotators independently label each \mbox{(question, model\_answer, ground\_truth)} triple as \textsc{Correct}, \textsc{Incorrect}, or \textsc{Unattempted}, following the same OI-specific grading guidelines used by the LLM judge (\S\ref{sec:appx_dqa_grading}). Annotators are shown the diagram image, question text, ground-truth answer, and model answer, but \emph{not} the automated judge's verdict.

\vspace{0.5em}\noindent\textbf{Agreement metrics.}
We report Cohen's Kappa ($\kappa$) and raw accuracy. Specifically, $\kappa_{\text{J-H}}$ measures agreement between the LLM judge and the human majority vote, while $\kappa_{\text{H-H}}$ measures Fleiss' Kappa among the three human annotators as a human reference. We additionally report raw accuracy because $\kappa$ can be deflated under extreme label imbalance, even when agreement is very high~\cite{badshah2025clev}.

\begin{table}[!tbp]
\centering
\caption{Human agreement validation of the DQA LLM judge. $\kappa_{\text{J-H}}$: Cohen's Kappa between the LLM judge and human majority vote. $\kappa_{\text{H-H}}$: Fleiss' Kappa among three human annotators. Acc: percentage agreement between judge and human majority. $^\dagger$Stage~2 $\kappa$ values are deflated by extreme label skew (99\%+ correct); raw accuracy better reflects the near-perfect agreement on this subset.}
\label{tab:judge_agreement}
\renewcommand{\arraystretch}{1.1}
\setlength{\tabcolsep}{4.mm}{
\resizebox{0.5\linewidth}{!}{%
\begin{tabular}{lcrrr}
\toprule
\textbf{Subset} & $N$ & $\kappa_{\text{J-H}}$ & $\kappa_{\text{H-H}}$ & \textbf{Acc (\%)} \\
\midrule
All            & 200 & 0.937 & 0.915 & 97.5 \\
\midrule
Stage 2 (Rule) & 130 & 0.000$^\dagger$ & 0.663$^\dagger$ & 99.2 \\
Stage 3 (LLM)  &  70 & 0.839 & 0.805 & 94.3 \\
\midrule
OI-NUM         & 133 & 0.961 & 0.949 & 98.5 \\
OI-TERM        &  53 & 0.911 & 0.809 & 96.2 \\
OI-LIST        &  14 & 0.811 & 1.000 & 92.9 \\
\bottomrule
\end{tabular}}
}
\end{table}

\vspace{0.5em}\noindent\textbf{Results.}
As shown in Table~\ref{tab:judge_agreement}, the LLM judge closely matches human consensus overall, achieving $\kappa_{\text{J-H}} = 0.937$ with 97.5\% accuracy, indicating \emph{almost perfect} agreement. This is comparable to the human agreement level ($\kappa_{\text{H-H}} = 0.915$), suggesting that the automated judge reliably tracks majority human judgments. For the 130 Stage~2 cases, only one judge--human disagreement occurs (99.2\% accuracy); the near-zero $\kappa$ is an artifact of extreme label skew rather than poor agreement. For the 70 Stage~3 cases---the most subjective subset---the judge still achieves $\kappa_{\text{J-H}} = 0.839$ (94.3\% accuracy), close to the corresponding human agreement ($\kappa_{\text{H-H}} = 0.805$). Across output instruction types, agreement is highest for OI-NUM ($\kappa_{\text{J-H}} = 0.961$) and lowest for OI-LIST ($\kappa_{\text{J-H}} = 0.811$), consistent with the greater ambiguity of list-valued answers.
\vspace{0.5em}

\section{Implementation Details}
\label{sec:appx_impl}

\subsection{Model Configurations}
\label{sec:appx_model_config}

Table~\ref{tab:appx_model_config} summarizes the generation configurations for all 12 evaluated models.
Closed-source models and several large open-source models are accessed via their respective provider APIs.
The remaining open-source models are deployed locally or remotely using vLLM\,(v0.14.0).
All models use a sampling temperature of $T{=}1.0$ except TikZero+\,10B ($T{=}0.8$).
The maximum output length $L_{\max}$ is set to 16{,}384 tokens for API-based models to accommodate long TikZ programs, while locally-served models use smaller limits (4{,}096--8{,}192) due to GPU memory constraints.

\begin{table}[!tbp]
\centering
\caption{Generation configurations of evaluated models. ``API'' denotes cloud-hosted model endpoints; ``vLLM'' denotes models deployed with vLLM. $T$: sampling temperature; $L_{\max}$: maximum output tokens.}
\label{tab:appx_model_config}
\renewcommand{\arraystretch}{1.1}
\setlength{\tabcolsep}{4.mm}{
\resizebox{0.6\linewidth}{!}{%
\begin{tabular}{lccrl}
\toprule
\textbf{Model} & \textbf{Inference} & $T$ & $L_{\max}$ & \textbf{Notes} \\
\midrule
\multicolumn{5}{c}{\textit{Closed-Source}} \\
\midrule
Gemini-3.1 Pro        & API  & 1.0 & 16{,}384 &  \\
Gemini-3.0 Pro        & API  & 1.0 & 16{,}384 &  \\
Gemini-3.0 Flash      & API  & 1.0 & 16{,}384 &  \\
GPT-5.2               & API  & 1.0 & 16{,}384 & reasoning\_effort\,=\,medium \\
Claude-4.6 Opus       & API  & 1.0 & 16{,}384 &  \\
Seed-2.0 Pro          & API  & 1.0 & 16{,}384 &  \\
\midrule
\multicolumn{5}{c}{\textit{Open-Source}} \\
\midrule
Qwen3.5-397B-A17B     & API  & 1.0 & 16{,}384 &  \\
Qwen3-VL-235B-A22B    & API  & 1.0 &  6{,}144 &  \\
Kimi-K2.5             & API  & 1.0 & 16{,}384 &  \\
Qwen3-VL-8B           & vLLM & 1.0 &  8{,}192 & 1$\times$A800 \\
InternVL3-38B         & vLLM & 1.0 &  8{,}192 &  \\
TikZero+\,10B         & vLLM & 0.8 &  4{,}096 & 1$\times$A800 \\
\bottomrule
\end{tabular}}
}
\end{table}

\subsection{Inference and Compilation}
\label{sec:appx_inference}

\noindent\textbf{Inference pipeline.}
All models receive diagram images encoded as base64 data URIs within multimodal messages.
For D2C-P and D2C-E, the system prompt instructs the model to output complete \LaTeX{} code inside a single \texttt{```latex} code block, starting from \verb|\documentclass| and ending with \verb|\end{document}|.
For DQA, models return structured JSON with \texttt{answer} and \texttt{reasoning} fields.
Generated code is extracted from the model response via regex-based markdown code-block parsing.
Each inference call is subject to a 300-second timeout; failed calls are retried up to 3 times with exponential backoff (2, 4, 8\,s delays).

\vspace{0.5em}
\noindent\textbf{Hardware.}
All experiments are conducted on a single machine with 8$\times$NVIDIA A800 80\,GB GPUs.
Open-source models served locally with vLLM each occupy one GPU.
Image-level metric computation (SSIM, LPIPS, CLIP) runs on a dedicated GPU to avoid memory contention with inference.
The DQA LLM judge (Qwen3-Next-80B-A3B, \S\ref{sec:appx_dqa_grading}) is deployed across 4 GPUs with tensor parallelism via vLLM.

\vspace{0.5em}
\noindent\textbf{\LaTeX{} compilation.}
Generated TikZ code is compiled into PDF using a multi-engine fallback strategy: \texttt{pdflatex} is attempted first, followed by \texttt{lualatex} and \texttt{xelatex} if earlier engines fail.
Each compilation attempt has a 60-second timeout.
Compiled PDFs are cropped using \texttt{pdfCropMargins} and converted to $256{\times}256$ PNG images (300\,DPI) via \texttt{pdftoppm} for image-level metric computation.
For object-level F1 evaluation, code is separately compiled to DVI format and converted to SVG for Structured Object Model (SOM) extraction.

\vspace{0.5em}
\noindent\textbf{Failure handling.}
When compilation fails or times out, all upward metrics (F1$_{avg}$, CrystalBLEU, SSIM, CLIP) default to~0, and downward metrics (LPIPS) default to~1.0, ensuring a fixed denominator across all models.
Inference failures, including API errors, empty responses, or timeouts after all retries, receive the same penalty scores.
This deterministic treatment prevents missing data from inflating any model's average performance.

\section{Evaluation Prompts}
\label{sec:appx_prompts}

We list 16 evaluation settings (\textbf{S1-S16}) in Table~\ref{tab:context_settings}. Herein, we provide the full system prompts per setting. Specifically, each inference call follows a two-message structure:
{(1)}~a \emph{system message} containing the task prompt (presented below), and
{(2)}~a \emph{user message} containing the diagram image (base64-encoded) together with a task-specific text query (\eg, ``Convert this diagram to LaTeX TikZ code'' for D2C-P, an editing instruction for D2C-E, or a question for DQA).
Placeholders enclosed in braces (\texttt{\{...\}}) are filled at inference time with instance-specific data: \texttt{\{perception\_data\}} is a structured JSON list of visual primitives extracted from ground-truth annotations.

\subsection{Foundational Evaluation Prompts}
\label{sec:appx_prompts_found}

The three foundational settings use system prompts that instruct the model to perform the task directly without additional context or tools.

\begin{tcolorbox}[colback=softgray, colframe=basetxt, title={\small S1\,/\,S6: D2C-P Direct Coding\,/\,D2C-E Direct Editing}, fonttitle=\bfseries\small, boxrule=0.5pt, arc=2pt, left=4pt, right=4pt, top=2pt, bottom=2pt, breakable]
\footnotesize
You are a LaTeX expert. Follow these rules strictly:
\begin{enumerate}[nosep,leftmargin=*]
\item Output the COMPLETE LaTeX code inside a single \texttt{```latex} code block.
\item Start from \texttt{\textbackslash documentclass} and end with \texttt{\textbackslash end\{document\}}.
\item Do NOT output any text, explanation, or reasoning outside the code block.
\end{enumerate}
\end{tcolorbox}

\noindent S1 and S6 share the same system prompt; the task is distinguished by the user message: S1 sends only the diagram image, while S6 additionally includes an editing textual instruction.

\begin{tcolorbox}[colback=softgray, colframe=basetxt, title={\small S12: DQA Direct Answering}, fonttitle=\bfseries\small, boxrule=0.5pt, arc=2pt, left=4pt, right=4pt, top=2pt, bottom=2pt, breakable]
\footnotesize
You are a scientific diagram analysis expert.
Answer the question based on the image and follow the required answer format exactly.
\end{tcolorbox}

\subsection{Agentic Evaluation Prompts}
\label{sec:appx_prompts_agent}

The agentic settings augment the foundational prompts with one or more capabilities: context utilization (providing structured object data), tool use (TikZ documentation search via MCP), state management (requiring intermediate code generation), and planning (coordinating multiple capabilities). Below, prompts are organized by task.

\subsubsection{Diagram-to-Code Parsing (S2--S5)}

\begin{tcolorbox}[colback=softgray, colframe=ctxtxt, title={\small S2: + Objects (Context Utilization)}, fonttitle=\bfseries\small, boxrule=0.5pt, arc=2pt, left=4pt, right=4pt, top=2pt, bottom=2pt, breakable]
\footnotesize
You are a LaTeX expert specializing in scientific diagrams.

\textbf{Visual Perception of This Diagram}

The following is a structured analysis of the visual elements in this diagram, extracted from ground truth data. These are the primitive building blocks that compose the diagram --- your generated code must faithfully reproduce all of them.

\texttt{\{perception\_data\}}

\textbf{Instructions}
\begin{itemize}[nosep,leftmargin=*]
\item The perception data above describes the fundamental visual primitives: their types, colors, text content, and spatial positions.
\item Treat these as the complete inventory of what the diagram contains. Every element listed must be present in your output.
\item Named colors (\eg, red, blue, gray) are standard color names.
\item Output the COMPLETE code inside a single \texttt{```latex} code block.
  Start from \texttt{\textbackslash documentclass} and end with \texttt{\textbackslash end\{document\}}.
\end{itemize}
\end{tcolorbox}

\begin{tcolorbox}[colback=softgray, colframe=tooltxt, title={\small S3: + TikZ Search Tool (Tool Use)}, fonttitle=\bfseries\small, boxrule=0.5pt, arc=2pt, left=4pt, right=4pt, top=2pt, bottom=2pt, breakable]
\footnotesize
You are a LaTeX expert with access to a TikZ documentation tool.

\textbf{Available Tool}
\begin{itemize}[nosep,leftmargin=*]
\item \texttt{SearchLaTeXKnowledgeBase}: Search TikZ/PGF/pgfplots documentation for syntax and examples.
\end{itemize}

\textbf{Workflow}
\begin{enumerate}[nosep,leftmargin=*]
\item Observe the diagram image carefully
\item If you need specific TikZ syntax or package usage, use the \texttt{SearchLaTeXKnowledgeBase} tool
\item Generate the COMPLETE LaTeX code inside a single \texttt{```latex} code block
\item Start from \texttt{\textbackslash documentclass} and end with \texttt{\textbackslash end\{document\}}
\end{enumerate}
\end{tcolorbox}

\begin{tcolorbox}[colback=softgray, colframe=statetxt, title={\small S4: + Model Generated Objects (State Management)}, fonttitle=\bfseries\small, boxrule=0.5pt, arc=2pt, left=4pt, right=4pt, top=2pt, bottom=2pt, breakable]
\footnotesize
You are a LaTeX expert. Before writing code, analyze the image systematically.

\textbf{Step 1 --- Perception} (inside \texttt{<perception>} tags):
\begin{enumerate}[nosep,leftmargin=*]
\item Element types
\item Text content
\item Colors
\item Spatial layout
\end{enumerate}

\textbf{Step 2 --- Code Generation:}
Based on your perception, output the COMPLETE LaTeX code inside a single \texttt{```latex} code block.
Start from \texttt{\textbackslash documentclass} and end with \texttt{\textbackslash end\{document\}}.
\end{tcolorbox}

\begin{tcolorbox}[colback=softgray, colframe=plantxt, title={\small S5: + Objects \& TikZ Search Tool (Planning)}, fonttitle=\bfseries\small, boxrule=0.5pt, arc=2pt, left=4pt, right=4pt, top=2pt, bottom=2pt, breakable]
\footnotesize
You are a LaTeX expert with access to a TikZ documentation tool.

\textbf{Visual Perception of This Diagram}

The following is a structured analysis of the visual elements in this diagram, extracted from ground truth data. These are the primitive building blocks that compose the diagram --- your generated code must faithfully reproduce all of them.

\texttt{\{perception\_data\}}

\textbf{Available Tool}
\begin{itemize}[nosep,leftmargin=*]
\item \texttt{SearchLaTeXKnowledgeBase}: Search TikZ/PGF/pgfplots documentation for syntax and examples.
\end{itemize}

\textbf{Instructions}
\begin{itemize}[nosep,leftmargin=*]
\item The perception data above describes the fundamental visual primitives of this diagram. Every element listed must be present in your output.
\item Named colors (\eg, red, blue, gray) are standard color names.
\item Use the \texttt{SearchLaTeXKnowledgeBase} tool when you need specific syntax or package usage.
\item Output the COMPLETE code inside a single \texttt{```latex} code block.
  Start from \texttt{\textbackslash documentclass} and end with \texttt{\textbackslash end\{document\}}.
\end{itemize}
\end{tcolorbox}

\subsubsection{Diagram-to-Code Editing (S7--S11)}

\begin{tcolorbox}[colback=softgray, colframe=ctxtxt, title={\small S7: + Objects (Context Utilization)}, fonttitle=\bfseries\small, boxrule=0.5pt, arc=2pt, left=4pt, right=4pt, top=2pt, bottom=2pt, breakable]
\footnotesize
You are a LaTeX expert.

\textbf{Visual Perception of This Diagram}

The following is a structured analysis of the visual elements in this diagram, extracted from ground truth data. These are the primitive building blocks that compose the diagram.

\texttt{\{perception\_data\}}

\textbf{Instructions}
\begin{itemize}[nosep,leftmargin=*]
\item The perception data above describes the fundamental visual primitives: their types, colors, text content, and spatial positions.
\item Use these primitives to understand the diagram's structure, then apply the user's modification instruction.
\item Output the final modified code inside a single \texttt{```latex} code block.
\item Start from \texttt{\textbackslash documentclass} and end with \texttt{\textbackslash end\{document\}}.
\end{itemize}
\end{tcolorbox}

\begin{tcolorbox}[colback=softgray, colframe=tooltxt, title={\small S8: + TikZ Search Tool (Tool Use)}, fonttitle=\bfseries\small, boxrule=0.5pt, arc=2pt, left=4pt, right=4pt, top=2pt, bottom=2pt, breakable]
\footnotesize
You are a LaTeX expert with access to a TikZ documentation tool.

\textbf{Available Tool}
\begin{itemize}[nosep,leftmargin=*]
\item \texttt{SearchLaTeXKnowledgeBase}: Search TikZ/PGF/pgfplots documentation for syntax and examples.
\end{itemize}

\textbf{Workflow}
\begin{enumerate}[nosep,leftmargin=*]
\item Observe the diagram image carefully
\item If you need specific TikZ syntax or package usage for the edit task, use the \texttt{SearchLaTeXKnowledgeBase} tool
\item Apply the user's modification instruction
\item Output the final modified code inside a single \texttt{```latex} code block
\item Start from \texttt{\textbackslash documentclass} and end with \texttt{\textbackslash end\{document\}}
\end{enumerate}
\end{tcolorbox}

\begin{tcolorbox}[colback=softgray, colframe=statetxt, title={\small S9: + Required TikZ Codes (State Management)}, fonttitle=\bfseries\small, boxrule=0.5pt, arc=2pt, left=4pt, right=4pt, top=2pt, bottom=2pt, breakable]
\footnotesize
You are a LaTeX expert. Complete this task in two steps within a single response.

\textbf{Step 1 --- Reconstruct Original Code:}
Look at the diagram image and generate complete LaTeX code for the original diagram.
Output it inside \texttt{<original\_reconstruction>} tag.

\textbf{Step 2 --- Apply Modification:}
Based on Step 1, apply the user instruction.
Output the final modified code inside a single \texttt{```latex} code block.

\textbf{Rules:}
\begin{itemize}[nosep,leftmargin=*]
\item The \texttt{```latex} code block is the final answer.
\item It must start from \texttt{\textbackslash documentclass} and end with \texttt{\textbackslash end\{document\}}.
\end{itemize}
\end{tcolorbox}

\begin{tcolorbox}[colback=softgray, colframe=plantxt, title={\small S10: + Optional TikZ Codes (Planning)}, fonttitle=\bfseries\small, boxrule=0.5pt, arc=2pt, left=4pt, right=4pt, top=2pt, bottom=2pt, breakable]
\footnotesize
You are a LaTeX expert. You may optionally reconstruct the code before editing.

If you find it helpful, you can first reconstruct the original diagram code inside \texttt{<original\_reconstruction>} tags. This is entirely optional --- skip it if you can apply the edit directly.

Then, apply the user's modification instruction.
Output the final modified code inside a single \texttt{```latex} code block.

\textbf{Rules:}
\begin{itemize}[nosep,leftmargin=*]
\item The \texttt{```latex} code block is the final answer.
\item It must start from \texttt{\textbackslash documentclass} and end with \texttt{\textbackslash end\{document\}}.
\end{itemize}
\end{tcolorbox}

\begin{tcolorbox}[colback=softgray, colframe=plantxt, title={\small S11: + Optional Codes \& TikZ Search Tool (Planning)}, fonttitle=\bfseries\small, boxrule=0.5pt, arc=2pt, left=4pt, right=4pt, top=2pt, bottom=2pt, breakable]
\footnotesize
You are a LaTeX expert with access to a TikZ documentation tool.

\textbf{Available Tool}
\begin{itemize}[nosep,leftmargin=*]
\item \texttt{SearchLaTeXKnowledgeBase}: Search TikZ/PGF/pgfplots documentation for syntax and examples.
\end{itemize}

\textbf{Workflow}

You may optionally reconstruct the original diagram code inside \texttt{<original\_reconstruction>} tags before editing. This is entirely optional --- skip it if you can apply the edit directly.

If you need specific TikZ syntax or package usage, use the \texttt{SearchLaTeXKnowledgeBase} tool.

Then, apply the user's modification instruction.
Output the final modified code inside a single \texttt{```latex} code block.

\textbf{Rules:}
\begin{itemize}[nosep,leftmargin=*]
\item The \texttt{```latex} code block is the final answer.
\item It must start from \texttt{\textbackslash documentclass} and end with \texttt{\textbackslash end\{document\}}.
\end{itemize}
\end{tcolorbox}

\subsubsection{Diagram Question Answering (S13--S16)}

\begin{tcolorbox}[colback=softgray, colframe=ctxtxt, title={\small S13: + Objects (Context Utilization)}, fonttitle=\bfseries\small, boxrule=0.5pt, arc=2pt, left=4pt, right=4pt, top=2pt, bottom=2pt, breakable]
\footnotesize
You are a scientific diagram analysis expert.

\textbf{Visual Perception of This Diagram}

The following is a structured analysis of the visual elements in this diagram, extracted from ground truth data. These are the primitive building blocks that compose the diagram.

\texttt{\{perception\_data\}}

\textbf{Instructions}
\begin{itemize}[nosep,leftmargin=*]
\item The perception data above describes the fundamental visual primitives: their types, colors, text content, and spatial positions.
\item Use these primitives along with the image to answer the question.
\item Follow the answer format specified in the question.
\end{itemize}
\end{tcolorbox}

\begin{tcolorbox}[colback=softgray, colframe=statetxt, title={\small S14: + Required TikZ Codes (State Management)}, fonttitle=\bfseries\small, boxrule=0.5pt, arc=2pt, left=4pt, right=4pt, top=2pt, bottom=2pt, breakable]
\footnotesize
You are a scientific diagram analysis expert.

\textbf{Step 1 --- Code Reconstruction:}
Generate the complete LaTeX code for this diagram.
Output it inside \texttt{<internal\_code\_representation>} tag.

\textbf{Step 2 --- Answer the Question:}
Using both image and code, answer the question.
Follow the answer format specified by the question.
Output final answer after \texttt{</internal\_code\_representation>}.
\end{tcolorbox}

\begin{tcolorbox}[colback=softgray, colframe=plantxt, title={\small S15: + Optional TikZ Codes (Planning)}, fonttitle=\bfseries\small, boxrule=0.5pt, arc=2pt, left=4pt, right=4pt, top=2pt, bottom=2pt, breakable]
\footnotesize
You are a scientific diagram analysis expert.

If you find it helpful, you can first reconstruct the diagram's LaTeX code inside \texttt{<internal\_code\_representation>} tags to aid your analysis. This is entirely optional --- skip it if you can answer directly from the image.

Then, answer the question based on the image (and your code reconstruction if you made one).
Follow the answer format specified by the question.
Output your final answer after any code reconstruction tags.
\end{tcolorbox}

\begin{tcolorbox}[colback=softgray, colframe=plantxt, title={\small S16: + Optional Codes \& TikZ Search Tool (Planning)}, fonttitle=\bfseries\small, boxrule=0.5pt, arc=2pt, left=4pt, right=4pt, top=2pt, bottom=2pt, breakable]
\footnotesize
You are a scientific diagram analysis expert with access to a TikZ documentation tool.

\textbf{Available Tool}
\begin{itemize}[nosep,leftmargin=*]
\item \texttt{SearchLaTeXKnowledgeBase}: Search TikZ/PGF/pgfplots documentation for syntax and examples.
\end{itemize}

\textbf{Workflow}

If you find it helpful, you can first reconstruct the diagram's LaTeX code inside \texttt{<internal\_code\_representation>} tags. This is entirely optional.

If you need to understand specific TikZ syntax or package conventions, use the \texttt{SearchLaTeXKnowledgeBase} tool.

Then, answer the question based on the image (and your analysis).
Follow the answer format specified by the question.
Output your final answer after any code reconstruction tags.
\end{tcolorbox}

\section{MCP-Based TikZ Documentation Server}
\label{sec:appx_mcp}

\subsection{Server Architecture and Implementation}
\label{sec:appx_mcp_arch}

As described in \S\ref{sec:eval_set}, we build the TikZ search tool as an MCP~\cite{anthropic2024mcp} server to give models on demand access to curated TikZ documentation, avoiding the noise of web search and the inefficiency of loading full manuals.

\vspace{0.5em}
\noindent\textbf{Pipeline overview.}
The server is built in three stages: knowledge extraction from official \LaTeX{} source files, unified knowledge base construction, and Mintlify based MCP deployment, as shown in Figure~\ref{fig:appx_mcp_pipeline}.

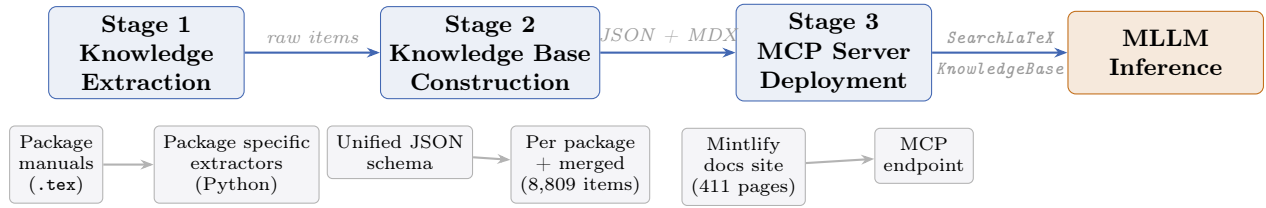
\begin{figure}[!tbp]
\centering
\resizebox{\linewidth}{!}{%
\begin{tikzpicture}[
    node distance=0.6cm and 0.4cm,
    stage/.style={rectangle, rounded corners=3pt, draw=headerblue, fill=headerblue!10,
                  minimum height=1.1cm, minimum width=2.6cm, align=center,
                  font=\small\bfseries},
    substage/.style={rectangle, rounded corners=2pt, draw=gray!60, fill=softgray,
                     minimum height=0.7cm, align=center, font=\scriptsize},
    arr/.style={-{Stealth[length=5pt]}, thick, headerblue},
    lbl/.style={font=\scriptsize\itshape, text=gray!80}
]
\node[stage] (s1) {Stage 1\\Knowledge\\Extraction};
\node[substage, below=0.35cm of s1, xshift=-1.2cm] (src) {Package\\manuals\\(\texttt{.tex})};
\node[substage, below=0.35cm of s1, xshift=1.2cm] (ext) {Package specific\\extractors\\(Python)};
\draw[arr, gray!60] (src) -- (ext);

\node[stage, right=1.8cm of s1] (s2) {Stage 2\\Knowledge Base\\Construction};
\node[substage, below=0.35cm of s2, xshift=-1.2cm] (json) {Unified JSON\\schema};
\node[substage, below=0.35cm of s2, xshift=1.2cm] (merge) {Per package\\+ merged\\(8{,}809 items)};
\draw[arr, gray!60] (json) -- (merge);

\node[stage, right=1.8cm of s2] (s3) {Stage 3\\MCP Server\\Deployment};
\node[substage, below=0.35cm of s3, xshift=-1.2cm] (mint) {Mintlify\\docs site\\(411 pages)};
\node[substage, below=0.35cm of s3, xshift=1.2cm] (mcp) {MCP\\endpoint};
\draw[arr, gray!60] (mint) -- (mcp);

\node[stage, right=1.8cm of s3, fill=tooltxt!15, draw=tooltxt] (mllm) {MLLM\\Inference};

\draw[arr] (s1) -- node[above, lbl] {raw items} (s2);
\draw[arr] (s2) -- node[above, lbl] {JSON + MDX} (s3);
\draw[arr] (s3) -- node[above, lbl] {\texttt{SearchLaTeX}} node[below, lbl] {\texttt{KnowledgeBase}} (mllm);
\end{tikzpicture}}
\caption{Three stage construction pipeline for the MCP based TikZ documentation server. \LaTeX{} source files from 19 package manuals are parsed into a unified JSON knowledge base and then deployed as a Mintlify hosted MCP server. During inference, models query the \texttt{SearchLaTeXKnowledgeBase} tool on demand.}
\label{fig:appx_mcp_pipeline}
\end{figure}

\vspace{0.5em}
\noindent\textbf{Stage 1: Knowledge extraction.}
We extract structured knowledge from the official \LaTeX{} source files of each TikZ package manual. Since packages use different documentation formats, we implement package specific extractors to parse examples, commands, options, and component catalogs. In total, the framework processes about 270 source files from 19 packages using seven Python scripts. Table~\ref{tab:appx_mcp_extractors} summarizes the extraction strategy for the six core packages.

\begin{table}[!tbp]
\centering
\caption{Extraction strategies for the six core TikZ packages aligned with \ourBench{}'s diagram domains.}
\label{tab:appx_mcp_extractors}
\renewcommand{\arraystretch}{1.1}
\setlength{\tabcolsep}{4.mm}{
\resizebox{0.85\linewidth}{!}{%
\begin{tabular}{l l r l l}
\toprule
\textbf{Package} & \textbf{Domain(s)} & \textbf{\#Files} & \textbf{Doc.\ Format} & \textbf{Extraction Strategy} \\
\midrule
\texttt{tikz-pgf}      & Graph, Geom., 3D & ${\sim}$150 & \texttt{codeexample}    & Recursive scan of PGF manual tree \\
\texttt{pgfplots}      & Charts, 3D       & ${\sim}$80  & \texttt{codeexample}    & Recursive file traversal with axis detection \\
\texttt{circuitikz}    & Circuit          & 4           & \texttt{LTXexample}     & Component catalog + showexpl examples \\
\texttt{tkz-euclide}   & Planar Geom.     & 32          & \texttt{tkzexample}     & Multi file orchestration across submodules \\
\texttt{chemfig}       & Chemistry        & 5           & \texttt{\textbackslash exemple} macro & Custom delimiter parsing \\
\texttt{tikz-network}  & Graph Struct.    & 1           & \texttt{docspec}/\texttt{lstlisting} & Command and option extraction \\
\bottomrule
\end{tabular}}
}
\end{table}

\vspace{0.5em}
\noindent\textbf{Stage 2: Knowledge base construction.}
All extracted items are normalized into a unified JSON schema. Each item stores its type, package, name, description, relevant code or syntax, and source file. The merged knowledge base contains 8{,}809 items in total, mainly executable examples and command specifications.

\vspace{0.5em}
\noindent\textbf{Stage 3: MCP server deployment.}
The structured knowledge base is converted into 411 Mintlify documentation pages and exposed through an MCP endpoint. The server provides a single tool, \texttt{SearchLaTeXKnowledgeBase}, which models can call during inference to retrieve relevant documentation fragments.

\vspace{0.5em}
\noindent\textbf{Query interface.}
The tool accepts a natural language query and optional filters, and returns ranked results such as code examples, command syntax, and usage notes. It supports keyword search, exact phrase search, command lookup, and Boolean operators. Table~\ref{tab:appx_mcp_queries} shows representative queries across benchmark domains.

\begin{table}[!tbp]
\centering
\caption{Example \texttt{SearchLaTeXKnowledgeBase} queries for benchmark domains.}
\label{tab:appx_mcp_queries}
\renewcommand{\arraystretch}{1.1}
\setlength{\tabcolsep}{4.mm}{
\resizebox{0.85\linewidth}{!}{%
\begin{tabular}{l l l}
\toprule
\textbf{Domain} & \textbf{Example Query} & \textbf{Returned Types} \\
\midrule
Charts        & \texttt{"addplot bar chart stacked"}                & Executable examples, command specs \\
Graph Struct. & \texttt{"tikz node positioning arrow styles"}       & Executable examples, command specs \\
Planar Geom.  & \texttt{"tkz-euclide angle bisector mark right angle"} & Executable examples, command specs \\
Circuit       & \texttt{"circuitikz resistor capacitor parallel"}   & Component definitions, examples \\
3D Shapes     & \texttt{"pgfplots 3d surface plot axis options"}    & Executable examples, command specs \\
Chemistry     & \texttt{"chemfig bond angle double bond ring"}      & Executable examples, key value options \\
\bottomrule
\end{tabular}}
}
\end{table}

\vspace{0.5em}
\noindent\textbf{Integration with MLLMs.}
During agentic evaluation, models receive the \texttt{SearchLaTeXKnowledgeBase} tool description in the system prompt and may invoke it as needed. The MCP design is model agnostic, so any MCP compatible client can connect to the same server without model specific tool implementations.

\subsection{Indexed Documentation List}
\label{sec:appx_mcp_docs}

The knowledge base indexes official documentation from 19 \LaTeX{} packages. The six core packages are directly aligned with the six diagram domains in \ourBench{}. The remaining packages either map to one of these six domains or provide domain agnostic support that can be used across all benchmark categories. Table~\ref{tab:appx_mcp_full_list} summarizes the indexed packages. For utility packages that are not tied to a single diagram type, we label the benchmark domain as \textit{All} rather than leaving the field empty.

\begin{table}[!tbp]
\centering
\caption{Indexed TikZ package documentation. Only the most important statistics are shown.}
\label{tab:appx_mcp_full_list}
\renewcommand{\arraystretch}{1.10}
\setlength{\tabcolsep}{4.mm}{
\resizebox{0.8\linewidth}{!}{%
\begin{tabular}{l l r r r r}
\toprule
\textbf{Package} & \textbf{Benchmark Domain(s)} & \textbf{Total} & \textbf{Exec.} & \textbf{Cmd.} & \textbf{\%} \\
\midrule
\rowcolor{softgray} \multicolumn{6}{l}{\textbf{Core Packages}} \\
\texttt{tikz-pgf}      & Graph, Geom., 3D   & 3{,}696 & 2{,}872 & 808  & 42.0 \\
\texttt{pgfplots}      & Charts, 3D         & 1{,}472 & 1{,}294 & 167  & 16.7 \\
\texttt{circuitikz}    & Circuit            & 816     & 501     & --   & 9.3  \\
\texttt{tkz-euclide}   & Planar Geom.       & 467     & 357     & 110  & 5.3  \\
\texttt{chemfig}       & Chemistry          & 290     & 248     & --   & 3.3  \\
\texttt{tikz-network}  & Graph Struct.      & 71      & 53      & 18   & 0.8  \\
\cmidrule(lr){1-6}
\textit{Core subtotal} &                    & \textit{6{,}812} & \textit{5{,}325} & \textit{1{,}103} & \textit{77.3} \\
\midrule
\rowcolor{softgray} \multicolumn{6}{l}{\textbf{Extended Packages}} \\
\texttt{xspace}        & All               & 1{,}107 & 107     & 1{,}000 & 12.6 \\
\texttt{pst-solides3d} & 3D Shapes         & 262     & 262     & --      & 3.0  \\
\texttt{fullpage}      & All               & 193     & 7       & 186     & 2.2  \\
\texttt{amscd}         & Graph Struct.     & 140     & 140     & --      & 1.6  \\
\texttt{tkz-base}      & Planar Geom.      & 97      & 97      & --      & 1.1  \\
\texttt{tkz-graph}     & Graph Struct.     & 92      & 92      & --      & 1.0  \\
\midrule
\rowcolor{softgray} \multicolumn{6}{l}{\textbf{Supplemental Packages}} \\
\texttt{tikz-cd}       & Graph Struct.     & 31      & 31      & --   & 0.4  \\
\texttt{comment}       & All               & 25      & 25      & --   & 0.3  \\
\texttt{tikz-3dplot}   & 3D Shapes         & 21      & 21      & --   & 0.2  \\
\texttt{tikz-qtree}    & Graph Struct.     & 15      & 15      & --   & 0.2  \\
\texttt{soul}          & All               & 8       & 4       & 4    & 0.1  \\
\texttt{forest}        & Graph Struct.     & 5       & 5       & --   & 0.1  \\
\texttt{tcolorbox}     & All               & 1       & 1       & --   & 0.0  \\
\midrule
\textbf{Total}         &                    & \textbf{8{,}809} & \textbf{6{,}132} & \textbf{2{,}293} & \textbf{100} \\
\bottomrule
\end{tabular}}
}
\end{table}

\vspace{0.5em}
\noindent\textbf{Core packages.}
The six core packages directly correspond to the six diagram domains in \ourBench{} and contribute 77.3\% of all indexed items. Among them, \path|tikz-pgf| and \path|pgfplots| provide the largest share of documentation, while \path|circuitikz|, \path|tkz-euclide|, \path|chemfig|, and \path|tikz-network| cover circuit, planar geometry, chemistry, and graph structure respectively.

\vspace{0.5em}
\noindent\textbf{Extended and supplemental packages.}
Among the remaining packages, several can still be mapped to benchmark domains. \path|pst-solides3d| and \path|tikz-3dplot| support 3D Shapes, while \path|amscd|, \path|tkz-graph|, \path|tikz-cd|, \path|tikz-qtree|, and \path|forest| support Graph Struct. \path|tkz-base| is mapped to Planar Geom. The rest, including \path|xspace|, \path|fullpage|, \path|comment|, \path|soul|, and \path|tcolorbox|, are domain agnostic utility packages, so we label them as \textit{All}.

\vspace{0.5em}
\noindent\textbf{Documentation sources.}
All items are extracted from official package documentation only. We do not include third party tutorials, forum posts, or community generated content, so the knowledge base remains authoritative and consistent.

\section{Additional Experimental Results}
\label{sec:appx_results}

\subsection{Foundational Results Per Diagram Type}
\label{sec:appx_per_domain}
We report Diagram-to-Code Parsing (D2C-P) results for 12 models and Diagram-to-Code Editing (D2C-E) results for 11 models, excluding TikZero+10B as it was not trained on editing samples. Figure~\ref{fig:appx_domain_heatmap} shows F1$_{\text{avg}}$ scores over each diagram domain: left panel for D2C-P, middle for preserve-only D2C-E, and right for edit-only D2C-E. The heatmap uses a diverging color scale from red (low) through white to {\color{metricgreen}green} (high). Figure~\ref{fig:appx_dqa_domain} reports DQA accuracy across 6 diagram domains.

For diagram-to-code parsing and editing, all models struggle on {3D Shapes}: D2C-P scores drop as low as 11.2\% (TikZero+10B), and even the best model stays below 40\%, indicating that current MLLMs fail to perceive three-dimensional spatial relationships and translate them into TikZ code. Most models perform well on {chemistry} diagrams, as the ChemFig package encodes molecular topology explicitly through bond-chain syntax. {Graph} diagrams are the relatively easiest editing domain for current MLLMs, likely because models easily ground the editing instructions into node--edge topology.  Closer inspection of the D2C-E heatmaps shows that preserve-only scores are higher than edit-only scores. For example, in charts, edit-only scores fall 10–20 points below preserve-only, indicating that current models retain untouched elements well but struggle to apply the instructed modifications. Moreover, the gap between open-source and closed-source models is larger on chemistry and circuit diagrams, especially for D2C-P.

\begin{figure}[!tbp]
\centering
\includegraphics[width=1.0\linewidth]{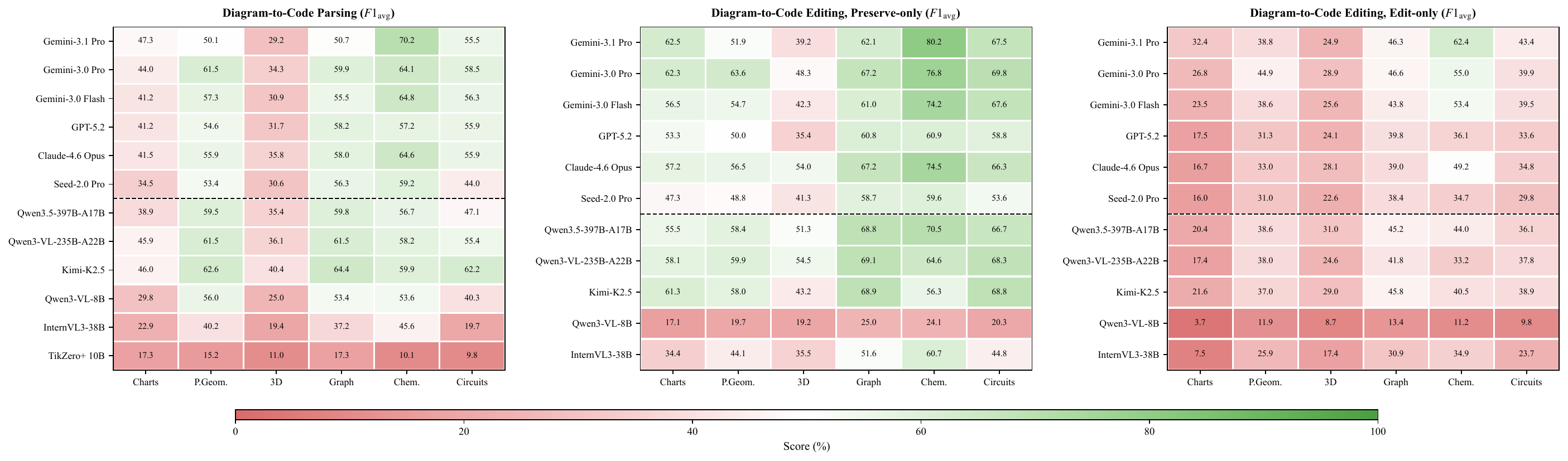}
\caption{(Updated) F1$_{\text{avg}}$ heatmap per diagram type: left panel for D2C-P, middle for preserve-only D2C-E, and right for edit-only D2C-E. Color encodes score magnitude (red$\to$white$\to${\color{metricgreen}green}); the dashed line separates closed-source (top) from open-source (bottom) models. 3D shapes is the most challenging domain across all models, while chemistry is relatively the easiest. }
\label{fig:appx_domain_heatmap}
\end{figure}

\subsection{F1 Scores Per Object}
\label{sec:appx_object_f1}


Figure~\ref{fig:appx_obj_radar} shows F1 score breakdowns for six representative models across three radars: D2C-P (left), D2C-E preserve-only (center), and D2C-E edit-only (right). Each radar plots five axes: F1$_{\text{avg}}$, F1$_{\text{type}}$, F1$_{\text{bbox}}$, F1$_{\text{color}}$, and F1$_{\text{text}}$, with axis scales fitted to each radar's value range.

A consistent finding is that F1$_{\text{bbox}}$ is dramatically lower than other object attributes. In D2C-P, all model polygons collapse toward the center on the F1$_{\text{bbox}}$ axis. This asymmetry persists in D2C-E preserve-only evaluation and becomes even more extreme in edit-only evaluation, confirming that spatial positioning remains the primary bottleneck across both parsing and editing tasks. Comparing the three panels shows a clear degradation: preserve-only scores are substantially higher than D2C-P on most axes, especially text and color where models benefit from retaining existing    
code, while edit-only scores drop sharply. Among the six models, Gemini-3.0 Pro achieves the largest polygon area across all three panels, while Qwen3-VL-8B shows the most compact polygons.

Table~\ref{tab:appx_d2cp_full} reports the per-object F1 scores on D2C-P, code-level (CrystalBLEU) and image-level (SSIM, CLIP, FID, LPIPS) metrics, complementing the aggregated F1$_{\text{avg}}$ in Table~\ref{tab:main_results}. F1$_{\text{bbox}}$ is the lowest across all models, confirming the spatial-positioning bottleneck identified in Figure~\ref{fig:appx_obj_radar}.   

\begin{table}[!tbp]
\centering
\caption{Three-level metric results of 12 models on D2C-P. $\downarrow$ means lower is better. \textbf{Bold} is best in class; \underline{underline} is second best.}
\label{tab:appx_d2cp_full}
\setlength{\tabcolsep}{4.mm}{
\resizebox{\linewidth}{!}{%
\begin{tabular}{@{}l ccccc c cccc@{}}
\toprule
& \multicolumn{5}{c}{\textit{Object-Level}} & \textit{Code-Level} & \multicolumn{4}{c}{\textit{Image-Level}} \\
\cmidrule(lr){2-6} \cmidrule(lr){7-7} \cmidrule(lr){8-11}
\textbf{Model} & F1$_{\text{type}}$ & F1$_{\text{text}}$ & F1$_{\text{color}}$ & F1$_{\text{bbox}}$ & F1$_{\text{avg}}$ & CBLEU & SSIM & CLIP & FID${\downarrow}$ & LPIPS${\downarrow}$ \\
\midrule
\multicolumn{11}{@{}l}{\textit{Closed-Source}} \\
Gemini-3.1 Pro       & 59.9 & 61.7 & 61.2 & \textbf{16.9} & 49.94 & \textbf{32.88} & 54.01 & 70.64 & \textbf{7.66} & 43.15 \\
Gemini-3.0 Pro       & \textbf{66.8} & \underline{69.2} & \textbf{69.9} & \underline{12.7} & \textbf{54.64} & \underline{32.76} & \underline{62.13} & 82.75 & \underline{8.88} & \textbf{35.14} \\
Gemini-3.0 Flash     & \underline{63.3} & 66.1 & 65.3 & 10.1 & 51.19 & 30.67 & 60.45 & 81.63 & 9.21 & \underline{36.59} \\
GPT-5.2              & 62.7 & 68.4 & \underline{66.2} & 8.0 & 51.35 & 28.81 & \textbf{62.66} & \textbf{88.29} & 12.74 & 37.53 \\
Claude-4.6 Opus      & 62.7 & \textbf{71.1} & 66.0 & 9.2 & \underline{52.23} & 31.27 & 61.11 & \underline{86.36} & 13.79 & 42.63 \\
Seed-2.0 Pro         & 59.8 & 62.4 & 60.3 & 7.8 & 47.57 & 32.56 & 60.14 & 84.93 & 14.66 & 44.30 \\
\midrule
\multicolumn{11}{@{}l}{\textit{Open-Source}} \\
Qwen3.5-397B-A17B    & 61.8 & 66.2 & 63.6 & 14.0 & 51.38 & 32.92 & 61.86 & 85.83 & \underline{11.06} & \underline{39.28} \\
Qwen3-VL-235B-A22B   & \underline{66.2} & \underline{70.9} & \underline{66.8} & \underline{16.6} & \underline{55.11} & \textbf{37.92} & \underline{61.87} & \underline{87.93} & 12.92 & 44.86 \\
Kimi-K2.5            & \textbf{69.0} & \textbf{73.3} & \textbf{70.5} & \textbf{17.1} & \textbf{57.48} & \underline{36.13} & \textbf{64.35} & \textbf{90.26} & \textbf{10.03} & \textbf{37.95} \\
Qwen3-VL-8B          & 52.7 & 58.4 & 54.5 & 13.0 & 44.66 & 33.31 & 54.78 & 78.84 & 15.90 & 55.52 \\
InternVL3-38B        & 38.1 & 41.2 & 38.7 & 7.9 & 31.47 & 30.91 & 50.75 & 71.28 & 19.67 & 62.41 \\
TikZero+ 10B         & 25.5 & 6.8 & 26.4 & 3.1 & 15.43 & 17.19 & 33.99 & 52.59 & 24.55 & 78.41 \\
\bottomrule
\end{tabular}%
}
}
\end{table}

Table~\ref{tab:appx_d2ce_full} reports the per-object F1 scores on D2C-E with preserve-only and edit-only splits, alongside code-level (CrystalBLEU) and image-level (SSIM, CLIP, FID, LPIPS) metrics. For preserve-only elements, models achieve moderate scores on type, text, and color, but F1$_{\text{bbox}}$ remains the lowest. For edit-only elements, all scores drop substantially, \eg, F1$_{\text{type}}$ shows the largest absolute gap, indicating that models struggle to instantiate correct object types for required editing elements.

\subsection{DQA Analysis Per Diagram Type}
\label{sec:appx_dqa_domain}

Table~\ref{tab:appx_dqa_domain} reports per-domain DQA accuracy across all 11 evaluated models (TikZero+\,10B is excluded as it is not trained on question answering). Fig.~\ref{fig:appx_dqa_domain} provides a complementary visual comparison using grouped bar charts.

{Graph structures} is the easiest domain for most models, with top-performing models exceeding 90\% accuracy (\eg, Gemini-3.1 Pro at 91.34\%). This is likely because graph-related questions often involve counting nodes or identifying connectivity patterns, which are relatively straightforward visual reasoning tasks.

{3D Shapes} is consistently the hardest DQA domain, with even the best models scoring below 69\%. The difficulty stems from the need to reason about three-dimensional spatial relationships, such as viewing angles and surface properties, from rendered diagrams.

\begin{table}[!tbp]
\centering
\caption{  Three-level metric results of 11 models on D2C-E with preserve-only (p) and edit-only (e) splits.  $\downarrow$ means lower is better. \textbf{Bold} is best in class; \underline{underline} is second best.}
\label{tab:appx_d2ce_full}
\resizebox{\linewidth}{!}{%
\begin{tabular}{@{}l ccccc ccccc cc cccc@{}}
\toprule
& \multicolumn{5}{c}{\textit{Preserve-Only}} & \multicolumn{5}{c}{\textit{Edit-Only}} & \multicolumn{2}{c}{\textit{CBLEU (p\,/\,e)}} & \multicolumn{4}{c}{\textit{Image-Level}} \\
\cmidrule(lr){2-6} \cmidrule(lr){7-11} \cmidrule(lr){12-13} \cmidrule(lr){14-17}
\textbf{Model} & F1$_{\text{type}}$ & F1$_{\text{text}}$ & F1$_{\text{color}}$ & F1$_{\text{bbox}}$ & F1$_{\text{avg}}$ & F1$_{\text{type}}$ & F1$_{\text{text}}$ & F1$_{\text{color}}$ & F1$_{\text{bbox}}$ & F1$_{\text{avg}}$ & p & e & SSIM & CLIP & FID${\downarrow}$ & LPIPS${\downarrow}$ \\
\midrule
\multicolumn{17}{@{}l}{\textit{Closed-Source}} \\
Gemini-3.1 Pro       & 70.4 & 70.7 & 69.1 & \textbf{32.2} & 60.60 & \textbf{43.6} & 60.4 & \textbf{52.1} & \textbf{7.3} & \textbf{40.84} & 42.02 & \textbf{2.98} & 57.63 & 75.32 & \textbf{4.62} & \underline{41.80} \\
Gemini-3.0 Pro       & \textbf{77.4} & \textbf{78.4} & \textbf{75.7} & \underline{28.9} & \textbf{65.12} & \underline{39.9} & \textbf{64.9} & \underline{51.2} & \underline{5.0} & \underline{40.24} & \textbf{44.77} & \underline{2.34} & \textbf{61.21} & \underline{83.55} & 5.84 & \textbf{39.29} \\
Gemini-3.0 Flash     & 71.2 & 72.7 & 69.0 & 24.2 & 59.29 & 37.1 & 58.7 & 48.8 & 4.3 & 37.24 & 42.21 & 2.17 & 57.17 & 77.71 & \underline{5.54} & 43.13 \\
GPT-5.2              & 66.8 & 70.3 & 64.5 & 20.8 & 55.59 & 28.7 & 53.3 & 40.3 & 2.3 & 31.15 & 38.75 & 1.15 & 54.71 & 75.94 & 8.61 & 47.77 \\
Claude-4.6 Opus      & \underline{75.1} & \underline{78.2} & \underline{72.7} & 24.3 & \underline{62.56} & 25.4 & \underline{60.5} & 39.0 & 3.0 & 31.95 & \underline{43.59} & 1.42 & \underline{59.54} & \textbf{84.02} & 8.75 & 46.14 \\
Seed-2.0 Pro         & 64.9 & 66.2 & 61.3 & 19.8 & 53.03 & 28.7 & 50.0 & 39.5 & 2.3 & 30.14 & 37.49 & 1.41 & 51.48 & 72.24 & 10.15 & 54.21 \\
\midrule
\multicolumn{17}{@{}l}{\textit{Open-Source}} \\
Qwen3.5-397B-A17B    & \underline{74.0} & \underline{75.4} & \textbf{71.1} & 30.5 & 62.75 & \underline{35.2} & \underline{60.1} & \textbf{45.3} & 4.7 & \underline{36.35} & \textbf{43.66} & \underline{1.91} & \textbf{59.14} & \underline{81.72} & \underline{7.08} & \textbf{44.18} \\
Qwen3-VL-235B-A22B   & \textbf{75.3} & \textbf{77.0} & 70.3 & \underline{32.6} & \textbf{63.79} & 30.0 & 57.9 & 40.9 & \underline{4.8} & 33.38 & \underline{43.05} & 1.85 & 58.06 & \textbf{82.08} & 8.63 & 50.03 \\
Kimi-K2.5            & 73.8 & 74.9 & \underline{70.5} & \textbf{33.3} & \underline{63.12} & \textbf{35.7} & \textbf{60.4} & \underline{44.6} & \textbf{5.4} & \textbf{36.52} & 42.11 & \textbf{2.02} & \underline{58.77} & 80.98 & \textbf{5.89} & \underline{45.11} \\
Qwen3-VL-8B          & 24.9 & 26.3 & 23.5 & 10.4 & 21.26 & 8.6 & 17.2 & 12.8 & 1.2 & 9.95 & 16.29 & 0.51 & 22.07 & 31.81 & 18.16 & 82.64 \\
InternVL3-38B        & 53.6 & 55.3 & 49.1 & 20.9 & 44.72 & 20.5 & 38.5 & 29.4 & 2.3 & 22.69 & 33.13 & 1.29 & 48.37 & 67.54 & 13.39 & 65.13 \\
\bottomrule
\end{tabular}%
}
\end{table}

Among open-source models, {Qwen3-VL-8B} shows particularly uneven domain performance: it achieves 66.71\% on Circuits but only 27.05\% on chemistry, suggesting that smaller models lack generalization across diverse diagram types. Conversely, {Qwen3.5-397B-A17B} maintains relatively balanced accuracy (66--88\%) across all domains, approaching closed-source performance.

\begin{table}[!t]
\centering
\caption{DQA accuracy (\%) per diagram type. \textbf{Bold} is best in class; \underline{underline} is second.}
\label{tab:appx_dqa_domain}
\renewcommand\arraystretch{1.1}
\setlength{\tabcolsep}{4.mm}{
 \resizebox{0.8\linewidth}{!}{
\begin{tabular}{lccccccc}
\toprule
\textbf{Model} & Charts & P.\,Geom. & 3D & Graph & Chem. & Circuits & \cellcolor{gray!15}\textit{Avg.} \\
\midrule
\multicolumn{8}{@{}l}{\textit{Closed-Source}} \\
Gemini-3.1 Pro       & \textbf{85.95} & \underline{83.63} & \underline{67.69} & \textbf{91.34} & \underline{86.07} & \underline{84.29} & \cellcolor{gray!15}\underline{86.29} \\
Gemini-3.0 Pro       & \underline{85.08} & \textbf{87.16} & 67.69 & \underline{90.88} & 85.79 & \textbf{84.58} & \cellcolor{gray!15}\textbf{86.46} \\
Gemini-3.0 Flash     & 82.95 & 78.80 & \textbf{68.57} & 89.29 & \textbf{86.89} & 84.01 & \cellcolor{gray!15}84.07 \\
GPT-5.2              & 82.08 & 81.93 & 63.96 & 83.87 & 81.42 & 83.43 & \cellcolor{gray!15}81.67 \\
Claude-4.6 Opus      & 70.42 & 79.96 & 60.00 & 61.22 & 73.77 & 78.39 & \cellcolor{gray!15}68.75 \\
Seed-2.0 Pro         & 70.75 & 76.12 & 60.22 & 79.73 & 80.05 & 82.42 & \cellcolor{gray!15}75.90 \\
\midrule
\multicolumn{8}{@{}l}{\textit{Open-Source}} \\
Qwen3.5-397B-A17B    & \textbf{82.03} & \textbf{83.45} & \textbf{66.15} & \textbf{87.50} & \textbf{86.34} & \textbf{81.12} & \cellcolor{gray!15}\textbf{83.42} \\
Qwen3-VL-235B-A22B   & 58.55 & 72.99 & 55.82 & 60.88 & 66.12 & 76.08 & \cellcolor{gray!15}63.60 \\
Kimi-K2.5            & \underline{76.20} & \underline{81.75} & \underline{64.40} & \underline{83.73} & \underline{80.60} & \underline{80.12} & \cellcolor{gray!15}\underline{79.74} \\
Qwen3-VL-8B          & 33.06 & 66.46 & 39.12 & 47.70 & 27.05 & 66.71 & \cellcolor{gray!15}47.12 \\
InternVL3-38B        & 55.17 & 75.31 & 49.01 & 48.23 & 48.36 & 69.74 & \cellcolor{gray!15}56.39 \\
\bottomrule
\end{tabular}
}
}
\end{table}

\subsection{Agentic Tool-Call Statistics}
\label{sec:appx_tool_stats}

Table~\ref{tab:appx_tool_adoption} reports tool-calling statistics in the agentic settings ({\color{tooltxt}S3}/{\color{tooltxt}S8}/{\color{plantxt}S16} in Table~\ref{tab:context_settings}).

\vspace{0.3em}
\noindent\textbf{Adoption.}
 All six models invoke the tool to some degree, but adoption varies widely (0.2\%–96.5\%) across tasks and models. On D2C-P and D2C-E, Claude-4.6~Opus adopts tools most aggressively (93.7\%/96.5\%), followed by Gemini-3.0~Pro (45.0\%/48.0\%) and Qwen3-VL-8B (38.7\%/43.8\%), while Seed-2.0~Pro is most conservative (8.3\%/10.0\%).
DQA triggers much lower tool use across all models, since question-answering rarely requires code-level documentation lookup.
Within D2C-E, domains requiring specialized packages, \eg, circuit (\texttt{circuitikz}, up to 80\%) and chemistry (\texttt{chemfig}, up to 65\%), elicit the highest adoption, while graph diagrams drawn with basic \texttt{tikz} commands show the lowest (1--13\% for most models).

\vspace{0.3em}
\noindent\textbf{Intensity.}
Claude-4.6~Opus averages $\bar{n}$=2.83 calls per question. We inspect its calling pattern: it first queries package syntax, then specific commands, final usage examples. This structured pattern  correlates with its consistent gains under tool use. Gemini-3.1~Pro shows the highest average calls among tool-invoking questions ($\bar{n}_{+}$=2.37 in D2C-E) despite low adoption (15\%), indicating excessive querying loops, \eg, repeated searches without synthesizing results, leading to context rot at the token limit.  
Qwen3-VL-8B shows the opposite: moderate adoption (43.8\%) but low $\bar{n}_{+}$=1.12, typically issuing only a single query per question.

\begin{figure}[t]
\centering
\includegraphics[width=0.96\linewidth]{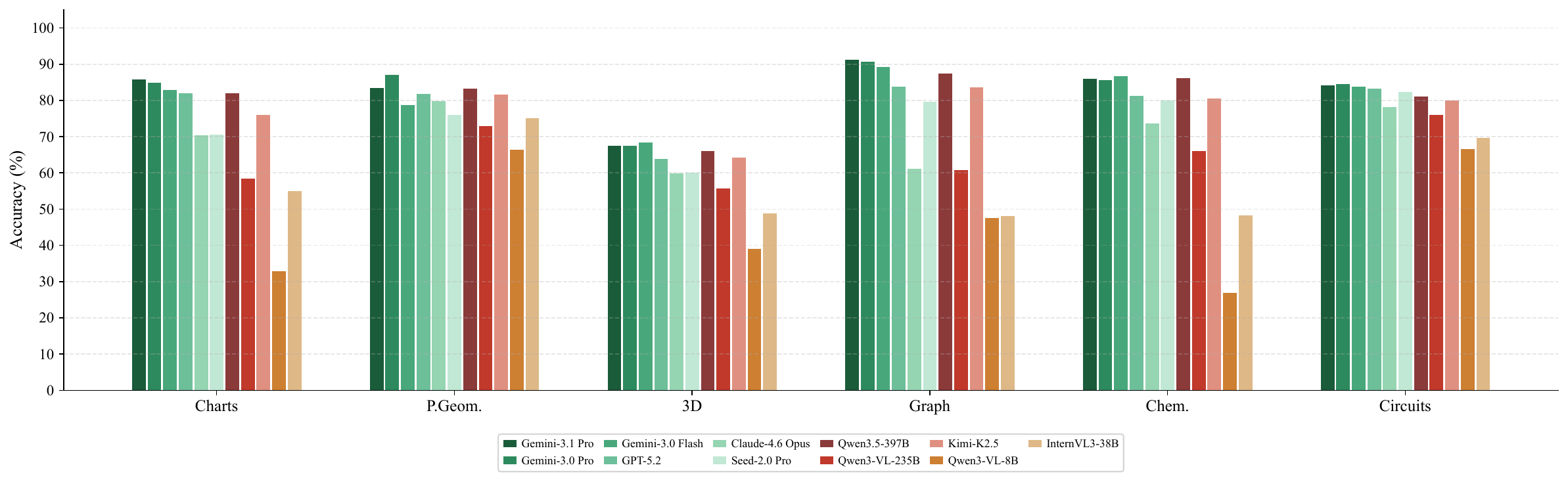}
\caption{(Updated) DQA accuracy (\%) per diagram type (an alternative visualization of Table~\ref{tab:appx_dqa_domain}). Bars distinguish closed-source from open-source models.}
\label{fig:appx_dqa_domain}
\end{figure}

\begin{table}[!tbp]
\centering
\caption{ Tool-calling statistics in tool-use agentic settings ({\color{tooltxt}S3}/{\color{tooltxt}S8} = \texttt{+Tool}; {\color{plantxt}S16} = \texttt{+Tool\&TikZ Codes}). $N$ is the number of questions; {Adopt.} is the percentage of questions where the model invokes the tool at least once ($\geq$1) out of all $N$ questions; $\bar{n}$ is the total call count divided by $N$; $\bar{n}_{+}$ is the total call count divided by the number of questions where the model called the tool at least once.}
\label{tab:appx_tool_adoption}
\renewcommand{\arraystretch}{1.1}
\setlength{\tabcolsep}{4.mm}{
\resizebox{0.95\linewidth}{!}{%
\begin{tabular}{lrrr rrr rrr}
\toprule
& \multicolumn{3}{c}{\textbf{D2C-P} ({\color{tooltxt}S3}, $N$=300)} & \multicolumn{3}{c}{\textbf{D2C-E} ({\color{tooltxt}S8}, $N$=600)} & \multicolumn{3}{c}{\textbf{DQA} ({\color{plantxt}S16}, $N$=600)} \\
\cmidrule(lr){2-4}\cmidrule(lr){5-7}\cmidrule(lr){8-10}
\textbf{Model} & {Adopt. (\% )} & $\bar{n}$ & $\bar{n}_{\!+}$ & {Adopt. (\% )} & $\bar{n}$ & $\bar{n}_{\!+}$ & {Adopt. (\% )} & $\bar{n}$ & $\bar{n}_{\!+}$ \\
\midrule
Claude-4.6 Opus & \textbf{93.7\%} & \textbf{2.70} & 2.88 & \textbf{96.5\%} & \textbf{2.83} & 2.93 & 2.7\% & 0.05 & 1.75 \\
Gemini-3.0 Pro  & 45.0\% & 0.75 & 1.67 & 48.0\% & 0.79 & 1.65 & \textbf{14.0\%} & \textbf{0.22} & 1.56 \\
Qwen3-VL-8B     & 38.7\% & 0.43 & 1.10 & 43.8\% & 0.49 & 1.12 & 0.2\% & 0.00 & 1.00 \\
GPT-5.2         & 26.3\% & 0.49 & 1.87 & 23.3\% & 0.43 & 1.84 & 5.7\% & 0.08 & 1.47 \\
Gemini-3.1 Pro  & 18.7\% & 0.44 & \textbf{2.34} & 15.0\% & 0.35 & \textbf{2.37} & 1.2\% & 0.02 & 1.43 \\
Seed-2.0 Pro    & 8.3\% & 0.09 & 1.04 & 10.0\% & 0.10 & 1.02 & 0.8\% & 0.01 & 1.00 \\
\bottomrule
\end{tabular}}
}
\end{table}

\vspace{0.3em}
\noindent\textbf{Response and execution rates.}
Table~\ref{tab:appx_tool_failure} compares two rates between direct ({\color{basetxt}S1}/{\color{basetxt}S6}) and tool-use agentic settings ({\color{tooltxt}S3}/{\color{tooltxt}S8}) for D2C-P/D2C-E: (1) {Response rate} (\%): the percentage of questions where the model outputs a complete response; failures occur when the model exceeds the context length or enters tool-calling loops without producing final code; (2) {Execution rate} (\%): the percentage of generated code that compiles successfully; failures occur when the output code contains syntax errors. Gemini-3.1~Pro suffers the most: tool access reduces the response rate by 25.0\% (D2C-P) and 23.8\% (D2C-E) due to the querying loops described above. Claude-4.6~Opus benefits most: tool access improves execution rate by +4.2\% on D2C-E. Other models show minimal changes.    

\begin{table}[!tbp]
\centering
\caption{Response and execution rates (\%) in direct ({\color{basetxt}S1}/{\color{basetxt}S6}) \vs tool-use settings ({\color{tooltxt}S3}/{\color{tooltxt}S8}). $\Delta$: change from direct to \texttt{+Tool} (\textcolor{ForestGreen}{$\uparrow$} gain, $\downarrow$ drop).}     
\label{tab:appx_tool_failure}
\renewcommand{\arraystretch}{1.1}
\setlength{\tabcolsep}{4.mm}{
\resizebox{\linewidth}{!}{%
\begin{tabular}{@{}l rrr rrr rrr rrr@{}}
\toprule
& \multicolumn{6}{c}{\textbf{D2C-P}} & \multicolumn{6}{c}{\textbf{D2C-E}} \\
\cmidrule(lr){2-7}\cmidrule(lr){8-13}
& \multicolumn{3}{c}{\textbf{Resp.\,Rate.}} & \multicolumn{3}{c}{\textbf{Exec.\,Rate.}} & \multicolumn{3}{c}{\textbf{Resp.\,Rate.}} & \multicolumn{3}{c}{\textbf{Exec.\,Rate.}} \\
\cmidrule(lr){2-4}\cmidrule(lr){5-7}\cmidrule(lr){8-10}\cmidrule(lr){11-13}
\textbf{Model} & {\color{basetxt}S1} & {\color{tooltxt}S3} & $\Delta$ & {\color{basetxt}S1} & {\color{tooltxt}S3} & $\Delta$ & {\color{basetxt}S6} & {\color{tooltxt}S8} & $\Delta$ & {\color{basetxt}S6} & {\color{tooltxt}S8} & $\Delta$ \\
\midrule
Claude-4.6 Opus & 100.0 & 100.0 & {\color{gray}0.0} & 91.3 & 93.3 & {\color{ForestGreen}+2.0} & 100.0 & 99.7 & -0.3 & 89.2 & 93.3 & {\color{ForestGreen}+4.2} \\
Seed-2.0 Pro    & 100.0 & 100.0 & {\color{gray}0.0} & 89.7 & 92.3 & {\color{ForestGreen}+2.7} & 99.7 & 100.0 & {\color{ForestGreen}+0.3} & 90.3 & 89.2 & -1.2 \\
GPT-5.2         & 100.0 & 97.7 & -2.3 & 95.3 & 91.7 & -3.7 & 100.0 & 98.2 & -1.8 & 91.3 & 92.7 & {\color{ForestGreen}+1.3} \\
Gemini-3.0 Pro  & 100.0 & 99.7 & -0.3 & 86.3 & 83.7 & -2.7 & 100.0 & 99.3 & -0.7 & 90.0 & 86.8 & -3.2 \\
Gemini-3.1 Pro  & 100.0 & 75.0 & -25.0 & 73.3 & 60.7 & -12.7 & 100.0 & 76.2 & -23.8 & 79.7 & 63.2 & -16.5 \\
Qwen3-VL-8B     & 99.7 & 100.0 & {\color{ForestGreen}+0.3} & 88.0 & 87.3 & -0.7 & 99.3 & 100.0 & {\color{ForestGreen}+0.7} & 86.8 & 86.3 & -0.5 \\
\bottomrule
\end{tabular}}
}
\end{table}

\section{Limitations and Future Work}
\label{sec:appx_limitations}

\noindent\textbf{Agentic evaluation.}
As acknowledged (\S\ref{sec:eval_set}), we do not claim \ourBench{} evaluates the full spectrum of agentic ability in the vibe writing framework. Vibe writing platforms (\eg, Prism~\cite{prism}) are becoming increasingly popular, as any researcher can prompt models in natural language to generate academic papers directly in \LaTeX{} format.  Common needs include inserting a diagram into the manuscript as compilable \LaTeX{} TikZ code, applying specific edits based on the paper content, or describing the domain-specific meaning of diagram symbols. Our evaluation settings target such multimodal aspects of vibe writing through diagram-to-code and diagram understanding tasks. 

However, the vibe writing workspace involves broader capabilities such as text drafting, citation management, and layout formatting, which remain outside our scope and would require designing agentic evaluation settings for purely textual abilities.  Moreover, broader agentic capabilities such as long-horizon task decomposition, multi-tool orchestration across heterogeneous environments, self-correction via execution feedback, and interactive collaboration with human users are also not covered by the current benchmark.

\vspace{0.5em}
\noindent\textbf{Code representation.} \ourBench{} adopts \LaTeX{} TikZ as the code representation, whose output compiles inline within \LaTeX{} authoring environments. But, in practice, researchers also create scientific
diagrams with Python libraries (\eg, Matplotlib, Seaborn), SVG/HTML tools, or other \LaTeX{} drawing packages (\eg, \texttt{pstricks}, \texttt{xy-pic}), which are currently not covered by the evaluation settings of \ourBench{}. 

\vspace{0.5em}
\noindent\textbf{Future directions.}
In future work, we plan to extend agentic settings to evaluate: (1)~broader textual and layout capabilities in scientific writing, such as text-to-TikZ generation, table construction from data, and mathematical derivation formatting; and (2)~richer agentic workflows, including execution feedback loops, iterative human-in-the-loop refinement, and multi-tool orchestration, that  reflect the real-world demands of vibe writing. For code representation, cross-language conversion pipelines that translate Python or SVG diagrams into TikZ would expand coverage while preserving inline \LaTeX{} compilation, and supporting diagram import via image upload would allow vibe writing workspaces to ingest figures.  Alternatively, extending evaluation to additional \LaTeX{} drawing packages beyond TikZ (\eg, \texttt{pstricks}, \texttt{xy-pic}) would broaden coverage within the \LaTeX{} ecosystem directly.  For diagram understanding and reasoning, we plan to move toward expert-level questions, and to incorporate richer context such as figure captions from the surrounding  paper, enabling evaluation of diagram comprehension grounded in the full manuscript.

We hope future versions of \ourBench{} can provide more comprehensive insights and guide targeted improvements by pinpointing where models fail: whether in foundational abilities (perception, coding, knowledge, reasoning) or agentic abilities (context utilization, tool use, state management, planning).  We do   
not foresee negative societal consequences; the benchmark is constructed from publicly available TikZ documentation and contains no private or sensitive data.

\section{Acknowledgement}
\label{sec:author_state}
We thank all 13 annotators for their contributions to data quality assurance; together, they verified over 2,000 scientific diagrams through cross-validation across six domains. Three annotators are co-authors of this paper. Among the remaining ten contributors, one is a research scientist at NetMind.ai, eight are PhD students at Nanjing University of Science and Technology (NJUST), and one is a PhD student at Westlake University. We gratefully acknowledge these ten collaborators and detail below the number of annotated samples and the domains each covered:

\begin{itemize}
    \item Xinyu Miao from NJUST — 419 cases, covering 3D Shape, Charts, Chemical Expressions, Graph Structures, Planar Geometry
    \item Yiyou Gao from NJUST — 331 cases, covering 3D Shape, Charts, Graph Structures, Planar Geometry
    \item Ruolin Wang from NJUST — 271 cases, covering 3D Shape, Charts, Graph Structures, Planar Geometry
    \item Xinyao Hu from NetMind.ai — 208 cases, covering 3D Shape, Charts, Graph Structures, Planar Geometry
    \item Jinhong Yang from NJUST — 200 cases, covering Graph Structures
    \item Zongxin Zhu  rom NJUST — 194 cases, covering Charts, Graph Structures, Planar Geometry
    \item Yu Wang from NJUST — 181 cases, covering Graph Structures, Planar Geometry
    \item Xinyu Zhang from NJUST — 170 cases, covering Charts, Circuit Diagrams, Graph Structures
    \item Xueji Fang  from Westlake University — 107 cases, covering 3D Shape, Charts, Graph Structures, Planar Geometry
    \item Wei Tang from NJUST — 11 cases, covering 3D Shape, Charts
\end{itemize}



\end{document}